%% file: main.tex
\documentclass{article} 
\usepackage{iclr2022_conference,times}

\usepackage{wrapfig}
\usepackage[utf8]{inputenc} 
\usepackage[T1]{fontenc}    
\usepackage[pagebackref=true]{hyperref}
\usepackage{url}            
\usepackage{booktabs}       
\usepackage{amsfonts}       
\usepackage{nicefrac}       
\usepackage{microtype}      
\usepackage{xcolor}         
\usepackage{graphicx}
\usepackage{enumitem}
\usepackage{caption}
\usepackage[ruled, vlined]{algorithm2e}
\SetKwComment{Comment}{$\triangleright$\ }{}
\newlength\savewidth

\input{math_definition}

\usepackage{xspace}
\newcommand{\ourmethod}{\method{unicorn-MAML}\xspace}
\newcommand{\eg}{{\em e.g.}}
\newcommand{\ie}{{\em i.e.}}

\title{How to Train Your MAML\\ to Excel in Few-Shot Classification}

\author{Han-Jia Ye \\
  \small{State Key Laboratory for Novel Software Technology, Nanjing University}\\
\And
  Wei-Lun Chao \\
  \small{The Ohio State University} \\
}

\usepackage{csquotes}

\iclrfinalcopy 
\begin{document}

\maketitle

\input{abstract}

\input{introduction}

\input{preliminary}
\input{approach-1}
\input{approach-2}
\input{approach-3}
\input{approach-4}
\input{disc}

\section*{Acknowledgments}
This research is supported by National Key R\&D Program of China (2020AAA0109401), CCF-Baidu Open Fund (2021PP15002000), NSFC (61773198, 61921006, 62006112, 62176117), Collaborative Innovation Center of Novel Software Technology and Industrialization, NSF of Jiangsu Province (BK20200313), NSF IIS-2107077, NSF OAC-2118240, NSF OAC-2112606, and the OSU GI Development funds. We are thankful for the generous support of computational resources by Ohio Supercomputer Center and AWS Cloud Credits for Research.
We thank S{\'e}bastien M.R. Arnold (USC) for helpful
discussions.

\section*{Reproducibility Statement}
The details of datasets, model architectures, hyper-parameters, and evaluation metrics are described in \autoref{ss_exp_setup} and \autoref{suppl_s_setup}.
Our code is available at \url{https://github.com/Han-Jia/UNICORN-MAML}, including the initialization weights pre-trained on the meta-training set and the checkpoints. 

{\small
\bibliographystyle{iclr2022_conference}
\bibliography{main}
}

\renewcommand{\thesection}{\Alph{section}}
\renewcommand{\thetable}{\Alph{table}}
\renewcommand{\thefigure}{\Alph{figure}}
\renewcommand{\theequation}{\Alph{equation}}

\newpage
\appendix
\begin{center}
	\textbf{\Large Appendix}
\end{center}
\input{suppl_content}

\end{document}

%% file: math_definition.tex
\usepackage{amssymb}
\usepackage{amsmath,amsfonts}
\usepackage{amsopn}
\usepackage{bm} 
\usepackage{multirow}
\usepackage{flushend}
\usepackage{tabularx}
\newcommand{\vct}[1]{\boldsymbol{#1}} 
\newcommand{\cst}[1]{\mathsf{#1}}  

\newcommand{\field}[1]{\mathbb{#1}}
\newcommand{\R}{\field{R}} 

\newcommand{\ProbOpr}[1]{\mathbb{#1}}

\newcommand{\expect}[2]{%
\ifthenelse{\equal{#2}{}}{\ProbOpr{E}_{#1}}
{\ifthenelse{\equal{#1}{}}{\ProbOpr{E}\left[#2\right]}{\ProbOpr{E}_{#1}\left[#2\right]}}} 
\newcommand{\var}[2]{%
\ifthenelse{\equal{#2}{}}{\ProbOpr{VAR}_{#1}}
{\ifthenelse{\equal{#1}{}}{\ProbOpr{VAR}\left[#2\right]}{\ProbOpr{VAR}_{#1}\left[#2\right]}}} 

\DeclareMathOperator{\argmax}{arg\,max}

\newcommand{\vtheta}{\vct{\theta}}

\newcommand{\vx}{{\vct{x}}}

\newcommand{\va}{\vct{a}}

\newcommand{\vw}{\vct{w}}

\newcommand{\vphi}{\vct{\phi}}

\newcommand{\sQ}{\mathcal{Q}}
\newcommand{\sS}{\mathcal{S}}

\newcommand{\sT}{\mathcal{T}}
\newcommand{\sL}{\mathcal{L}}
\newcommand{\IL}{\cst{InLoop}}
\newcommand{\eat}[1]{}
\newcommand{\method}[1]{\textsc{#1}}

%% file: abstract.tex

\begin{abstract}
Model-agnostic meta-learning (MAML) is arguably one of the most popular meta-learning algorithms nowadays. Nevertheless, its performance on few-shot classification is far behind many recent algorithms dedicated to the problem. In this paper, we point out several key facets of how to train MAML to excel in few-shot classification. First, we find that MAML needs a large number of gradient steps in its inner loop update, which contradicts its common usage in few-shot classification. Second, we find that MAML is sensitive to the class label assignments during meta-testing. Concretely, MAML meta-trains the initialization of an $N$-way classifier. These $N$ ways, during meta-testing, then have ``$N!$'' different permutations to be paired with a few-shot task of $N$ novel classes. We find that these permutations lead to a huge variance of accuracy, making MAML unstable in few-shot classification. Third, we investigate several approaches to make MAML permutation-invariant, among which meta-training a \emph{single vector to initialize all the $N$ weight vectors in the classification head} performs the best. On benchmark datasets like \emph{Mini}ImageNet and \emph{Tiered}ImageNet, our approach, which we name \ourmethod, performs on a par with or even outperforms many recent few-shot classification algorithms, \emph{without sacrificing MAML's simplicity.}
\end{abstract}

%% file: introduction.tex

\section{Introduction}
\label{s_intro}

Meta-learning is a sub-field of machine learning which attempts to search for the best learning strategy as the learning
experiences increases~\citep{thrun2012learning,lemke2015metalearning}.
Recent years have witnessed an abundance of new approaches on meta-learning~\citep{vanschoren2018meta,hospedales2020meta}, among which model-agnostic meta-learning (MAML)~\citep{FinnAL17Model,finn2018learning3} is one of the most popular algorithms, owing to its ``model-agnostic'' nature to incorporate different model architectures and its principled formulation to be applied to different problems. 
Concretely, MAML aims to learn a good \emph{model initialization} (through the outer loop optimization), which can then be quickly adapted to novel tasks given few examples (through the inner loop optimization).

However, in few-shot classification~\citep{VinyalsBLKW16Matching,SnellSZ17Prototypical} which many meta-learning algorithms are dedicated to, MAML's performance has been shown to fall far behind~\citep{wang2019simpleshot,chen2019closer,Triantafillou2019Meta}.

In this paper, we take a closer look at MAML on few-shot classification. The standard setup involves two phases, meta-training and meta-testing, in which MAML learns the model initialization during meta-training and applies it during meta-testing. In both phases, MAML receives multiple $N$-way $K$-shot tasks. 
Each task is an $N$-class classification problem provided with $K$ labeled support examples per class. After the (temporary) inner loop optimization using the support examples, the updated model from the initialization is then evaluated on the query examples of the same $N$ classes. The loss calculated on the query examples during mete-training is used to optimize the meta-parameters (\ie, the model initialization) through the outer loop. 
\emph{For consistency, we mainly study the scenario where meta-training and meta-testing use the same number of gradient steps in the inner loop optimization.}
 
More specifically, what MAML learns for few-shot classification is the initialization of an $N$-class classifier. Without loss of generality, we denote a classifier by $\hat{y} = \argmax_{c\in[N]} \vw_c^\top f_{\vphi}(\vx)$, where $f_{\vphi}$ is the feature extractor on an example $\vx$; $\{\vw_c\}_{c=1}^N$ are the weight vectors in the linear classification head. We use $\vtheta$ to represent the collection of meta-parameters $\{\vphi, \vw_1, \cdots, \vw_N\}$.

\begin{figure}[t]
    \centering
    \minipage{1\textwidth}
    \centering
    {\includegraphics[width=0.71\textwidth]{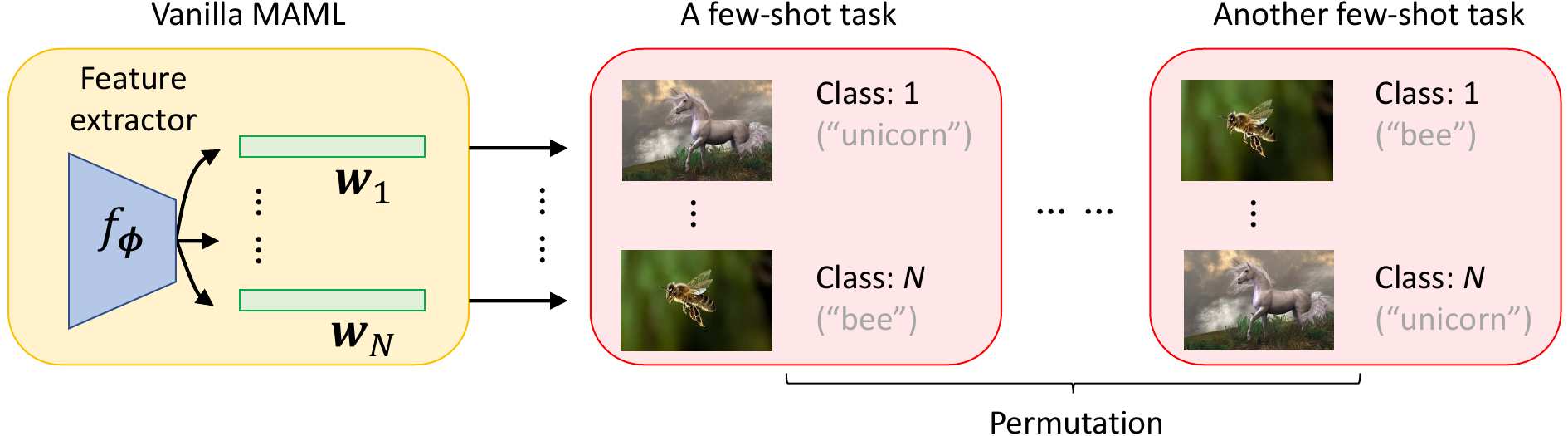}}
    \mbox{\small (a)}
    \endminipage
    \\ \vskip 5pt
    \minipage{1\textwidth}
    \centering
    {\includegraphics[width=0.71\textwidth]{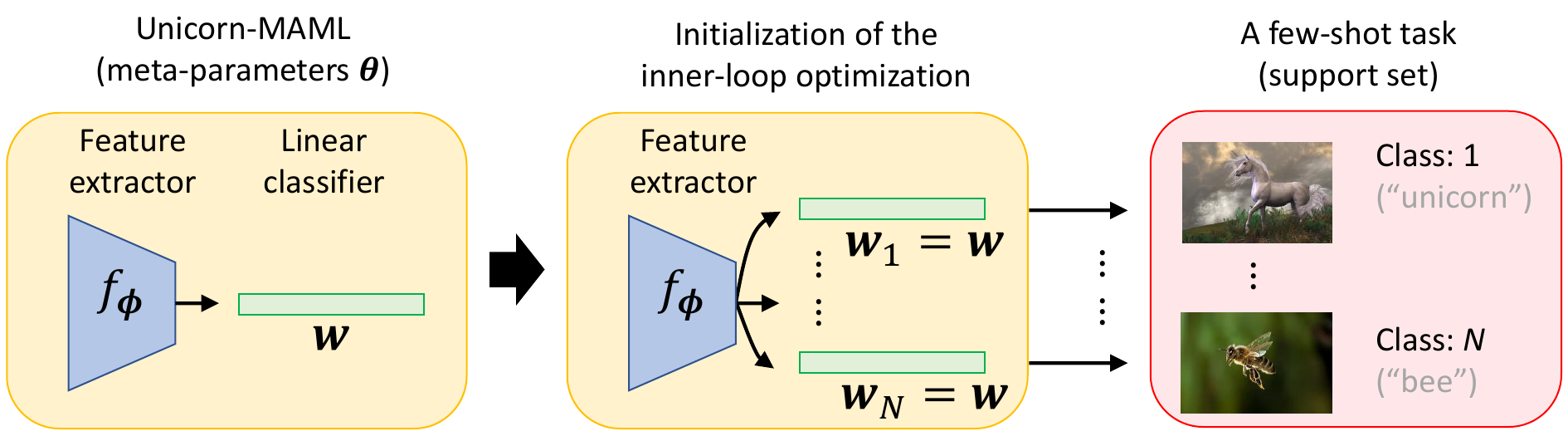}}
    \mbox{\small (b)}
    \endminipage
    \vskip -5pt
    \caption{\small \textbf{The problem of permutations in label assignments, and the illustration of \ourmethod.} (a) A vanilla MAML learns the initialization of $\vphi$ and $\{\vw_c\}_{c=1}^{N}$ (\ie, the $N$ weight vectors). Each of $\{\vw_c\}_{c=1}^{N}$ is paired with the corresponding class label $c\in[N]$ of a few-shot task. A few-shot task, however, may consist of the same set of semantic classes but in different permutations of class label assignments, leading to a larger variance in meta-testing accuracy. (b) In contrast, our \textbf{\ourmethod}, besides learning $\vphi$, learns only a single weight vector $\vw$ and uses it to initialize all the $N$ weight vectors $\{\vw_c\}_{c=1}^{N}$ at the beginning of the inner loop.
    That is, \ourmethod directly forces the learned model initialization to be permutation-invariant.}
    \label{fig:ourmethod}
    \vskip -10pt
\end{figure}

Our first observation is that \textbf{MAML needs a large number of gradient steps in the inner loop}. For example, on \emph{Mini}ImageNet~\citep{VinyalsBLKW16Matching} and \emph{Tiered}ImageNet~\citep{ren2018meta}, MAML's accuracy improves along with the increased number of gradient steps and achieves the highest around $15\sim20$ steps, which are much larger than the conventional usage of MAML~\citep{Antoniou2018How}. We attribute this to the behavior of the model initialization learned from mutually-exclusive tasks~\citep{yin2020meta}, which, without any further inner loop optimization, performs at the chance level (\ie, $\frac{100}{N}\%$ accuracy) on query examples, not only for the meta-testing tasks but also for the meta-training tasks. 
In other words, the initialized model needs many gradient steps to attain a high accuracy.

Our second observation is that \textbf{MAML is sensitive to the permutations of class label assignments during meta-testing}.
Concretely, 
when an $N$-way $K$-shot task arrives, MAML pairs the learned initialization of $\vw_c$ with the corresponding class label $c\in[N]$ of that task. The issue resides in the ``meaning'' of $c$ in a task. In the standard setup, each $N$-way task is created by drawing $N$ classes from a bigger pool of semantic classes (\eg, ``dog'', ``cat'', ``bird'', etc.), followed by a \emph{random} label re-assignment into $c\in\{1,\cdots,N\}$. In other words, the same set of $N$ semantic classes can be labeled totally differently into $\{1,\cdots,N\}$ and thus be paired with $\{\vw_c\}_{c=1}^N$ differently. Taking a five-way task for example, there are $5!=120$ permutations to pair the same set of five semantic classes to the linear classification head. In some of them, a class ``dog'' may be assigned to $c=1$; in some others, it may be assigned to $c\neq1$. \emph{While this randomness has been shown crucial in meta-training to help MAML prevent \emph{over-fitting}~\citep{rajendran2020meta,yao2021improving,yin2020meta}, we find that it makes the meta-testing phase unstable.} 
Specifically, different permutations can lead to drastically different meta-testing accuracy --- on average, the best permutation for each five-way one-shot task has $\sim 15\%$ higher accuracy than the worst permutation, on both datasets.

Building upon this observation, we investigate multiple approaches to \textbf{make MAML permutation-invariant}, either in the meta-testing phase alone or in both phases. We find that a simple solution --- \emph{meta-training only a single vector $\vw$ and using it to initialize the $N$ linear classifiers $\{\vw_c\}_{c=1}^{N}$} --- performs the best. 
We name this approach \ourmethod, as illustrated in \autoref{fig:ourmethod} (b).  
Concretely, at the beginning of the inner loop, \ourmethod duplicates $\vw$ into $\vw_c, \forall c\in[N]$. After the inner loop optimization, the meta-gradient with respect to $\{\vw_c\}_{c=1}^{N}$ are aggregated to update $\vw$ in the outer loop (during meta-training). This design not only makes \ourmethod permutation-invariant, but also ensures that no single model can solve all tasks at once without inner loop optimization, to prevent memorization over-fitting~\citep{yin2020meta}. 
On \emph{Mini}ImageNet~\citep{VinyalsBLKW16Matching}, \emph{Tiered}ImageNet~\citep{ren2018meta}, and CUB datasets~\citep{WahCUB_200_2011}, \ourmethod performs on a par with or even outperforms many recent few-shot classification algorithms, while preserving the simplicity of MAML without adding any extra network modules or learning strategies. 
Our code is available at \url{https://github.com/Han-Jia/UNICORN-MAML}.

%% file: preliminary.tex

\section{MAML for Few-Shot Classification}
\label{s_pre}
\subsection{Problem definition}
\label{ss_problem_def}
The goal of few-shot classification is to construct a classifier using limited labeled examples. The challenge is the potential over-fitting or poor generalization problem.
Following~\citep{VinyalsBLKW16Matching}, we define a few-shot classification problem as an $N$-way $K$-shot task, which has $N$ classes and each class has $K$ labeled support examples. We denote the labeled support set by $\mathcal{S} = \{(\vx_{i}, y_{i})\}_{i=1}^{N\times K}$; 
each $(\vx_{i}, y_{i})$ is a pair of an input (\eg, image) and a class label, where $y_i\in\{1,2,\cdots,N\}=[N]$.
The value of $K$ is small, \eg, $K=1$ or $K=5$. 
To evaluate the quality of the resulting classifier, each task is associated with a query set $\mathcal{Q}$, which is composed of examples of the same $N$ classes.

The core idea of meta-learning for few-shot classification is to sample few-shot tasks $\sT = (\sS, \sQ)$ from a set of ``base'' classes, of which we have ample examples per class.
Meta-learning then learns the ability of \emph{``how to build a classifier using limited data''} from these tasks.
After this \textbf{meta-training} phase, we then proceed to the \textbf{meta-testing} phase to tackle the true few-shot tasks that are composed of examples from ``novel'' classes. By default, the ``novel'' and ``base'' classes are disjoint.
It is worth noting that the total number of ``base'' (and ``novel'') classes is usually larger than $N$ (see \autoref{ss_exp_setup}). Thus, to construct an $N$-way $K$-shot task in each phase, one usually first samples $N$ classes from the corresponding set of classes, and randomly re-labels each sampled class by an index $c\in[N]$. This randomness results in the so-called mutually-exclusive tasks~\citep{yin2020meta}. 
In this paper, we will use base (novel) and meta-training (meta-testing) classes interchangeably.

\subsection{Model-agnostic meta-learning (MAML)}
\label{ss_MAML}

As introduced in~\autoref{s_intro}, MAML aims to learn the initialization of an $N$-way classifier, such that when provided with the support set $\sS$ of an $N$-way $K$-shot task, the classifier can be quickly and robustly updated to perform well on the task (\ie, classify the query set $\sQ$ well). Let us denote a classifier by $\hat{y} = h_{\vtheta}(\vx) = \argmax_{c\in[N]} \vw_c^\top f_{\vphi}(\vx)$, where $f_{\vphi}$ is the feature extractor, $\{\vw_c\}_{c=1}^N$ are the weight vectors of the classification head, and $\vtheta=\{\vphi, \vw_1, \cdots, \vw_N\}$ collects the parameters of both.
MAML evaluates $h_{\vtheta}$ on $\sS$ and uses the gradient to update $\vtheta$ into $\vtheta'$, so that $h_{\vtheta'}$ can be applied to $\sQ$. This procedure is called the \textbf{inner loop optimization}, which usually takes $M$ gradient steps.
\begin{align}
    & \vtheta' \leftarrow \vtheta\nonumber\\
    & \text{for } m\in[M] \text{ do}\label{e_inner}\\
    & \quad \quad \vtheta' = \vtheta' - \alpha\nabla_{\vtheta'}\sL(\sS, \vtheta')\nonumber
\end{align}
Here, $\sL(\sS, \vtheta') = \sum_{(\vx,y)\in\sS}\ell(h_{\vtheta'}(\vx), y)$ is the loss computed on examples of $\sS$ and $\alpha$ is the learning rate (or step size). The cross-entropy loss is commonly used for $\ell$.
As suggested in the original MAML paper~\citep{FinnAL17Model} and \citep{Antoniou2018How}, $M$ is usually set to a small integer (\eg, $\leq 5$). For ease of notation, let us denote the output $\vtheta'$ after $M$ gradient steps by $\vtheta' = \IL(\sS, \vtheta, M)$.
 
To learn the initialization $\vtheta$, MAML leverages the few-shot tasks sampled from the base classes. Let us denote by $p(\sT)$ the distribution of tasks from the base classes, where each task is a pair of support and query sets $(\sS, \sQ)$. MAML aims to minimize the following meta-training objective w.r.t. $\vtheta$:
\begin{align}
\mathop{\mathbb{E}}_{(\sS,\sQ)\sim p(\sT)} \sL(\sQ, \vtheta'_{\sS}) = 
\mathop{\mathbb{E}}_{(\sS,\sQ)\sim p(\sT)} \sL(\sQ, \IL(\sS, \vtheta, M)).
\label{eq:maml_obj}
\end{align}
Namely, MAML aims to find a shared $\vtheta$ among tasks, which, after inner loop updates using $\sS$, can lead to a small loss on the query set $\sQ$. (We add the subscript $_\sS$ to $\vtheta'$ to show that $\vtheta'_\sS$ depends on $\sS$.) 
To optimize \autoref{eq:maml_obj}, MAML applies stochastic gradient descent (SGD) but at the task level. That is, at every iteration, MAML samples a task $\sT=(\sS,\sQ)$ and computes the meta-gradient w.r.t. $\vtheta$:
\begin{align}
\nabla_{\vtheta} \sL(\sQ, \vtheta'_{\sS}) = \nabla_{\vtheta} \sL(\sQ, \IL(\sS, \vtheta, M)).
\label{eq:maml_obj_gradient}
\end{align}
In practice, one may sample a mini-batch of tasks and compute the mini-batch meta-gradient w.r.t. $\vtheta$ to optimize $\vtheta$. This SGD for $\vtheta$ is known as the \textbf{outer loop optimization} for MAML.
It is worth noting that calculating the gradient in \autoref{eq:maml_obj_gradient} can impose considerable computational and memory burdens because it involves a gradient
through a gradient (along the inner loop but in a backward order)~\citep{FinnAL17Model}. Thus in practice, it is common to apply the first-order approximation~\citep{FinnAL17Model,Nichol2018Reptile}, \ie, $\nabla_{\vtheta} \sL(\sQ, \vtheta'_{\sS})\approx \nabla_{\vtheta'_{\sS}} \sL(\sQ, \vtheta'_{\sS})$. 

\textbf{For additional related work on meta-learning and few-shot learning, please see \autoref{suppl_s_related}.}

\subsection{Experimental setup}
\label{ss_exp_setup}
As our paper is heavily driven by empirical observations, we first introduce the three main datasets we experiment on, the neural network architectures we use, and the implementation details.

\textbf{Dataset.} We work on \textbf{\emph{Mini}ImageNet} \citep{VinyalsBLKW16Matching}, \textbf{\emph{Tiered}ImageNet}~\citep{ren2018meta}, and \textbf{CUB} datasets~\citep{WahCUB_200_2011}. {\it Mini}ImageNet contains $100$ semantic classes; each has 600 images. Following~\citep{Sachin2017}, the $100$ classes are split into meta-training/validation/testing sets with 64/16/20 (non-overlapped) classes, respectively. That is, there are $64$ base classes and $20$ novel classes; the other $16$ classes are used for hyper-parameter tuning.
{\it Tiered}ImageNet~\citep{ren2018meta} has $608$ semantic classes, which are split into the three sets with 351/97/160 classes, respectively. On average, each class has $\sim 1,300$ images.
{CUB}~\citep{WahCUB_200_2011} has $200$ classes, which are split into the three sets with 200/50/50 classes
following~\citep{ye2020fewshot}.
All images are resized to $84\times84$, following~\citep{Lee2019Meta,ye2020fewshot}.

\textbf{Training and evaluation.} During meta-training, meta-validation, and meta-testing, we sample $N$-way $K$-shot tasks from the corresponding classes and images. We follow the literature~\citep{SnellSZ17Prototypical,VinyalsBLKW16Matching} to study the five-way one-shot and five-way five-shot tasks. As mentioned in \autoref{ss_problem_def}, every time we sample five distinct classes, we randomly assign each of them an index $c\in[N]$. During meta-testing, we follow the evaluation protocol in \citep{Zhang2020Deep,Rusu2018LEO,ye2020fewshot} to sample $10,000$ tasks. In each task, the query set contains $15$ images per class. We report the mean accuracy (in \%) and the $95\%$ confidence interval.

\textbf{Model architecture.} 
We follow~\citep{Lee2019Meta} to use a  ResNet-12~\citep{he2016deep} architecture for $f_{\vphi}$ (cf. \autoref{ss_MAML}), which has wider widths and Dropblock modules~\citep{Ghiasi2018Drop}. We note that many recent few-shot learning algorithms use this backbone. We also follow the original MAML~\citep{FinnAL17Model} to use a 4-layer convolutional network (ConvNet)~\citep{VinyalsBLKW16Matching}.

\textbf{Implementation details.} 
Throughout the paper, for simplicity and consistency, we use
\begin{itemize} [nosep,topsep=0pt,parsep=0pt,partopsep=0pt, leftmargin=*] 
    \item the first-order approximation for calculating the meta-gradient in the outer loop;
    \item the same number of gradient steps in the inner loop during meta-training and meta-testing;
    \item the weights pre-trained on the entire meta-training set to initialize $\vphi$, following the recent practice~\citep{ye2020fewshot,Rusu2018LEO,qiao2018few}. \emph{We note that in meta-training we still optimize this pre-trained $\vphi$ in the ``outer'' loop to search for a better initialization for MAML.}
\end{itemize}
MAML has several hyper-parameters and we select them on the meta-validation set.
Specifically, for the outer loop, we learn with at most $10,000$ tasks: we group every $100$ tasks into an epoch. 
We apply SGD with momentum $0.9$ and weight decay $0.0005$. We start with an outer loop learning rate $0.002$ for ConvNet and $0.001$ for ResNet-12, which are decayed by $0.5$ and $0.1$ after every $20$ epochs for ConvNet and ResNet-12, respectively.
For the inner loop, we have to set the number of gradient steps $M$ and the learning rate $\alpha$ (cf. \autoref{e_inner}).
We provide more details in the next section.

%% file: approach-1.tex

\section{MAML Needs A Large Number of Inner Loop Gradient Steps}
\label{s_approach_1}

\begin{figure}[t]
\centering
\minipage{0.26\textwidth}
\centering
{\includegraphics[width=1\textwidth]{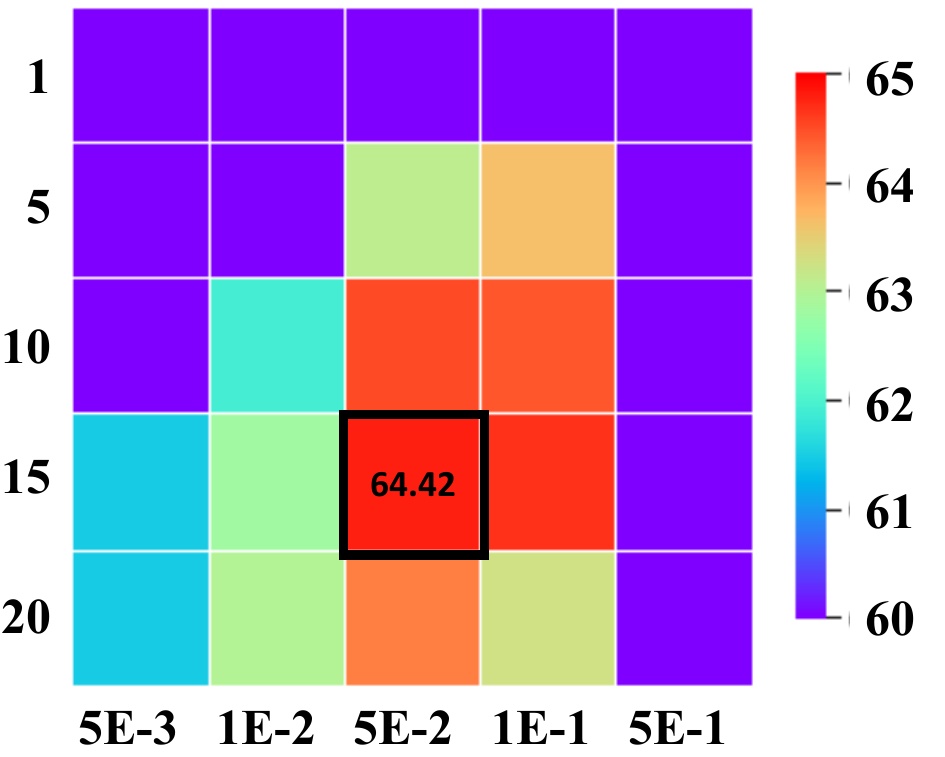}}\\
\mbox{\small {\it Mini}ImageNet, ResNet}
\endminipage
\hfill
\minipage{0.26\textwidth}
\centering
{\includegraphics[width=1\textwidth]{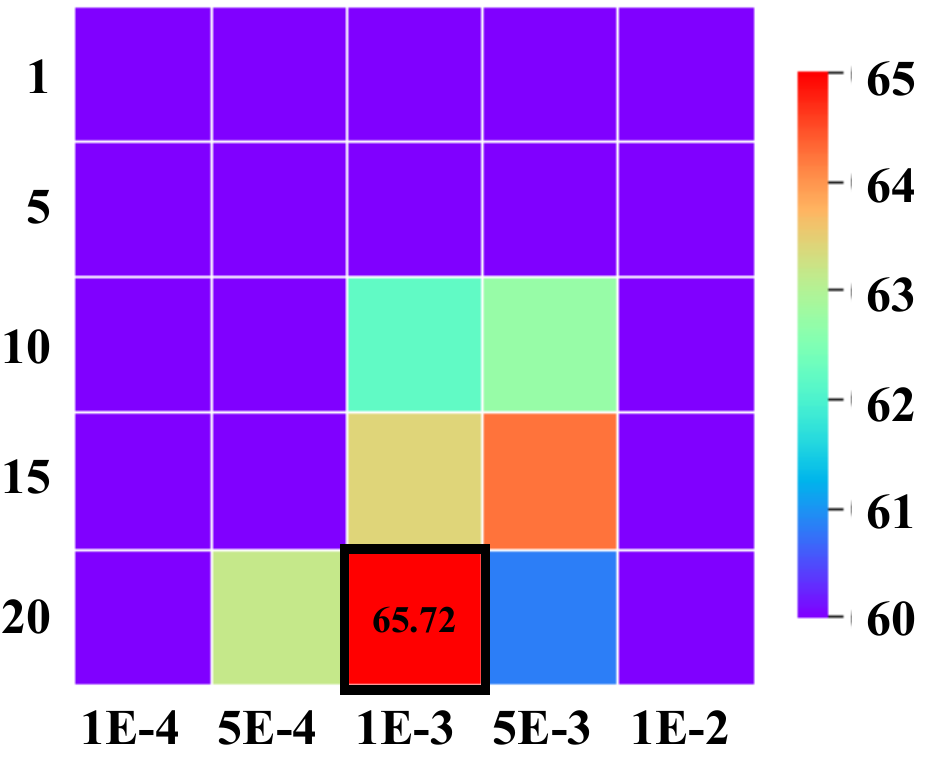}}\\
\mbox{\small {\it Tiered}ImageNet, ResNet}
\endminipage
\hfill
\minipage{0.26\textwidth}
\centering
{\includegraphics[width=1\textwidth]{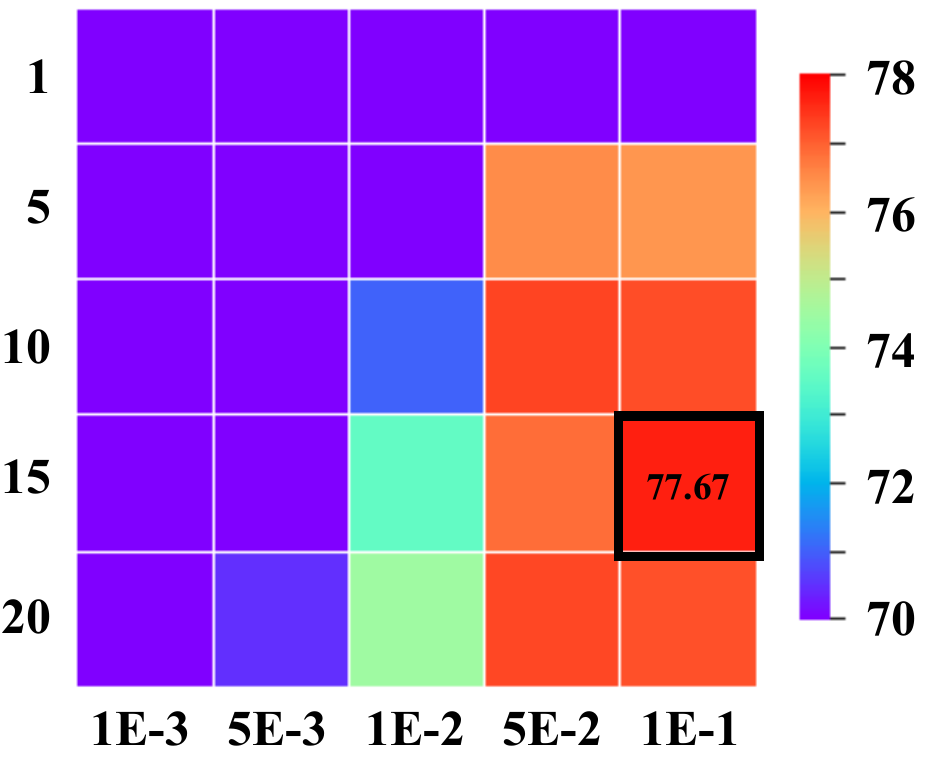}}\\
\mbox{\small CUB, ResNet}
\endminipage
\\ \vskip 5pt
\minipage{0.26\textwidth}
\centering
{\includegraphics[width=1\textwidth]{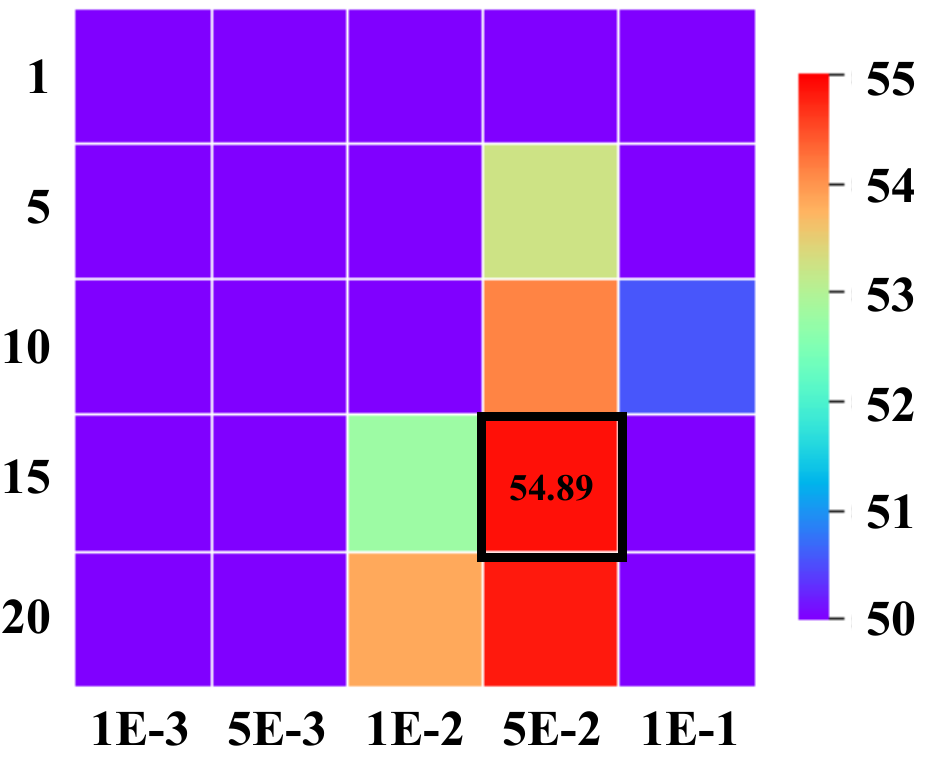}}\\
\mbox{\small {\it Mini}ImageNet, ConvNet}
\endminipage
\hfill
\minipage{0.26\textwidth}
\centering
{\includegraphics[width=1\textwidth]{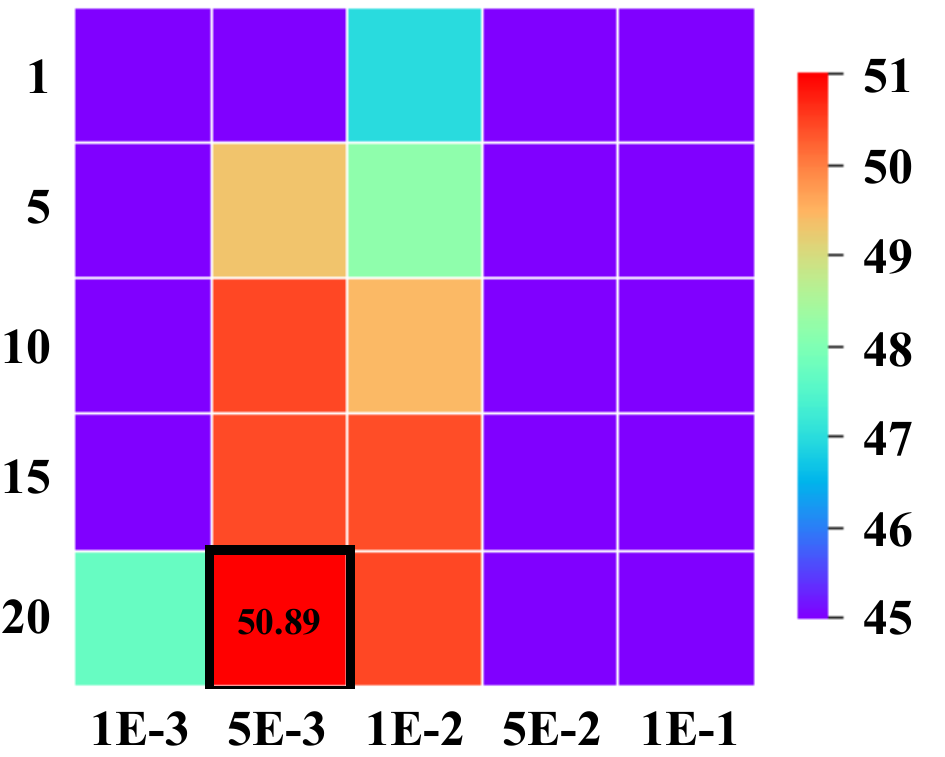}}\\
\mbox{\small {\it Tiered}ImageNet, ConvNet}
\endminipage
\hfill
\minipage{0.26\textwidth}
\centering
{\includegraphics[width=1\textwidth]{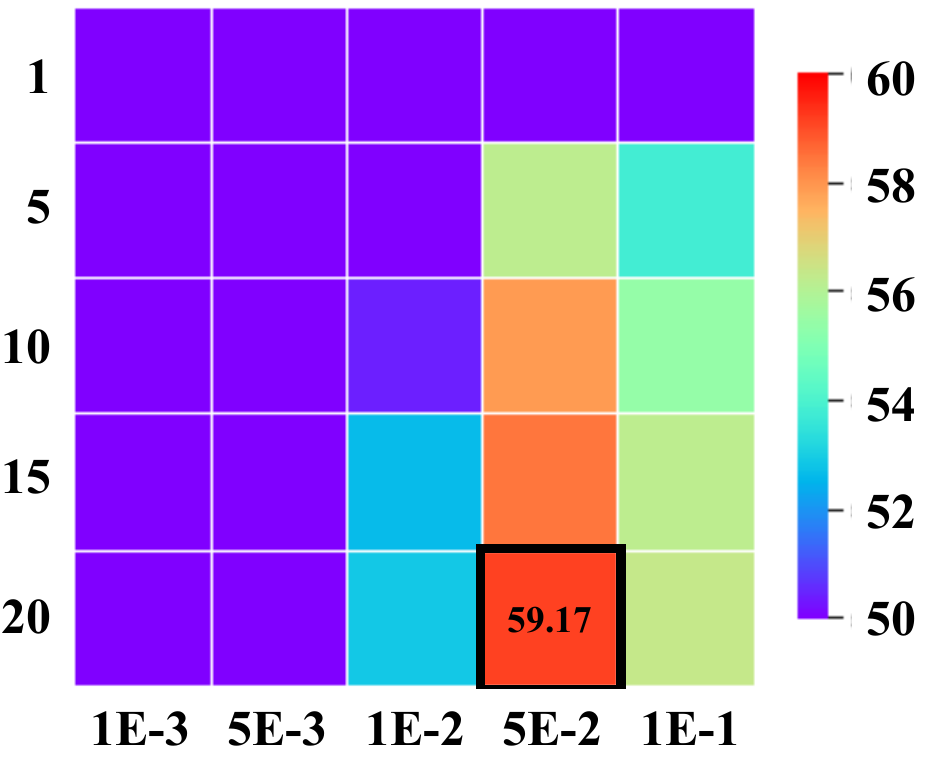}}\\
\mbox{\small CUB, ConvNet}
\endminipage
\vskip -7pt
\caption{\small Heat maps of MAML's five-way one-shot accuracy on {\it Mini}ImageNet, {\it Tiered}ImageNet, and CUB w.r.t. the inner loop learning rate $\alpha$ (x-axis) and the number of inner loop updates $M$ (y-axis).  
For each heat map, \textbf{we set accuracy below a threshold to a fixed value for clarity}; we denote the best accuracy by a black box.}\label{fig:heatmap}
\vskip -7pt
\end{figure}

We find that for MAML's inner loop, the number of gradient updates $M$ (cf. \autoref{e_inner}) is usually selected within a small range close to $1$, \eg, $M\in[1, 5]$~\citep{Antoniou2018How}. At first glance, this makes sense according to the motivation of MAML~\citep{FinnAL17Model} --- with a small
number of gradient steps, the resulting model will have a good generalization performance.

In our experiment, we however observe that it is crucial to explore a larger $M$\footnote{We reiterate that for simplicity and consistency we apply the same $M$ in meta-training and meta-testing.}. Specifically, we consider $M\in[1, 20]$ along with $\alpha\in[10^{-4}, 10^{0}]$. We plot the meta-testing accuracy of five-way one-shot tasks on the three datasets in \autoref{fig:heatmap}\footnote{We tune hyper-parameters on the meta-validation set and find that the accuracy there reflects the meta-testing accuracy well. We show the meta-testing accuracy here simply for a direct comparison to results in the literature.}, using both ResNet and ConvNet backbones. We find that MAML achieves higher and much more stable accuracy (w.r.t. the learning rate) when $M$ is larger, \eg, larger than $10$. Specifically, for {\emph{Mini}ImageNet} with ResNet, the highest accuracy $64.42\%$ is obtained with $M=15$, higher than $62.90\%$ with $M=5$; for \emph{Tiered}ImageNet with ResNet, the highest accuracy $65.72\%$ is obtained with $M=15$, higher than $59.08\%$ with $M=5$. As will be seen in \autoref{s_benchmark}, these results with a larger $M$ are already close to the state-of-the-art performance.

\begin{figure}[t]
\centering
\minipage{0.16\textwidth}
\centering
{\includegraphics[width=1\textwidth]{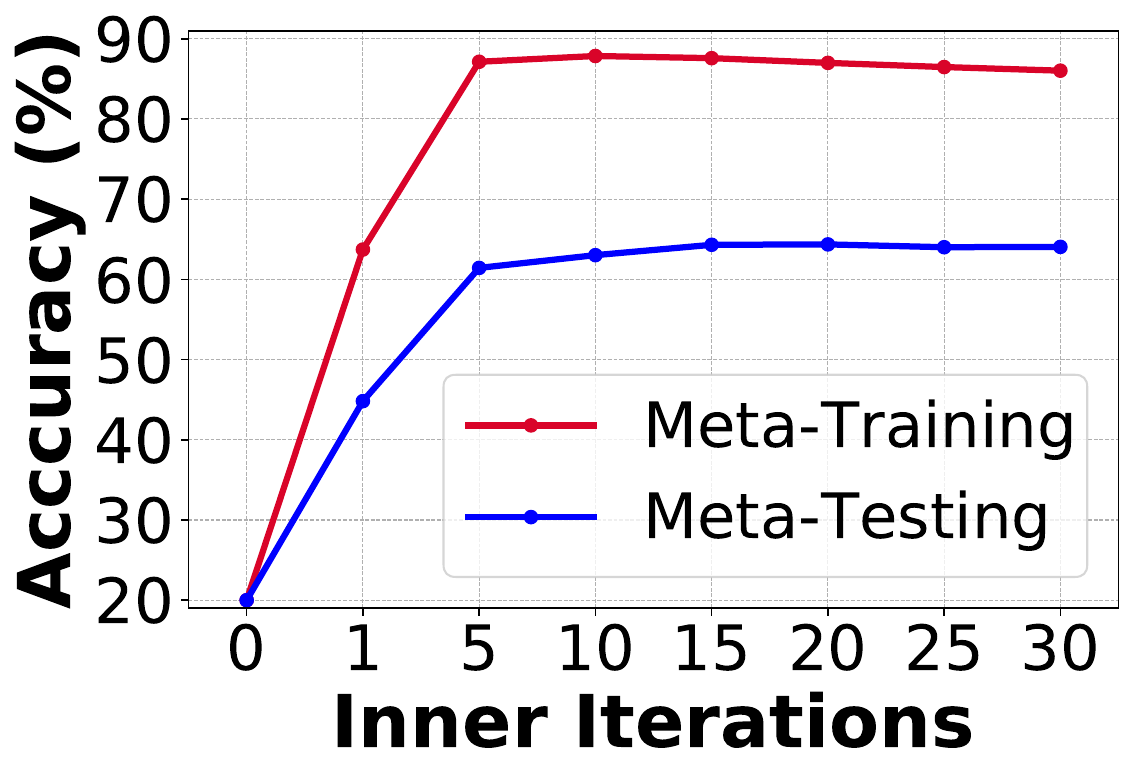}}\\
\mbox{\small {\it Mini}ImageNet, R}
\endminipage
\hfill
\minipage{0.16\textwidth}
\centering
{\includegraphics[width=1\textwidth]{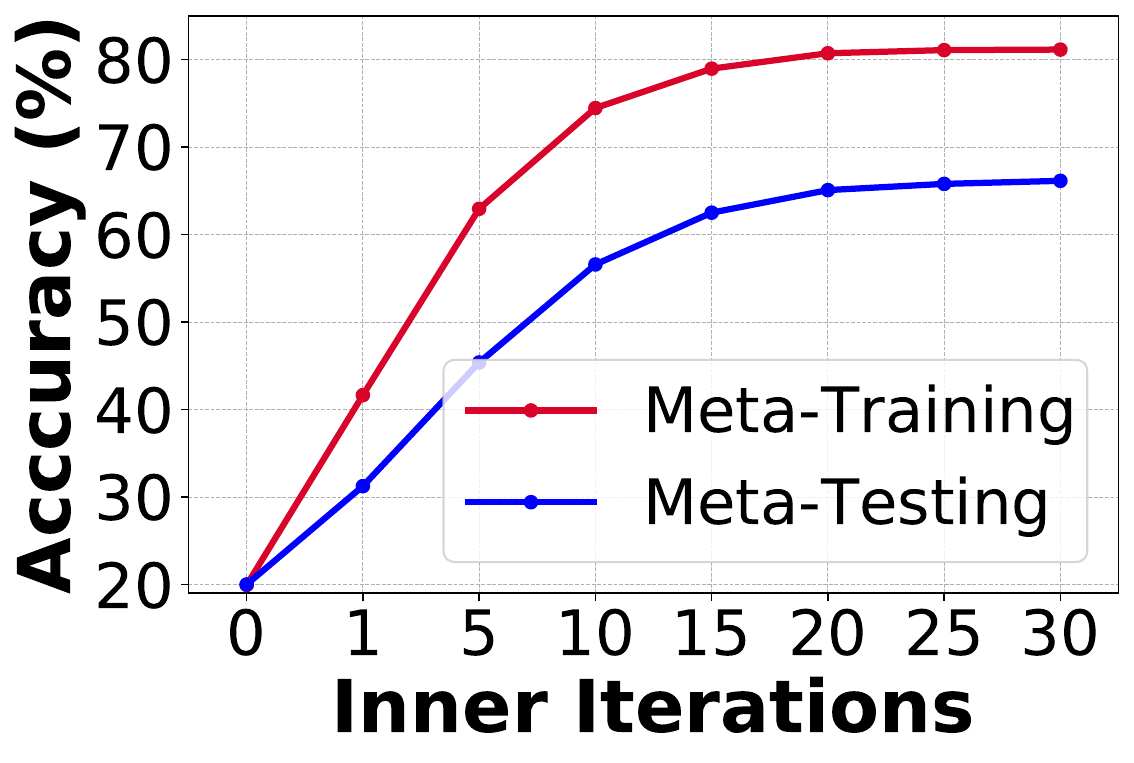}}\\
\mbox{\small {\it Tiered}ImageNet, R}
\endminipage
\hfill
\minipage{0.16\textwidth}
\centering
{\includegraphics[width=1\textwidth]{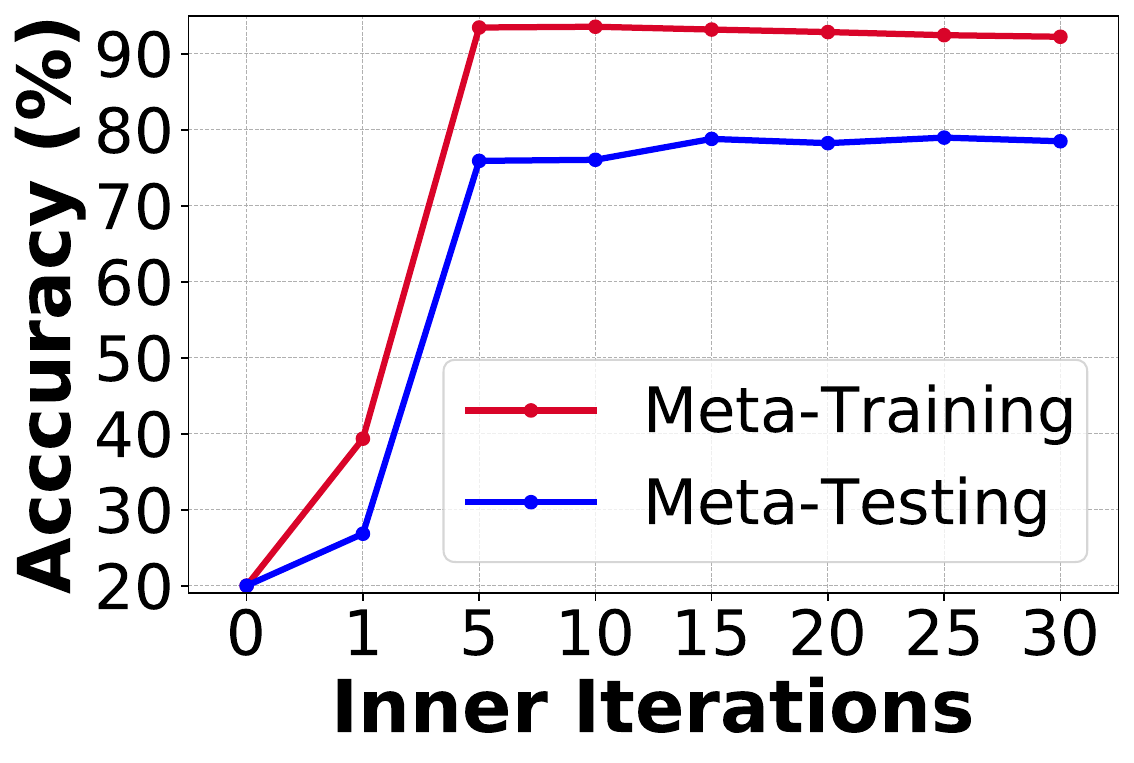}}\\
\mbox{\small CUB, R}
\endminipage
\hfill
\minipage{0.16\textwidth}
\centering
{\includegraphics[width=1\textwidth]{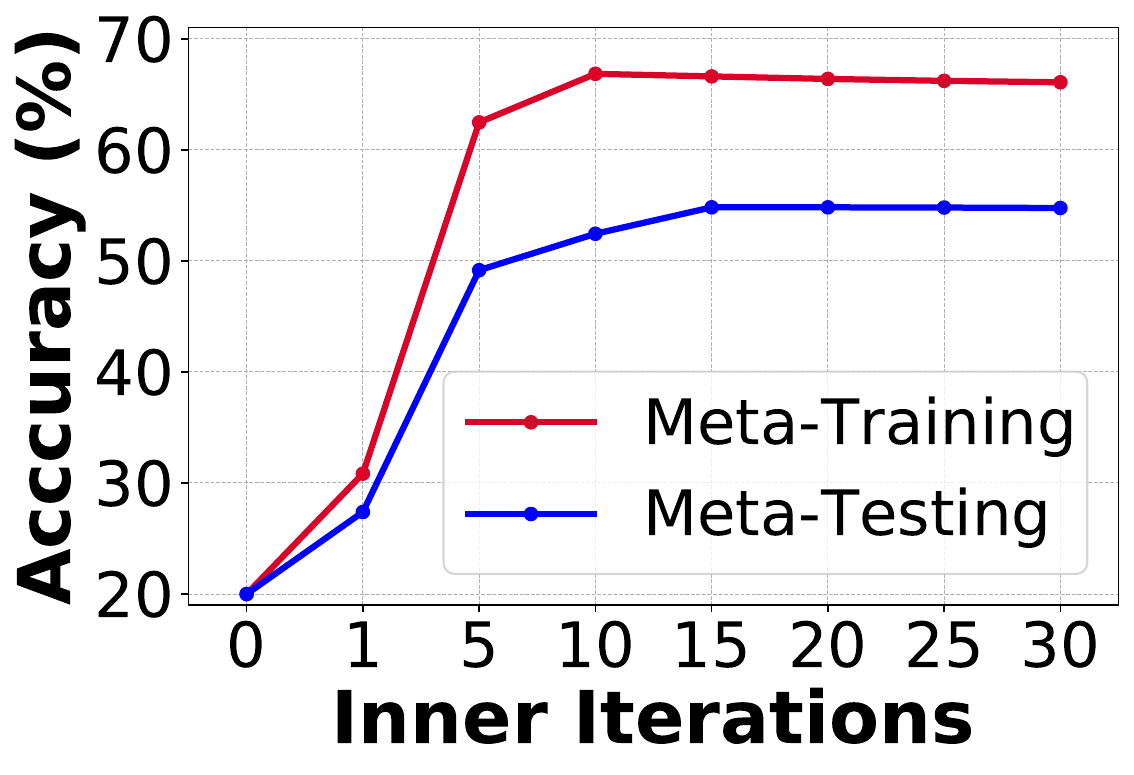}}\\
\mbox{\small {\it Mini}ImageNet, C}
\endminipage
\hfill
\minipage{0.16\textwidth}
\centering
{\includegraphics[width=1\textwidth]{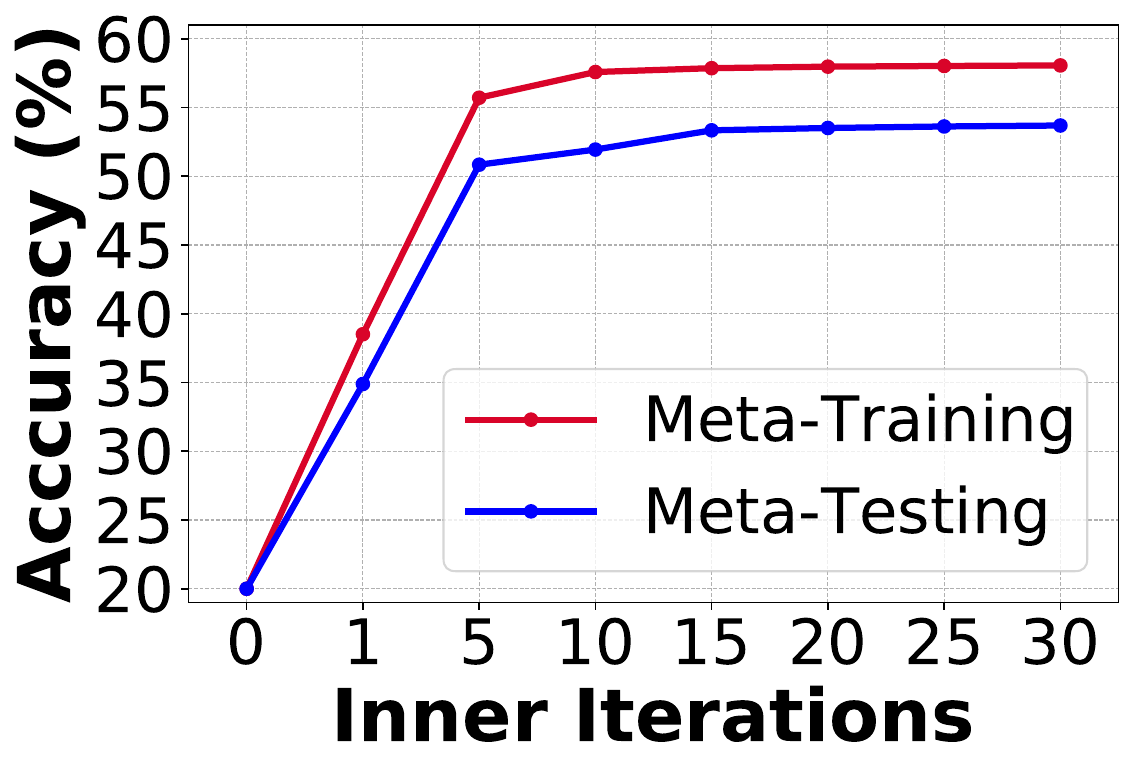}}\\
\mbox{\small {\it Tiered}ImageNet, C}
\endminipage
\hfill
\minipage{0.16\textwidth}
\centering
{\includegraphics[width=1\textwidth]{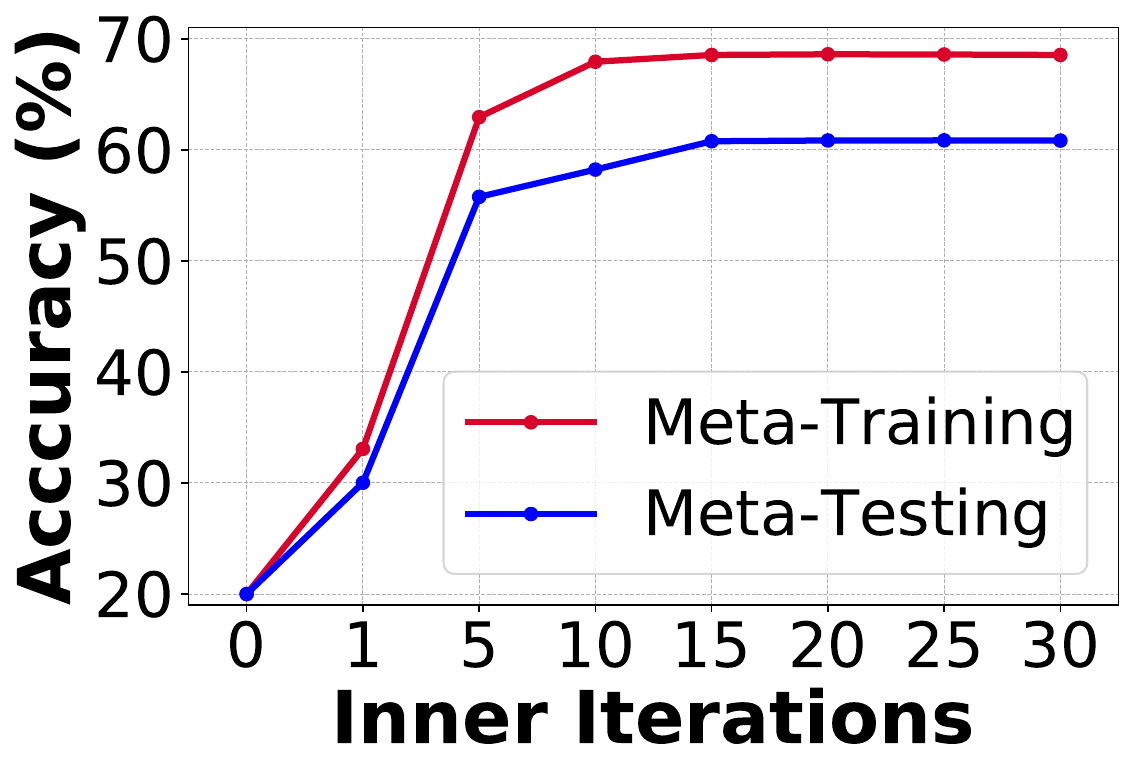}}\\
\mbox{\small CUB, C}
\endminipage
\vskip -5pt
\caption{\small We plot the change of the five-way one-shot classification accuracy (on the query set), averaged over 10,000 tasks sampled from either the meta-training (red) or meta-testing classes (blue), along with the process of inner loop updates, using the best model initialization learned by MAML. C: ConvNet; R: ResNet.}\label{fig:multistep}
\vskip -10pt
\end{figure}

To analyze why MAML needs a large $M$, we plot the change of classification accuracy along with the inner loop updates in \autoref{fig:multistep}. Specifically, we first perform meta-training using the $M$ value selected by meta-validation, for each pair of dataset and backbone.  
We then analyze the learned model initialization on few-shot tasks sampled from the meta-training and meta-testing classes, by performing $0\sim30$ inner loop updates using the support set and reporting the accuracy on the query set. \textbf{We conduct the same experiments for five-way five-shot tasks in \autoref{suppl_s_CUB}.}

We have two observations. First, in general, the more inner loop updates we perform for a few-shot task, 
the higher the accuracy is, no matter if it is a meta-training or meta-testing task. 
This trend aligns with the few-shot regression study in \citep{FinnAL17Model}.
Second, before any inner loop update, the learned initialization $\vtheta$ has a $\sim20\%$ accuracy on average, \ie, the accuracy by random classification. 
Interestingly, this is the case not only for meta-testing tasks but also for meta-training tasks, even though the learned initialization does contain the classification head $\{\vw_c\}_{c=1}^N$. \emph{This explains why a larger number of inner loop gradient steps is needed: the learned initialized model has to be updated from performing random predictions to achieving a much higher classification accuracy}.

We attribute the second observation to the \emph{random} class label assignments in creating few-shot tasks (cf.~\autoref{ss_problem_def} and \autoref{ss_exp_setup}), which make the created tasks mutually-exclusive~\citep{yin2020meta} --- \ie, a single model cannot solve them all at once before inner loop optimization. 
Concretely, for a few-shot task of a specific set of $N$ semantic classes (\eg, $\{$``dog'', ``cat'', $\cdots$,``bird''$\}$), such a randomness can turn it into different tasks from MAML's perspective.
For instance, the class ``dog'' may be assigned to $c=1$ and paired with $\vw_1$ at the current task, but to $c=2$ and paired with $\vw_2$ when it is sampled again. \emph{For a five-way task, the same set of five semantic classes can be assigned to $\{1,\cdots,5\}$ via $120$ (\ie, $5!$) different permutations.} 
As a result, if we directly apply the learned initialization of MAML without inner loop updates, the accuracy on few-shot tasks of the same set of semantic classes (but in different permutations) can cancels each other out. (Please see \autoref{supp_ss_math} for details.) Besides, since the randomness occurs also in meta-training, each $\vw_c$ will be discouraged to learn specific knowledge towards any semantic class~\citep{rajendran2020meta,yao2021improving,yin2020meta}, hence producing an accuracy at the chance level even on meta-training tasks.

\noindent\textbf{Related work.} Several following-up works of MAML, \eg, \citep{Hsu2018Unsupervised}, use different numbers of inner loop steps in meta-training (\eg, $M=1\sim5$) and meta-testing (\eg, $M=20\sim50$). 
We make $M$ equal in the two phases for consistency, and provide a detailed analysis of why MAML needs a large $M$ value. On large-scale tasks beyond few-shot learning, \citet{shin2021large} also found the necessity of a larger number of inner loop steps, but from a different perspective than ours.

%% file: approach-2.tex
 
\section{MAML is Sensitive to The Label Permutations in Meta-Testing}
\label{s_approach_2}

The randomness in class label assignments raises an interesting question: \emph{do different permutations result in different meta-testing accuracy after inner loop updates?} More specifically, if $\{\vw_c\}_{c=1}^N$ are paired with the $N$ classes differently, will the updated model after the inner loop perform differently?

\begin{figure}[t]
\centering
\begin{minipage}{0.22\linewidth}
\centering
\includegraphics[width=\linewidth]{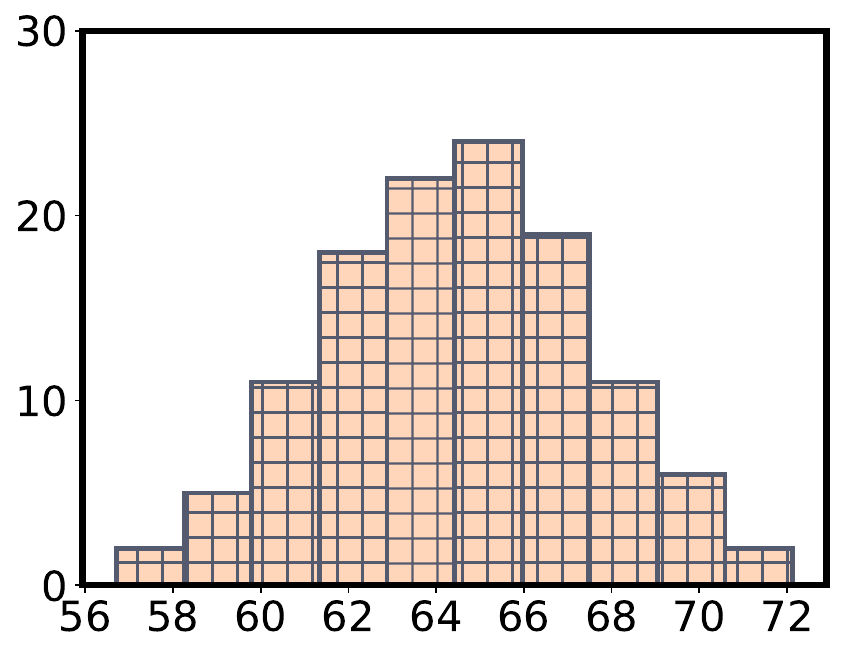}
\vskip-5pt
\centering\mbox{\scriptsize(a) {\it Mini}ImageNet, 1-Shot}
\end{minipage}
\hfill
\begin{minipage}{0.22\linewidth}
\centering
\includegraphics[width=\linewidth]{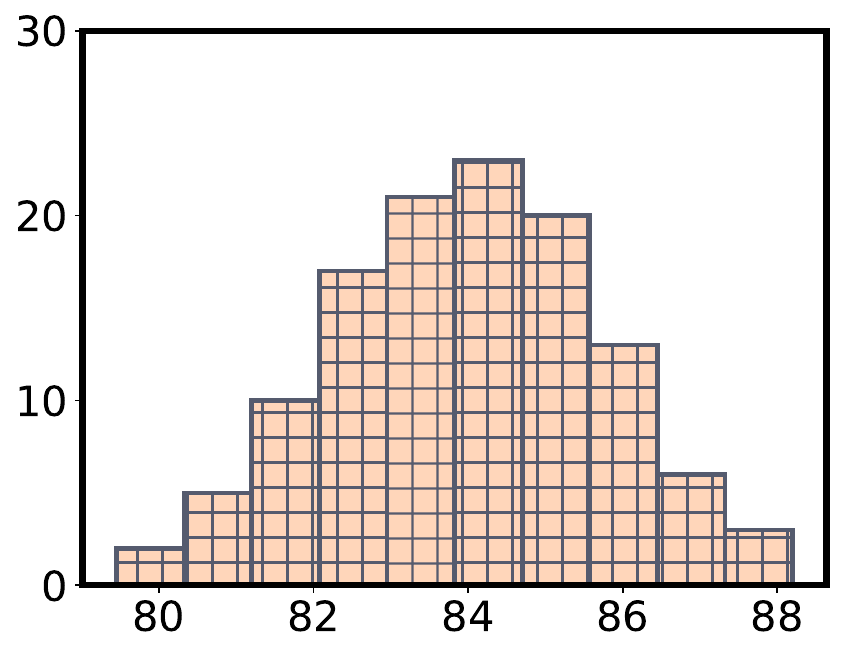}
\vskip-5pt
\centering\mbox{\scriptsize(b) {\it Mini}ImageNet, 5-Shot}
\end{minipage}
\hfill
\begin{minipage}{0.22\linewidth}
\centering
\includegraphics[width=\linewidth]{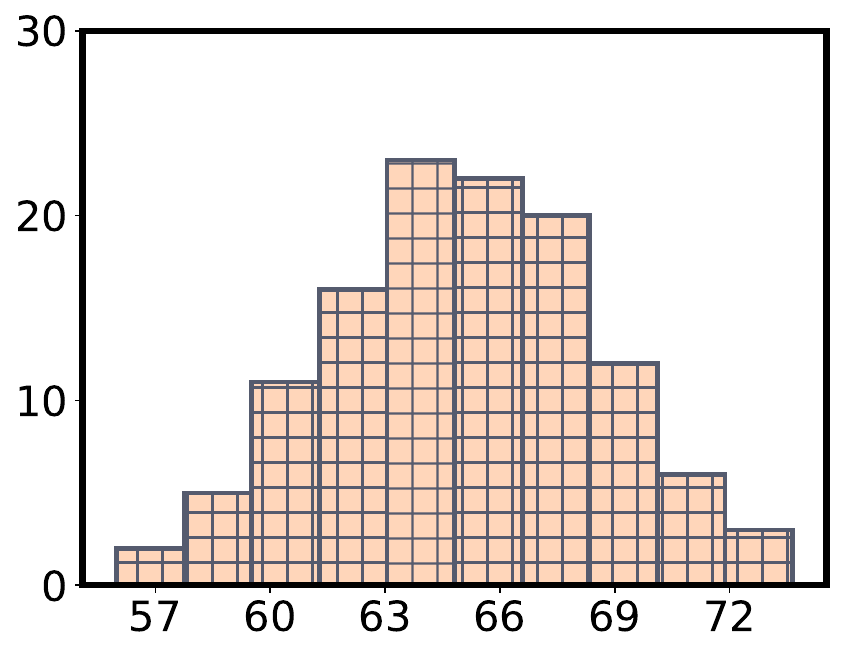}
\vskip-5pt
\centering\mbox{\scriptsize(c) {\it Tiered}ImageNet, 1-Shot}
\end{minipage}
\hfill
\begin{minipage}{0.22\linewidth}
\centering
\includegraphics[width=\linewidth]{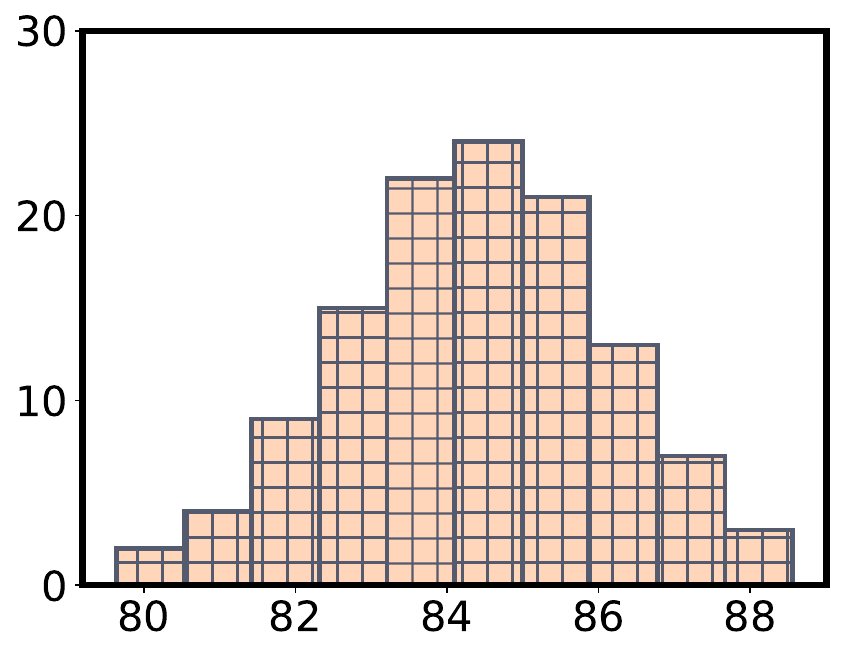}
\vskip-5pt
\centering\mbox{\scriptsize (d) {\it Tiered}ImageNet, 5-Shot}
\end{minipage}
\vskip -4pt
\caption{\small The histogram of the $120$ meta-testing accuracy (averaged over $2,000$ tasks), each corresponds to a specific position in the sorted list of each task's accuracy among $120$ permutations.
The x-axis corresponds to accuracy (range); the y-axis corresponds to counts. The backbone is ResNet-12.}\label{fig:permutation} 
\vskip -6pt
\end{figure}

To answer this question, we conduct a detailed experiment: Algorithm \ref{a_MAML} summarizes the procedure. We focus on \textbf{five-way one/five-shot} tasks on {\emph{Mini}ImageNet} and \emph{Tiered}ImageNet, using the ResNet backbone. For each task type and dataset combination, we first meta-train the model initialization using MAML, and then evaluate the learned initialized on $2,000$ meta-testing tasks. For each task, there are $120$ permutations; each permutation, after the inner loop, would likely lead to a different model and query set accuracy. \emph{We sort the $120$ accuracy for each task, and take average for each position in the sorted list over $2,000$ tasks.} This results in $120$ averaged accuracy, each for a specific position in the sorted list. Specifically, the highest accuracy corresponds to the case that each task cherry-picks its best permutation according to the query set accuracy after inner loop optimization.

\setlength{\textfloatsep}{8pt}
\begin{algorithm}[t]
\small
\SetAlgoLined
\caption{\small Evaluation of the effect of class label permutations on meta-testing tasks.}
\textbf{Given} the learned initialization $\vtheta$ by MAML\\
\For{$t\in\{1,\cdots,2000\}$}{
\textbf{Sample} a meta-testing task
$\mathcal{T} = (\mathcal{S}, \mathcal{Q})$ and initialize an accuracy vector $\va_t\in\R^{120}$
\\
\For{$p\in\{1,\cdots,120\}$}{
\textbf{Shuffle} the class labels with a specific permutation $\pi: [N]\mapsto[N]$; \ie, $(\mathcal{S}, \mathcal{Q})$ becomes $(\mathcal{S}_{\pi}, \mathcal{Q}_{\pi})$\\
\textbf{Update} $\vtheta$ to get $\vtheta' = \IL(\sS_{\pi}, \vtheta, M)$, 
evaluate $\vtheta'$ on $\mathcal{Q}_{\pi}$, and record the accuracy into $\va_t[p]$\\
}
\textbf{Sort} values in the accuracy vector $\va_t$ in the descending order and get $\va_t^\text{(sort)}$
}
\textbf{Average} over $2000$ tasks by $\frac{1}{2000}\sum_t \va_t^\text{(sort)}$
\label{a_MAML}
\end{algorithm}

We show the histogram of the $120$ average meta-testing accuracy in~\autoref{fig:permutation}. There exists a huge variance. Specifically, the best permutation can be $15\%/8\%$ higher than the worst in one/five-shot tasks. The best permutation is also much higher than vanilla MAML's results (from \autoref{s_approach_1}), which are
64.42\%/83.44\%/65.72\%/84.37\%, corresponding to the four sub-figures from left to right in~\autoref{fig:permutation}. What's more, the best permutation can easily achieve state-of-the-art accuracy (see~\autoref{s_benchmark}).

Of course, so far we find the best permutation via cherry-picking --- by looking at the meta-testing accuracy --- so it is like an upper bound. However, if we can find the best permutation without looking at the query sets' labels
or make MAML permutation-invariant, we may practically improve MAML.

\noindent\textbf{Related work.} We note that we investigate the effect of the permutations of class label assignments differently from~\citep{rajendran2020meta,yao2021improving,yin2020meta}. There, the permutations lead to mutually-exclusive meta-training tasks, which help prevent MAML from over-fitting. Here, we look at the meta-testing tasks, and permutations result in a huge variance of meta-testing accuracy.

%% file: approach-3.tex

\section{Making MAML Permutation-Invariant during Meta-Testing}\label{s_pi_test}

We study approaches that can 
make MAML permutation-invariant during meta-testing.
That is, we take the same learned initialization $\vtheta$ as in \autoref{s_approach_2} without changing the meta-training phase.

We first investigate \textbf{searching for the best permutation for each task}. As we cannot access query sets' labels, we use the support sets' data as a proxy. We choose the best permutation according to which permutation, either before or after inner loop updates (less practical), leads to the highest accuracy or smallest loss on the support set.
\autoref{tab:permutation_select} summarizes the results: none of them leads to consistent gains. We hypothesize two reasons. First, due to mutually-exclusive tasks in meta-training, the learned $\vtheta$ by MAML would produce chance-level predictions before updates (see \autoref{suppl_s_intro}). Second, after updates, the support set accuracy quickly goes to $100\%$ and is thus not informative.

In stead of choosing one from many, we further explore \textbf{taking ensemble}~\citep{breiman1996bagging,zhou2012ensemble,dietterich2000ensemble} \textbf{over the predictions made by updated models of different permutations.} We note that this makes MAML permutation-invariant but inevitably needs more computations.
To make the ensemble process clear, 
we permute the weight vectors in $\{\vw_c\}_{c=1}^N$ rather than the class label assignments in a task: the two methods are equivalent but the former is easier for aggregating predictions.
We study two variants: (a) full permutations (\ie, $120$ of them in five-way tasks), which is intractable for larger $N$; (b) rotated permutations, which rotates the index $c$ in $\{\vw_c\}_{c=1}^N$\footnote{That is, we consider re-assign $\vw_c$ to $\vw_{(c+\gamma \mod N) +1}$, where $\gamma\in[N]$.}, leading to $N$ permutations. \autoref{tab:ensemble} shows the results --- ensemble consistently improves MAML. Even with the rotated version that has much fewer permutations than the full version, the gains are comparable. 

\emph{We emphasize that our focus here is to make MAML permutation-invariant in meta-testing, not to explore and compare all potential ways of performing ensemble on MAML.}

\begin{table}[t]
  \centering
  \small 
  \caption{\small The meta-testing accuracy over 2,000 tasks given different permutation selection strategies.}
  \vskip-10pt
    \begin{tabular}{l|cccc}
    \addlinespace
    \toprule
          & \multicolumn{ 2}{c}{{\it Mini}ImageNet} & \multicolumn{ 2}{c}{{\it Tiered}ImageNet} \\
    \midrule
    Select the permutation by & 1-Shot & 5-Shot & 1-Shot & 5-Shot \\
    \midrule
    None  & 64.42 $\pm$ 0.20 & 83.44 $\pm$ 0.13 & 65.72 $\pm$ 0.20 & 84.37 $\pm$ 0.16 \\
    Initial Support Acc & 64.42 $\pm$ 0.20 & 83.95 $\pm$ 0.13 & 65.06 $\pm$ 0.20 & 84.32 $\pm$ 0.16 \\
    Initial Support Loss & 64.42 $\pm$ 0.20 & 83.91 $\pm$ 0.13 & 65.42 $\pm$ 0.20 & 84.23 $\pm$ 0.16 \\
    Updated Support Acc & 64.42 $\pm$ 0.20 & 83.95 $\pm$ 0.13 & 65.01 $\pm$ 0.20 & 84.37 $\pm$ 0.16 \\
    Updated Support Loss & 64.67 $\pm$ 0.20 & 84.05 $\pm$ 0.13 & 65.43 $\pm$ 0.20 & 84.22 $\pm$ 0.16 \\
    \bottomrule
    \end{tabular}
  \label{tab:permutation_select}
  \vskip -5pt
\end{table}

\begin{table}[] 
\centering
\begin{minipage}[c]{0.57\textwidth}
\centering
\tabcolsep 1.5pt
\small
\caption{\small Ensemble over updated models of different permutations. (The confidence interval is omitted due to space limit.)}
\vspace{-10pt}
\begin{tabular}{c|ccc||c|ccc}
    \addlinespace
    \toprule
    {\it Mini} & Vanilla & Full & Rotated & {\it Tiered} & Vanilla & Full & Rotated \\
          \midrule
    1-Shot & 64.42 & 65.50 & 65.37 & 1-Shot & 65.72 & 66.68 & 66.63 \\
    5-Shot & 83.44 & 84.43 & 84.40 & 5-Shot & 84.37 & 84.83 & 84.81 \\
    \bottomrule
    \end{tabular}\label{tab:ensemble}
\end{minipage}
\hfill
\begin{minipage}[c]{0.4\textwidth}
\centering
\small
\tabcolsep 4pt
\caption{\small  We average the top-layer classifiers and expand it to $N$-way during meta-testing.}
\vspace{-10pt}
\begin{tabular}{c|cc}
\addlinespace
\toprule
      & {\it Mini}ImageNet & {\it Tiered}ImageNet \\
\midrule
1-Shot &  64.40 $\pm$ 0.21 & 66.24 $\pm$ 0.24 \\
5-Shot & 84.24 $\pm$ 0.13 & 84.52 $\pm$ 0.16 \\
\bottomrule
\end{tabular}\label{tab:test_avg}
\end{minipage}
\vskip 5pt
\end{table}

We further explore an efficient approach to make MAML permutation-invariant, which is to \textbf{manipulate the learned initialization of $\{\vw_c\}_{c=1}^C$}. Concretely, MAML is sensitive to the permutations in class assignments because $\vw_c\neq\vw_{c'}$ for $c\neq c'$. One method to overcome this is to make $\vw_c=\vw_{c'}$ during meta-testing. Here, we investigate re-initializing each $\vw_c$ by their average in meta-testing:  $\vw_c\leftarrow\frac{1}{N}\sum_{c'=1}^{N} \vw_{c'}$. 
By doing so, no matter which permutation we perform, the model after inner loop optimization will be the same and lead to the same query set accuracy.
\autoref{tab:test_avg} summarizes the results; this approach improves vanilla MAML (see \autoref{tab:ensemble}) in three of the four cases.

At first glance, this approach may not make sense since the resulting $\{\vw_c\}_{c=1}^C$, before inner loop updates, are identical and simply make random predictions. However, please note that even the original $\{\vw_c\}_{c=1}^C$ have an averaged chance accuracy (cf. \autoref{fig:multistep}). In~\autoref{suppl_s_dropout}, we provide an explanation of this approach by drawing an analogy to dropout~\citep{srivastava2014dropout}. In short, in meta-training, we receive a task with an arbitrary permutation, which can be seen as drawing a permutation at random for the task. In meta-testing, we then take expectation over permutations, which essentially lead to the averaged $\vw_c$. 

%% file: approach-4.tex

\section{\ourmethod: Learning a Single Weight Vector}
\label{s_benchmark}

The experimental results in \autoref{s_pi_test} are promising: by making MAML permutation-invariant in meta-testing, we can potentially improve vanilla MAML. While ensemble inevitably increases the computational burdens, the method by manipulating the learned initialization of $\{\vw_c\}_{c=1}^N$ keeps the same run time as vanilla MAML. In this section, we investigate the latter approach further. We ask:
\begin{displayquote} 
 If we directly learn \emph{a single weight vector $\vw$} to initialize $\{\vw_c\}_{c=1}^N$ in meta-training, making the inner loop optimization in meta-training and meta-testing \emph{consistent} and both \emph{permutation-invariant}, can we further improve the accuracy?  \end{displayquote}

Concretely, we redefine the learnable meta-parameters $\vtheta$ of MAML, which become $\vtheta=\{\vphi, \vw\}$, where $\vphi$ is for the feature extractor. We name this method \textbf{\ourmethod}, as we meta-train only a single weight vector $\vw$ in the classification head. The inner loop optimization and outer loop optimization of \ourmethod very much follow MAML, with some slight changes. 
\begin{itemize} [nosep,topsep=0pt,parsep=0pt,partopsep=0pt, leftmargin=*] 
\item \textbf{Inner loop optimization:} At the beginning, $\vw$ is duplicated into $\{\vw_c = \vw\}_{c=1}^N$. That is, we use $\vw$ to initialize every $\vw_c$, $\forall c\in[N]$.
These $\{\vw_c = \vw\}_{c=1}^N$ then undergo the same inner loop optimization process as vanilla MAML (cf. \autoref{e_inner}).
\item \textbf{Outer loop optimization:} Let us denote the updated model by $\vtheta'=\{\vphi', \vw_1', \cdots, \vw'_N\}$. To perform the outer loop optimization for $\vw$ in meta-training, we collect the gradients derived from the query set $\sQ$. Let us denote by $\nabla_{\vw_c}\sL(\sQ,\vtheta')$ the gradient w.r.t. the initial $\vw_c$ (cf. \autoref{ss_MAML}). Since $\vw_c$ is duplicated from $\vw$, we obtain the gradient w.r.t. $\vw$ by $\sum_{c\in[N]}\nabla_{\vw_c}\sL(\sQ,\vtheta')$. 
\end{itemize}

\begin{table}[tbp]
  \centering
  \small
  \caption{\small 5-Way 1-Shot and 5-Shot classification accuracy and 95\% confidence interval on {\it Mini}ImageNet and {\it Tiered}ImageNet (over 10,000 tasks), using ResNet-12 as the backbone. $\dagger$: MAML with 5 inner loop steps in meta-training/testing. $\star$: we carefully select the number of inner loop steps, based on the meta-validation set.
  }
  \vspace{-10pt}
  \tabcolsep 3pt
    \begin{tabular}{c|cc|cc}
    \addlinespace
    \toprule
    \multicolumn{ 1}{c}{Dataset $\rightarrow$} & \multicolumn{ 2}{|c}{{\it Mini}ImageNet} & \multicolumn{ 2}{|c}{{\it Tiered}ImageNet} \\
    \midrule
    \multicolumn{ 1}{c|}{Setups $\rightarrow$} & 1-Shot & 5-Shot & 1-Shot & 5-Shot \\
    \midrule
    ProtoNet~{\citep{SnellSZ17Prototypical}} & 62.39 $\pm$ 0.20 & 80.53 $\pm$ 0.20 & 68.23 $\pm$ 0.23 & 84.03 $\pm$ 0.16 \\
    ProtoMAML~{\citep{Triantafillou2019Meta}} & 64.12 $\pm$ 0.20 & 81.24 $\pm$ 0.20 & 68.46 $\pm$ 0.23 & 84.67 $\pm$ 0.16 \\
    MetaOptNet~{\citep{Lee2019Meta}} & 62.64 $\pm$ 0.35 & 78.63 $\pm$ 0.68 & 65.99 $\pm$ 0.72 & 81.56 $\pm$ 0.53 \\
    MTL+E3BM~{\citep{Sun2019Meta}} & 63.80 $\pm$ 0.40 & 80.10 $\pm$ 0.30 & 71.20 $\pm$ 0.40 & 85.30 $\pm$ 0.30 \\
    RFS-Distill~{\citep{Tian2020Rethinking}} & 64.82 $\pm$ 0.60 & 82.14 $\pm$ 0.43 & 69.74 $\pm$ 0.72 & 84.41 $\pm$ 0.55 \\
    DeepEMD~{\citep{Zhang2020Deep}} & 65.91 $\pm$ 0.82 & 82.41 $\pm$ 0.56 & {\bf 71.52 $\pm$ 0.69} & 86.03 $\pm$ 0.49 \\
    MATE+MetaOpt~{\citep{Chen2020MATE}} & 62.08 $\pm$ 0.64 & 78.64 $\pm$ 0.46 & 71.16 $\pm$ 0.87 & 86.03 $\pm$ 0.58 \\
    DSN-MR~{\citep{Simon2020Adaptive}} & 64.60 $\pm$ 0.72 & 79.51 $\pm$ 0.50 & 67.39 $\pm$ 0.82 & 82.85 $\pm$ 0.56 \\
    FEAT~{\citep{ye2020fewshot}}  & {\bf 66.78 $\pm$ 0.20} & 82.05 $\pm$ 0.14 & 70.80 $\pm$ 0.23 & 84.79 $\pm$ 0.16 \\
    \midrule
    MAML (5-Step$^\dagger$) & 62.90 $\pm$ 0.20 & 80.81 $\pm$ 0.14 & 59.08 $\pm$ 0.20 & 80.04 $\pm$ 0.16 \\
    MAML (our reimplementation$^\star$) & 64.42 $\pm$ 0.20 & 83.44 $\pm$ 0.14 & 65.72 $\pm$ 0.20 & 84.37 $\pm$ 0.16 \\
    \ourmethod & 65.17 $\pm$ 0.20 & {\bf 84.30 $\pm$ 0.14} & 69.24 $\pm$ 0.20 & {\bf 86.06 $\pm$ 0.16} \\
    \bottomrule
    \end{tabular}
  \label{tab:mini}
  \vskip 5pt
\end{table}

\autoref{tab:mini} summarizes the results of \ourmethod, MAML, and many existing few-shot learning algorithms. \ourmethod consistently improves MAML: on {\it Mini}ImageNet, \ourmethod has a $0.7\%$ gain on one-shot tasks and a $0.8\%$ gain on five-shot tasks; on {\it Tiered}ImageNet, \ourmethod has significant improvements (a $3.5\%$ gain on one-shot tasks and a $1.6\%$ gain on five-shot tasks).
More importantly, \ourmethod performs on a par with many recent algorithms on one-shot tasks, and achieves the highest accuracy on five-shot tasks. Specifically, compared to ProtoMAML and MetaOptNet, which are both permutation-invariant variants of MAML (see the related work paragraph at the end of this section), \ourmethod notably outperforms them.

\noindent\textbf{Other results.}
We evaluate \ourmethod on \textbf{CUB} \citep{WahCUB_200_2011} and use the ConvNet backbone on {\it Mini}ImageNet in the appendix. \ourmethod achieves promising improvements.

\noindent\textbf{Why does \ourmethod work?} The design of \ourmethod ensures that, without using the support set to update the model, the model simply performs at the chance level on the query set. In other words, its formulation inherently helps prevent memorization over-fitting~\citep{yin2020meta}.

\begin{figure}[h]
\centering
\begin{minipage}{0.22\linewidth}
\centering
\includegraphics[width=\linewidth]{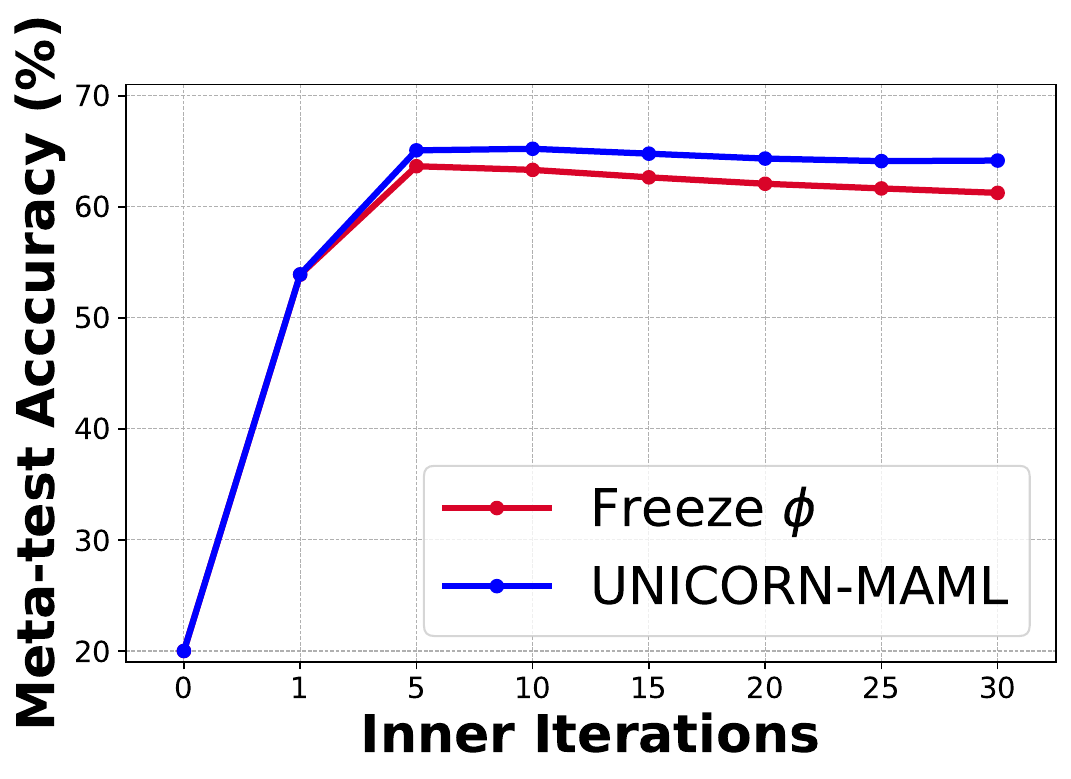} \vskip -5pt
\centering\mbox{\scriptsize(a) {\it Mini}ImageNet, 1-Shot}
\end{minipage}
\hfill
\begin{minipage}{0.22\linewidth}
\centering
\includegraphics[width=\linewidth]{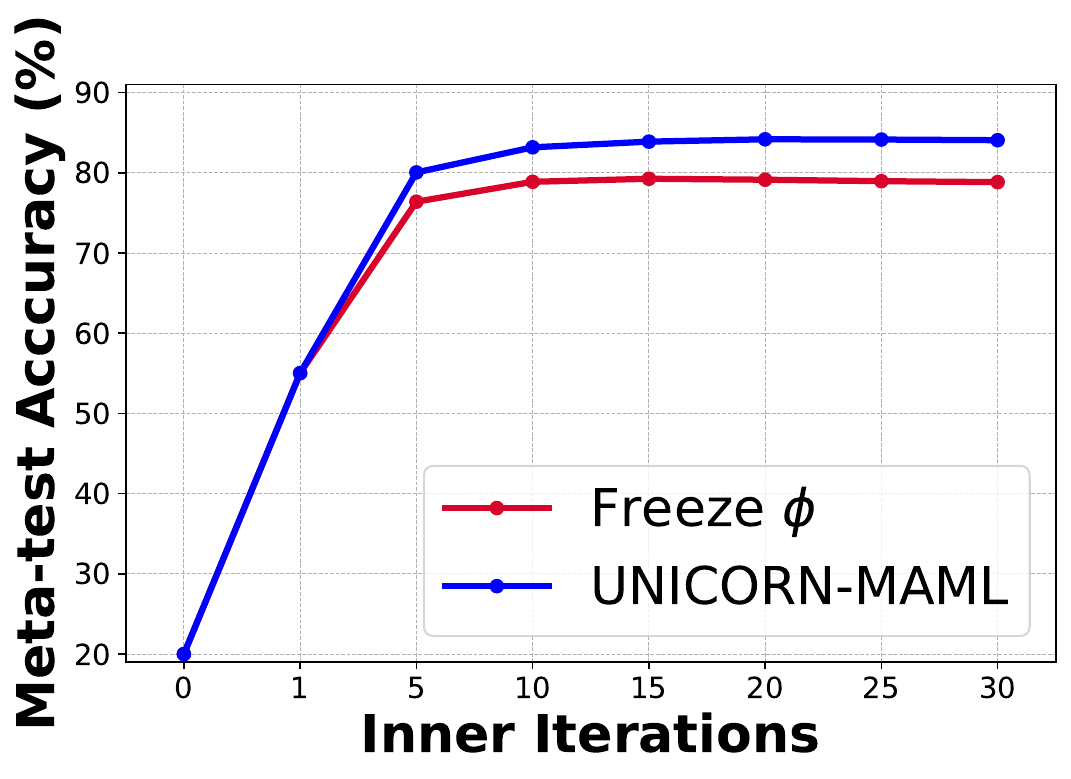} \vskip -5pt
\centering\mbox{\scriptsize(b) {\it Mini}ImageNet, 5-Shot}
\end{minipage}
\hfill
\begin{minipage}{0.22\linewidth}
\centering
\includegraphics[width=\linewidth]{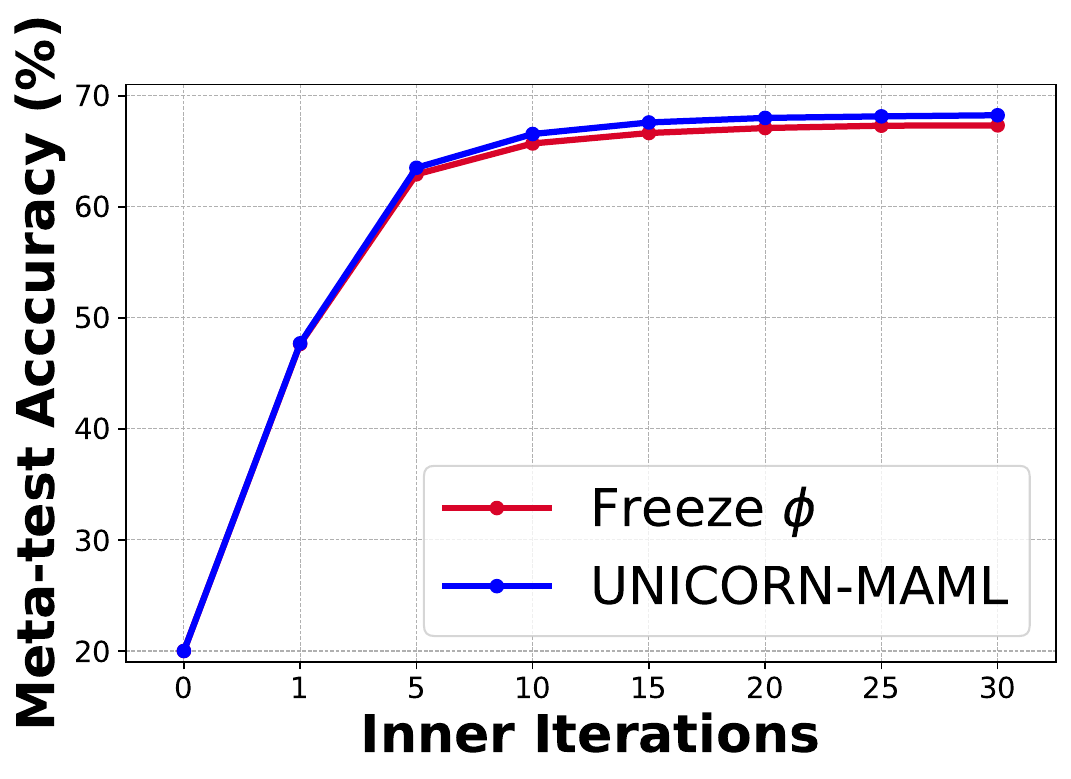} \vskip -5pt
\centering\mbox{\scriptsize(c) {\it Tiered}ImageNet, 1-Shot}
\end{minipage}
\hfill
\begin{minipage}{0.22\linewidth}
\centering
\includegraphics[width=\linewidth]{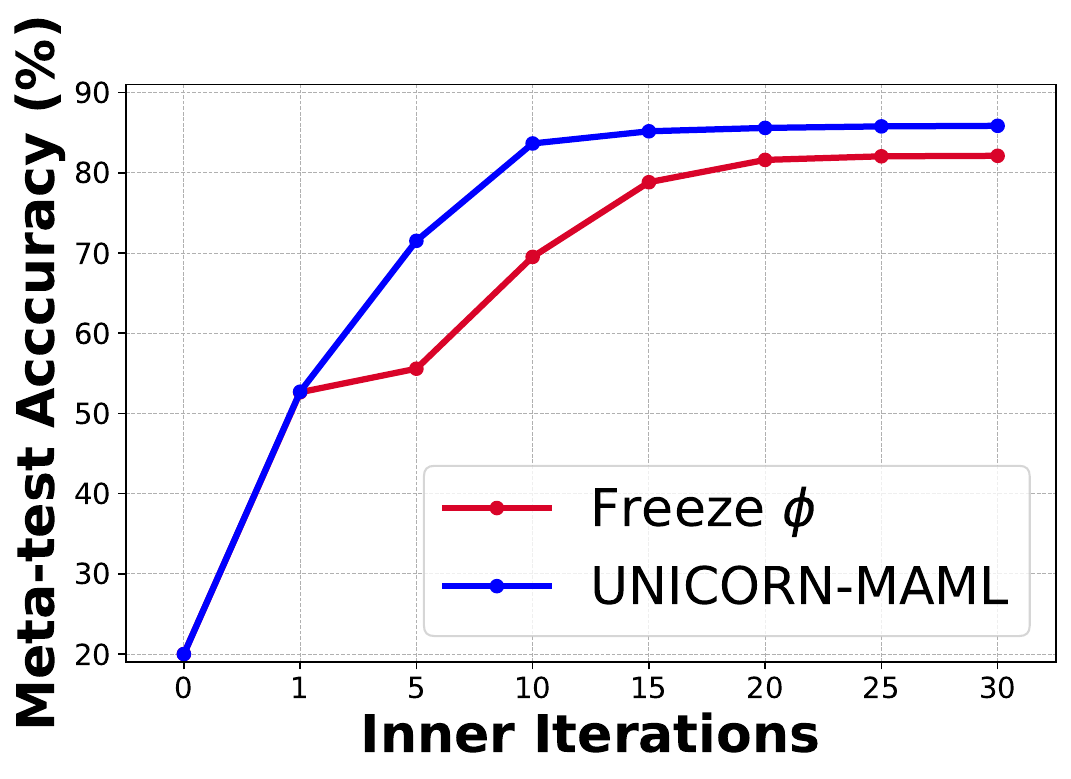}
\centering\mbox{\scriptsize(d) {\it Tiered}ImageNet, 5-Shot}
\end{minipage}
\vskip -10pt
\caption{\small The change of meta-testing accuracy (over 10,000 tasks) along with the process of inner loop updates based on \ourmethod. We investigate updating/freezing the feature extractor $\vphi$.}\label{fig:vanilla_vs_unicorn}
\vskip 2pt
\end{figure}

\textbf{Embedding adaptation is needed.} We analyze \ourmethod in terms of its inner loop updates during meta-testing, similar to \autoref{fig:multistep}. This time, we also investigate updating or freezing the feature extractor $\vphi$. \autoref{fig:vanilla_vs_unicorn} shows the results on five-way one- and five-shot tasks on both datasets. \ourmethod's accuracy again begins with $20\%$ but rapidly increases along with the inner loop updates. In three out of four cases,
adapting the feature extractor $\vphi$ is necessary for claiming a higher accuracy, even if the backbone has been well pre-trained, which aligns with the recent claim by~\cite{Seb2021Embedding}: \emph{``Embedding adaptation is still needed for few-shot learning.''}

\textbf{Experiments on larger shots and transferability.} MAML or similar algorithms that involve a bi-level optimization problem (\ie, inner loop and outer loop optimization) are often considered more {complicated} and {computationally expensive} than algorithms without bi-level optimization, such as ProtoNet~\citep{SnellSZ17Prototypical} or SimpleShot~\citep{wang2019simpleshot}. Nevertheless, the inner loop optimization does strengthen MAML's adaptability during meta-testing, especially (a) when the meta-testing tasks are substantially different from the meta-training tasks (\eg, meta-training using {\it Mini}ImageNet but meta-testing on CUB) or (b) when the number of shots increases (\eg, from $1\sim 5$ to $10\sim 50$). In \autoref{tab:multi-shot} and \autoref{tab:cross_data}, we conduct further experiments to justify these aspects. 

\noindent\textbf{Related work.} We note that some variants of MAML are permutation-invariant, even though they are not designed for the purpose. For example, LEO~\citep{Rusu2018LEO} computes class prototypes (\ie, averaged features per class) to encode each class and uses them to produce task-specific initialization. However, it introduces additional sub-networks. MetaOptNet~\citep{Lee2019Meta} performs inner loop optimization only on $\{\vw_c\}_{c=1}^N$ (till convergence), making it a convex problem which is not sensitive to the initialization and hence the permutations. This method, however, has a high computational burden and needs careful hyper-parameter tuning for the additionally introduced regularizers. Proto-MAML \citep{Triantafillou2019Meta} initializes the linear classifiers $\{\vw_c\}_{c=1}^N$ with the prototypes, which could be permutation-invariant but cannot achieve accuracy as high as our \ourmethod.

%% file: disc.tex
\begin{table}[] 
\centering
\begin{minipage}[c]{0.56\textwidth}
\centering
\tabcolsep 0.5pt
\small
\vspace{-10pt}
\caption{\small 5-Way 10/20/30/50-Shot classification accuracy on {\it Mini}ImageNet over 10,000 tasks with a ResNet backbone. (The confidence interval is omitted due to space limit.)}
\vspace{-10pt}
\begin{tabular}{c|cccc}
    \addlinespace
    \toprule
     Multi-Shot & 10-shot & 20-shot & 30-shot & 50-shot \\
    \midrule
    SimpleShot{\scriptsize~\citep{wang2019simpleshot}} & 84.89 & 86.91 & 87.53  & 88.08  \\
    ProtoNet{\scriptsize~\citep{SnellSZ17Prototypical}} & 82.83 &  84.61 & 85.07  & 85.57 \\
    FEAT{\scriptsize~\citep{ye2020fewshot}}  & 85.15  & 87.09 & 87.82  & 87.83 \\
    \midrule
    MAML (our implementation)  & 88.08  & 90.23  & 91.06  & 92.14 \\
    \ourmethod & \bf 88.38  & \bf 90.96 & \bf 91.96  & \bf 92.86\\
    \bottomrule
    \end{tabular}
  \label{tab:multi-shot}
\end{minipage}
\hfill
\begin{minipage}[c]{0.41\textwidth}
\centering
\small
\tabcolsep 0.5pt
\vspace{-10pt}
\caption{\small  5-Way 1/5-Shot classification accuracy on CUB based on the best learned model with a ResNet backbone from {\it Mini}ImageNet.}
\vspace{-10pt}
\begin{tabular}{c|cc}
\addlinespace
\toprule
 {\it Mini}ImageNet $\rightarrow$ CUB & 1-Shot & 5-Shot \\
\midrule
Baseline++{\scriptsize~\citep{chen2019closer}} & 50.37 & 73.30\\
ProtoNet{\scriptsize~\citep{SnellSZ17Prototypical}} & 50.01 & 72.02\\
Neg-Cosine{\scriptsize~\citep{liu2020negative}} & 47.74 & 69.30\\
\midrule
MAML (our implementation)  & 51.25 & 73.86\\
\ourmethod & \bf 51.80 & \bf 75.67\\
\bottomrule
\end{tabular}\label{tab:cross_data}
\end{minipage}
\end{table}

\section{Discussion and Conclusion}
\label{s_disc}

There have been an abundance of ``novel'' algorithms proposed for few-shot classification~\citep{hospedales2020meta,wang2020generalizing}. In these papers, MAML~\citep{FinnAL17Model} is frequently considered as a baseline, but shows inferior results. This raises our interests. Is it because MAML is not suitable for few-shot classification, or is it because MAML has not been applied appropriately to the problem?

We thus conduct a series of analyses on MAML for few-shot classification, including hyper-parameter tuning and the sensitivity to the permutations of class label assignments in few-shot tasks. 
We find that by using a large number of inner loop gradient steps (in both meta-training and meta-testing), MAML can achieve comparable results to many existing algorithms. By further making MAML permutation-invariant to the class label assignments, we present \ourmethod, which outperforms many existing algorithms on five-shot tasks, without the need to add extra sub-networks.
We hope that \ourmethod could serve as a strong baseline for future work in few-shot classification.

%% file: suppl_content.tex

We provide contents that we omit in the main text.

\begin{itemize}
    \item \autoref{suppl_s_related}: Additional background and related work.
    \item \autoref{suppl_s_setup}: details of experimental setups (cf. \autoref{s_pre}).
    \item \autoref{suppl_s_intro}: details and analyses of permutations of class label assignments (cf. \autoref{s_approach_2}).
    \item \autoref{suppl_s_CUB}: Additional experimental results.
    \item \autoref{suppl_s_dropout}: Additional explanation of our studied methods.
\end{itemize}

\input{related}

\section{Details of Experimental Setups}
\label{suppl_s_setup}

We follow the pre-training procedure in~\citep{ye2020fewshot}. During pre-training we append the feature extractor backbone with a fully-connected layer for classification, and train it to classify all classes in the base class set (\eg, 64 classes in the {\it Mini}ImageNet) with the cross-entropy loss. In this stage, we apply standard ImageNet image augmentations (\eg, random crop and random flip). The best pre-trained feature extractor (\ie, epoch) is selected based on the one-shot classification accuracy on the meta-validation set. Specifically, we sample one-shot tasks from the meta-validation set, and apply the nearest neighbor classifier on top of the extracted features to evaluate the quality of the backbone. Finally, the best pre-trained backbone is used to initialize the feature extractor for MAML.

\section{Permutations of Class Label Assignments}
\label{suppl_s_intro}

We provide more analyses and discussions on the permutation issue in the class label assignment. As illustrated in \autoref{fig:ourmethod} (a), few-shot tasks of the same set of $N$ semantic classes (\eg, ``unicorn'', ``bee'', etc.) can be associated with different label assignments (\ie, $c\in[N]$) and are paired with the learned initialization $\{\vw_c\}_{c=1}^N$ of MAML differently. For five-way tasks, there are $120$ permutations.

In \autoref{s_approach_2}, we study how the permutations affect the meta-testing accuracy (after the inner loop optimization of $\vphi$ and $\{\vw_c\}_{c=1}^N$) and see a high variance among the permutations.
We note that, the inner loop optimization updates not only the linear classifiers $\{\vw_c\}_{c=1}^N$, but also $\vphi$ of the feature extractor. Different permutations therefore can lead to different feature extractors.

\begin{figure}[t]
    \centering
    \minipage{0.49\columnwidth}   
    \centering
    \includegraphics[width=1.0\linewidth]{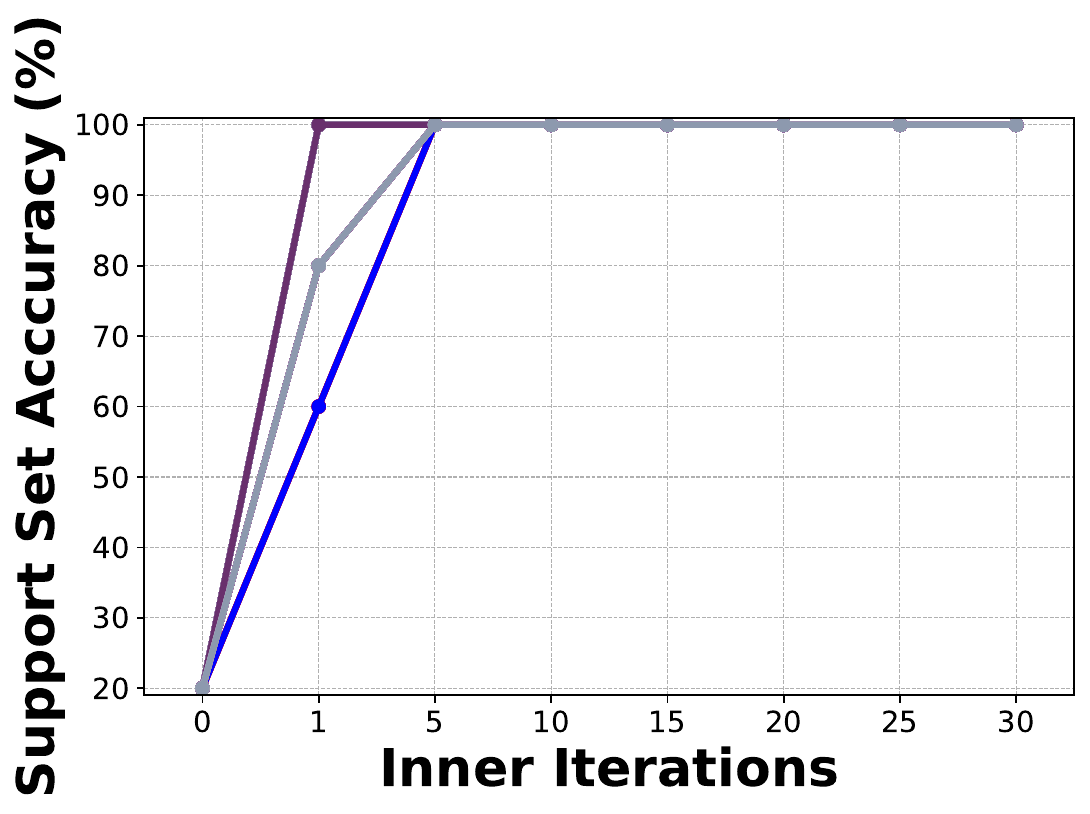}
    \endminipage\hfill
    \minipage{0.49\columnwidth}
    \centering
    \includegraphics[width=1.0\linewidth]{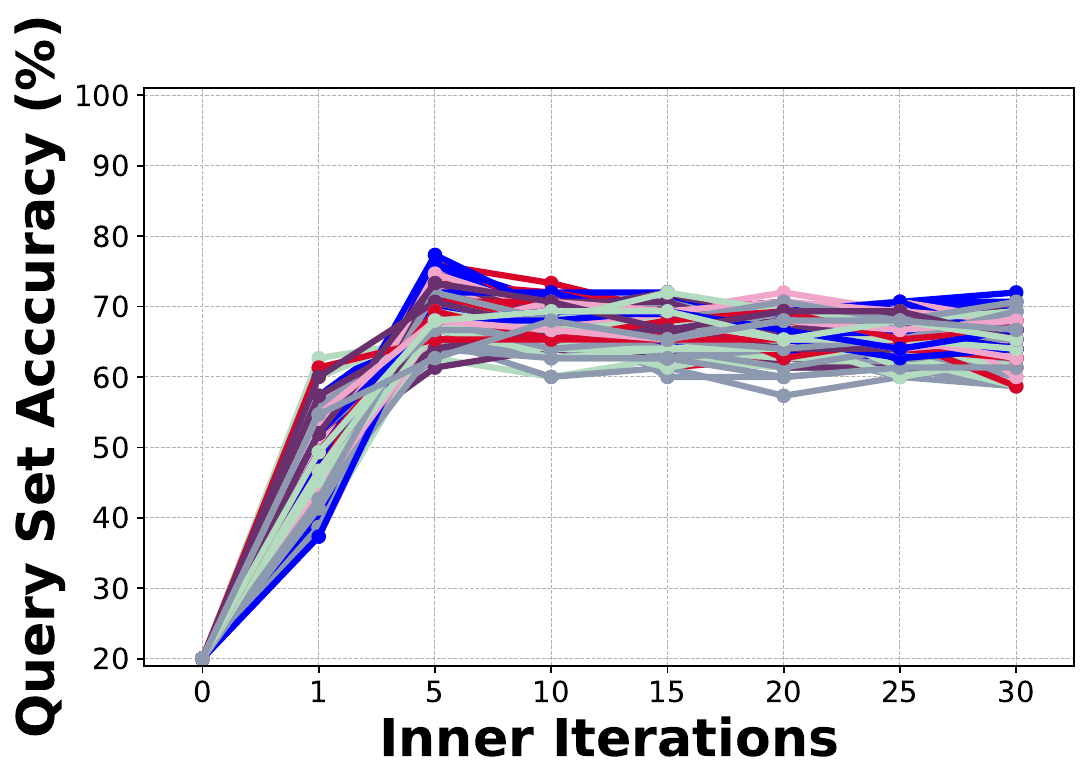}
    \endminipage\hfill
    \caption{The support (left) and query (right) set accuracy on a randomly sampled five-way one-shot meta-testing task from {\it Mini}ImageNet. We plot the accuracy of each permutation (totally 120) along with the process of inner loop optimization (the same permutation is colored the same in the left and right images).}
    \label{fig:support_query}
\end{figure}

Here, we further sample a five-way one-shot meta-testing task, and study the change of accuracy along with the inner loop updates (using a MAML trained with a fixed number of inner loop updates). Specifically, we plot both the support set and query set accuracy for each permutation. As shown in \autoref{fig:support_query}, there exists a high variance of query set accuracy among permutations after the inner loop optimization.
This is, however, not the case for the support set. (The reason that only three curves appear for the support set is because there are only five examples, and all the permutations reach $100\%$ support set accuracy within five inner loop steps.)
Interestingly, for all the permutations, their initialized accuracy (\ie, before inner loop optimization) is all $20\%$. After an investigation, we find that the meta-learned $\{\vw_c\}_{c=1}^N$ (initialization) is dominated by one of them; \ie, all the support or query examples are classified into one class. While this may not always be the case for other few-shot tasks or if we re-train MAML, for the task we sampled, it explains why we obtain $20\%$ for all permutations. We note that, even with an initial accuracy of $20\%$, the learned initialization can be updated to attain high classification accuracy.

We further compare the change of support and query set accuracy along with the inner loop optimization in \autoref{fig:support_query}. We find that, while both accuracy increases, since the support set accuracy converges quickly and has a smaller variance among permutations, it is difficult to use its information to determine which permutation leads to the highest query set accuracy.
This makes sense since the support set is few-shot: its accuracy thus cannot robustly reflect the query set accuracy. 
This explains why the methods studied in \autoref{tab:permutation_select} cannot determine the best permutation for the query set.

\subsection{Mathematical explanation}
\label{supp_ss_math}
We provide a simple mathematical explanation for why, on average, the query set accuracy is at the chance level if we directly apply the learned initialized model.
Suppose we have a five-way task with five semantic classes $\{\text{``dog'', ``cat'', ``bird'', ``car'', ``person''}\}$. 
Without loss of generality, let us assume that the query set has only five examples, one from each class.
Let us also assume that the best permutation --- \ie, the best assignment of $\{\vw_1, \cdots, \vw_5\}$ to these classes --- 
gives a $100\%$ query set accuracy using the initialized model. 
Since there are in total $120$ possible permutations, there will be $10$ of them with $60\%$ accuracy (\ie, by switching two-class indices), $20$ of them with $40\%$ accuracy (\ie, by shuffling the indices of three classes such that they do not take their original indices), $45$ of them with $20\%$ accuracy, and $54$ of them with $0\%$ accuracy. Taking an average over these permutations gives a $20\%$ accuracy. In other words, even if one of the permutations performs well, on average the accuracy will be close to random.

\begin{figure}[t]
\centering
\minipage{0.28\textwidth}
\centering
{\includegraphics[width=1\textwidth]{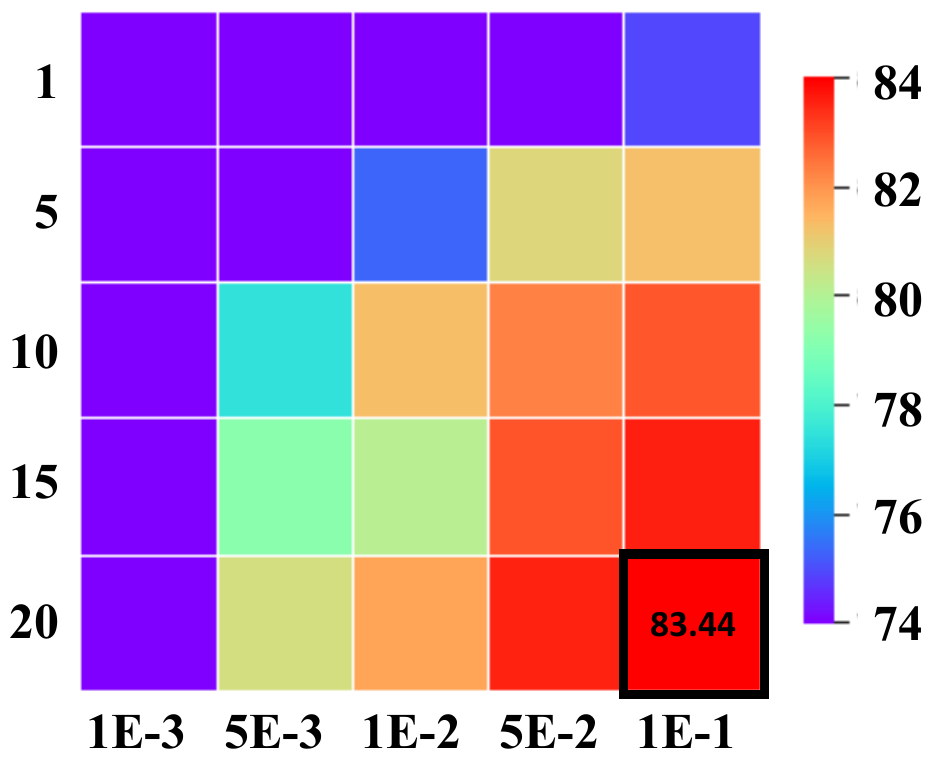}}\\
\mbox{{\it Mini}ImageNet, ResNet}
\endminipage
\hfill
\minipage{0.28\textwidth}
\centering
{\includegraphics[width=1\textwidth]{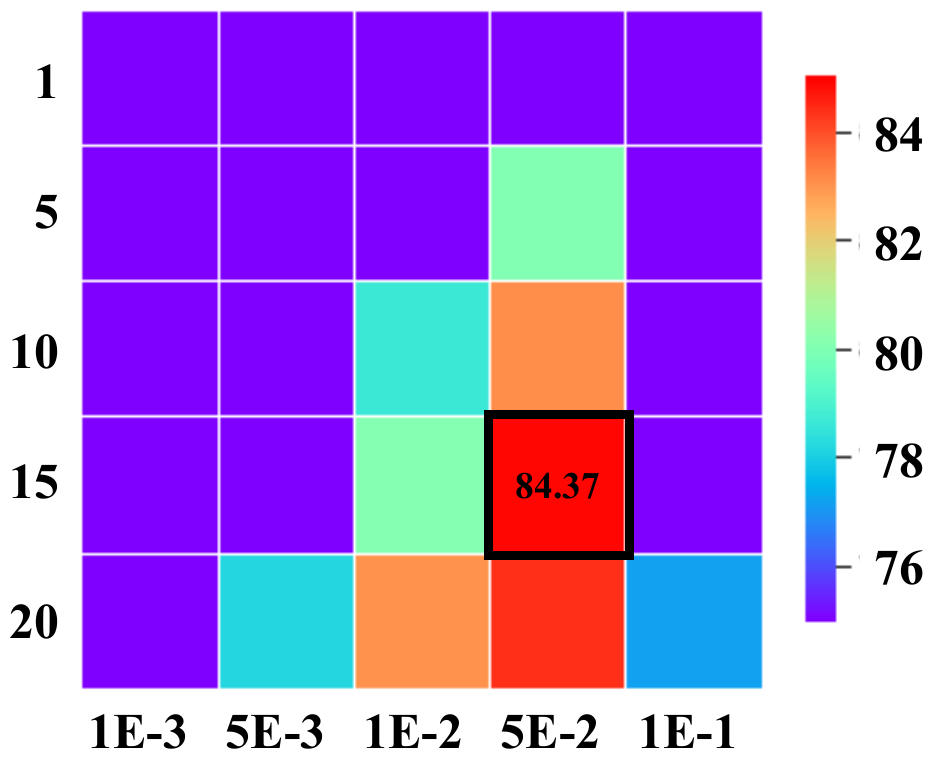}}\\
\mbox{{\it Tiered}ImageNet, ResNet}
\endminipage
\hfill
\minipage{0.28\textwidth}
\centering
{\includegraphics[width=1\textwidth]{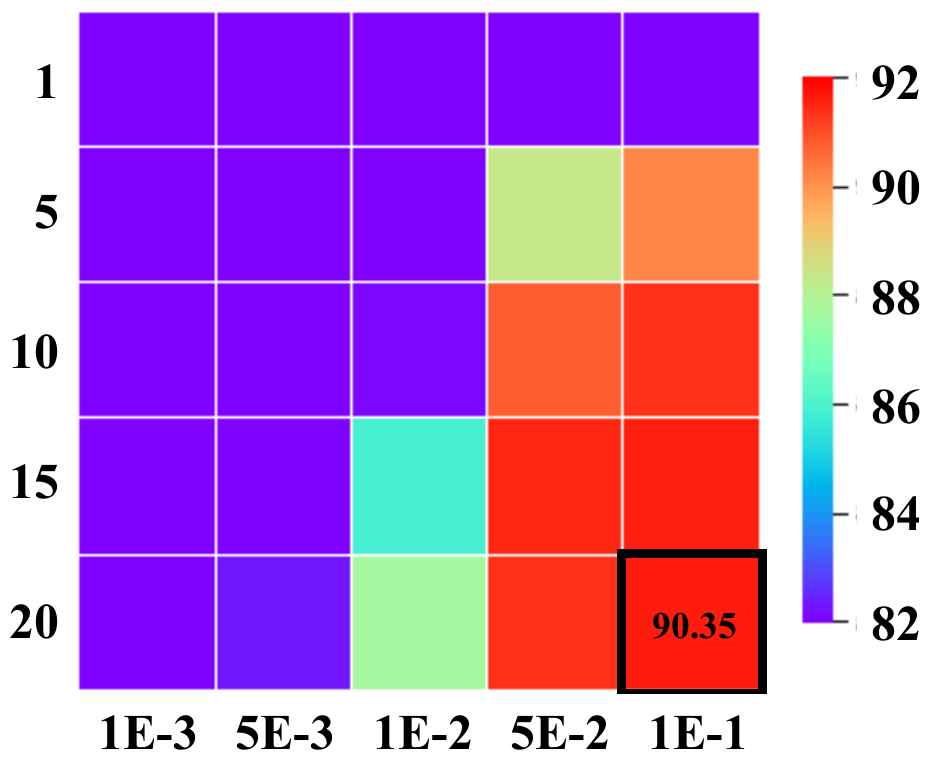}}\\
\mbox{CUB, ResNet}
\endminipage

\minipage{0.28\textwidth}
\centering
{\includegraphics[width=1\textwidth]{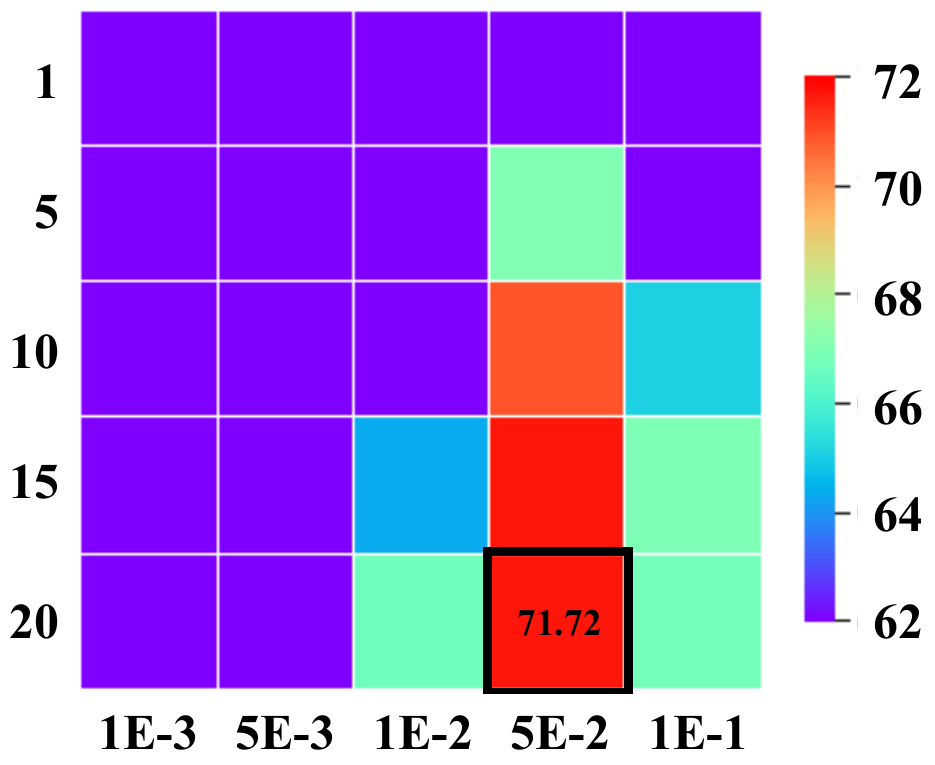}}\\
\mbox{{\it Mini}ImageNet, ConvNet}
\endminipage
\hfill
\minipage{0.28\textwidth}
\centering
{\includegraphics[width=1\textwidth]{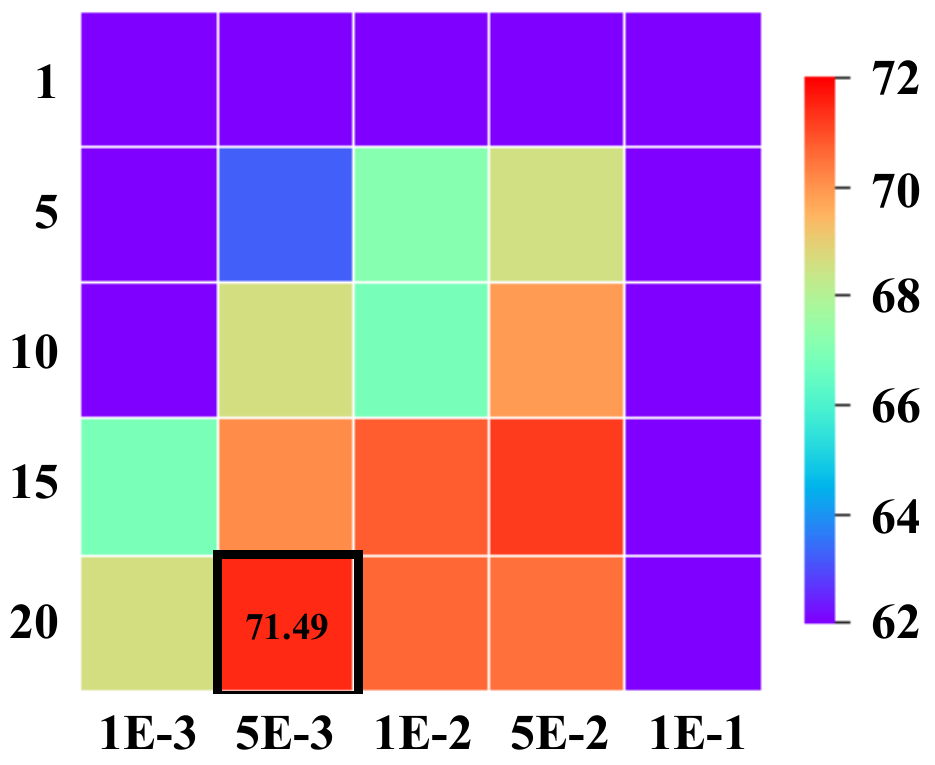}}\\
\mbox{{\it Tiered}ImageNet, ConvNet}
\endminipage
\hfill
\minipage{0.28\textwidth}
\centering
{\includegraphics[width=1\textwidth]{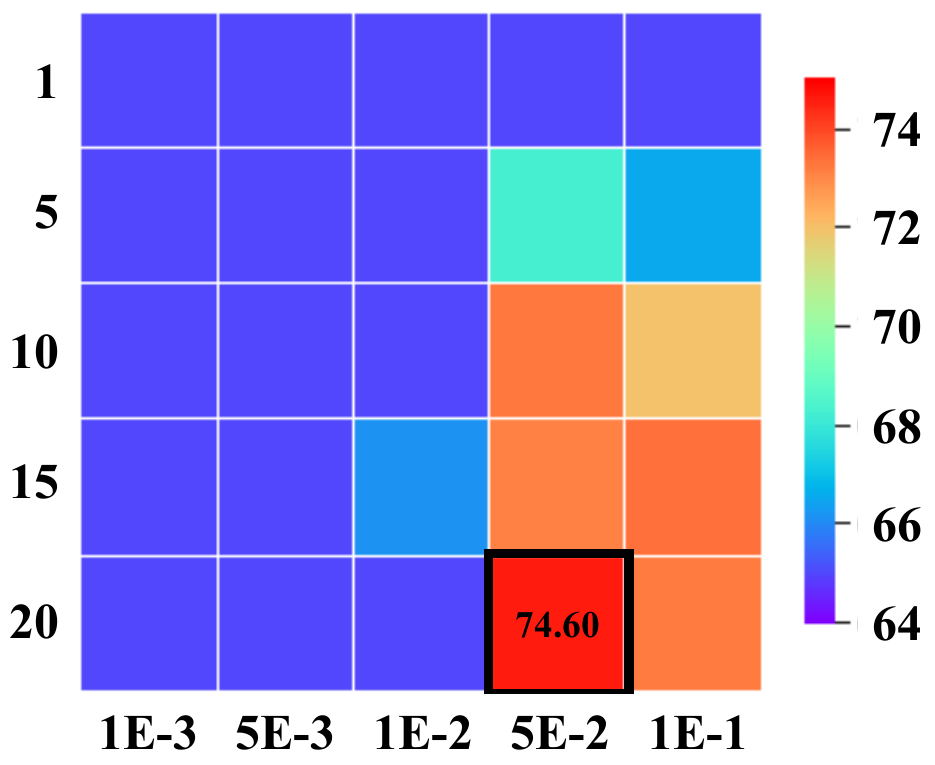}}\\
\mbox{CUB, ConvNet}
\endminipage
\caption{Heat maps of MAML's five-way five-shot accuracy on {\it Mini}ImageNet, {\it Tiered}ImageNet, and CUB w.r.t. the inner loop learning rate $\alpha$ (x-axis) and the number of inner loop updates $M$ (y-axis). 
For each map, \textbf{we set accuracy below a threshold to a fixed value for clarity}; we denote the best accuracy by a black box.}\label{fig:heatmap2}
\end{figure}

\begin{figure}[t]
\centering
\minipage{0.32\textwidth}
\centering
{\includegraphics[width=1\textwidth]{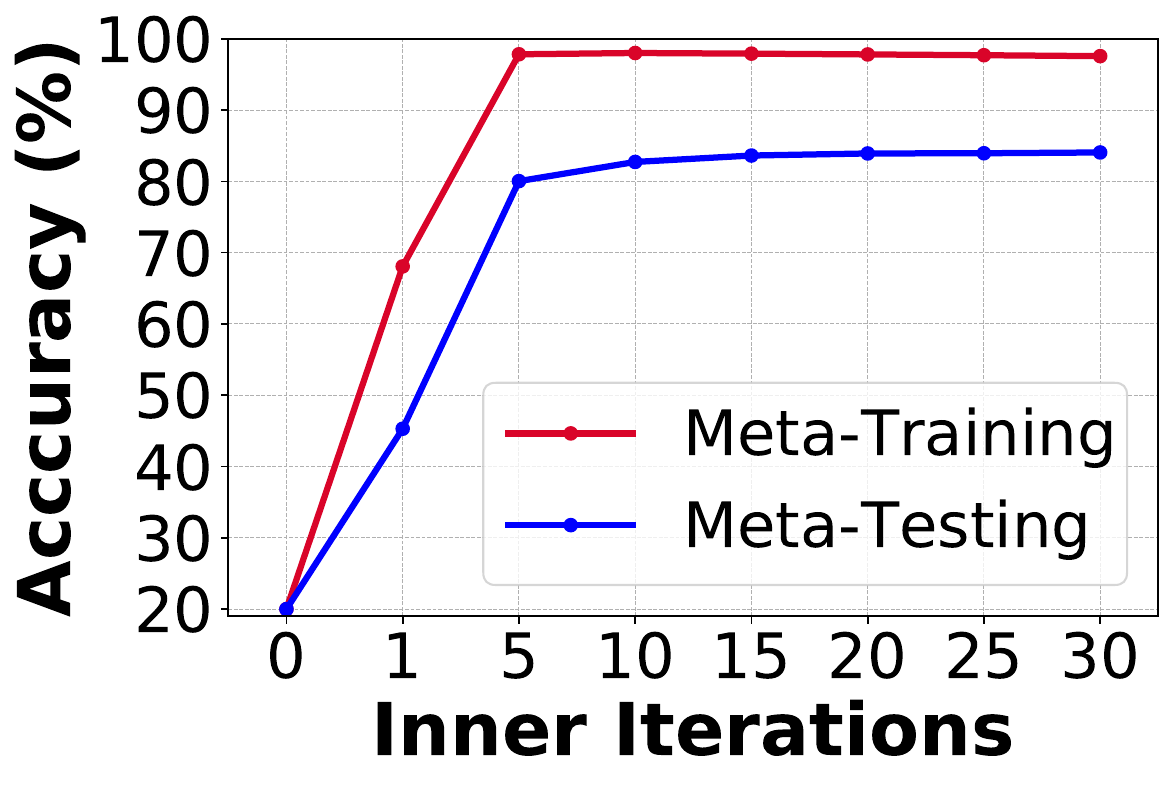}}\\
\mbox{{\it Mini}ImageNet, ResNet}
\endminipage
\hfill
\minipage{0.32\textwidth}
\centering
{\includegraphics[width=1\textwidth]{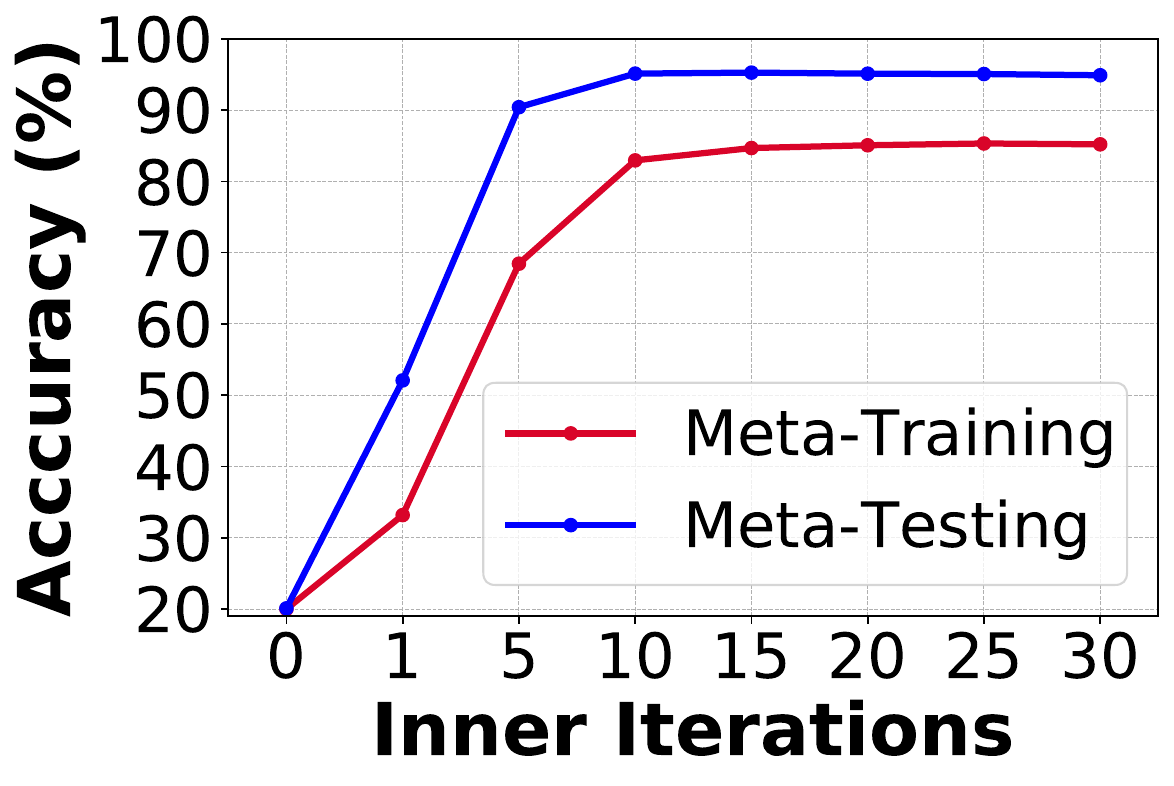}}\\
\mbox{{\it Tiered}ImageNet, ResNet}
\endminipage
\hfill
\minipage{0.32\textwidth}
\centering
{\includegraphics[width=1\textwidth]{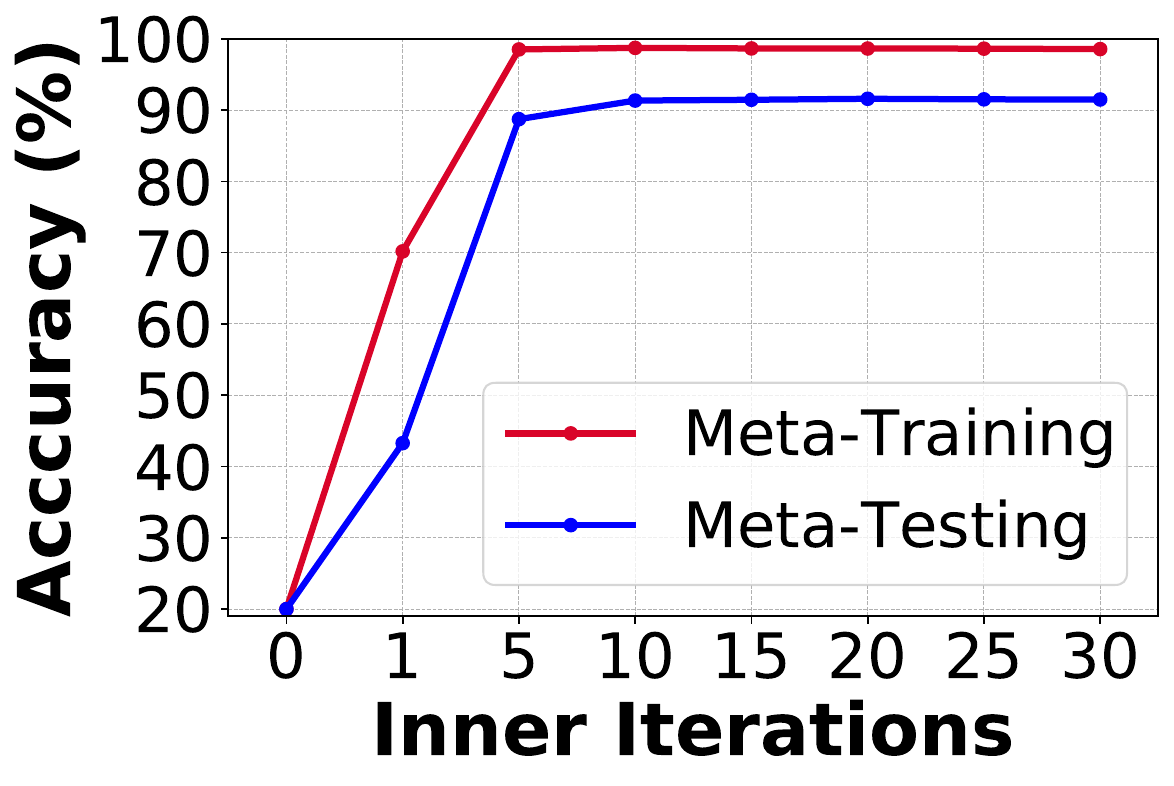}}\\
\mbox{CUB, ResNet}
\endminipage

\minipage{0.32\textwidth}
\centering
{\includegraphics[width=1\textwidth]{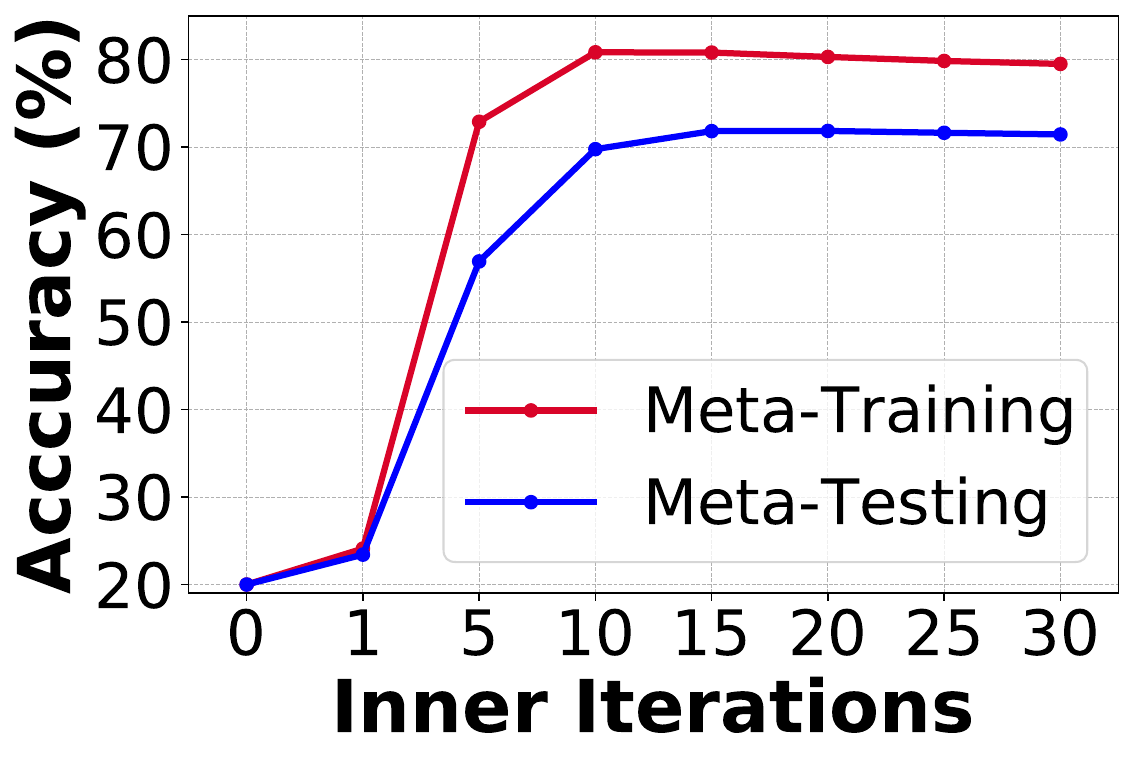}}\\
\mbox{{\it Mini}ImageNet, ConvNet}
\endminipage
\hfill
\minipage{0.32\textwidth}
\centering
{\includegraphics[width=1\textwidth]{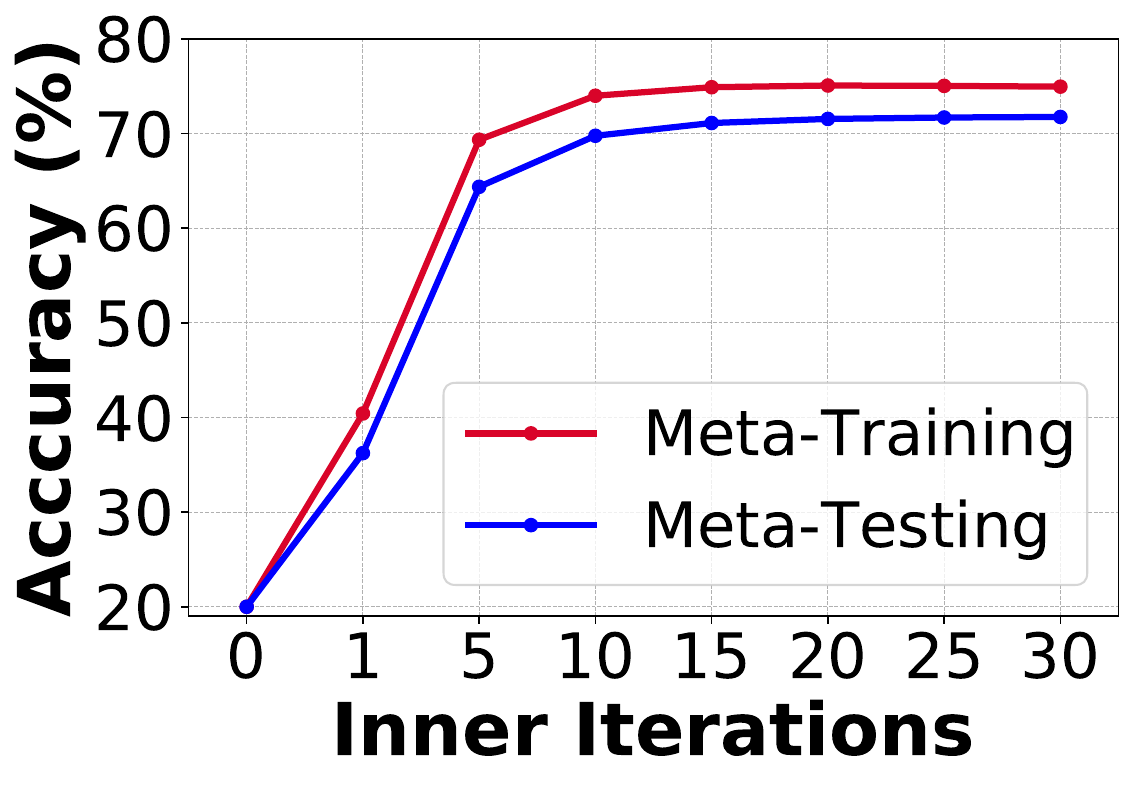}}\\
\mbox{{\it Tiered}ImageNet, ConvNet}
\endminipage
\hfill
\minipage{0.32\textwidth}
\centering
{\includegraphics[width=1\textwidth]{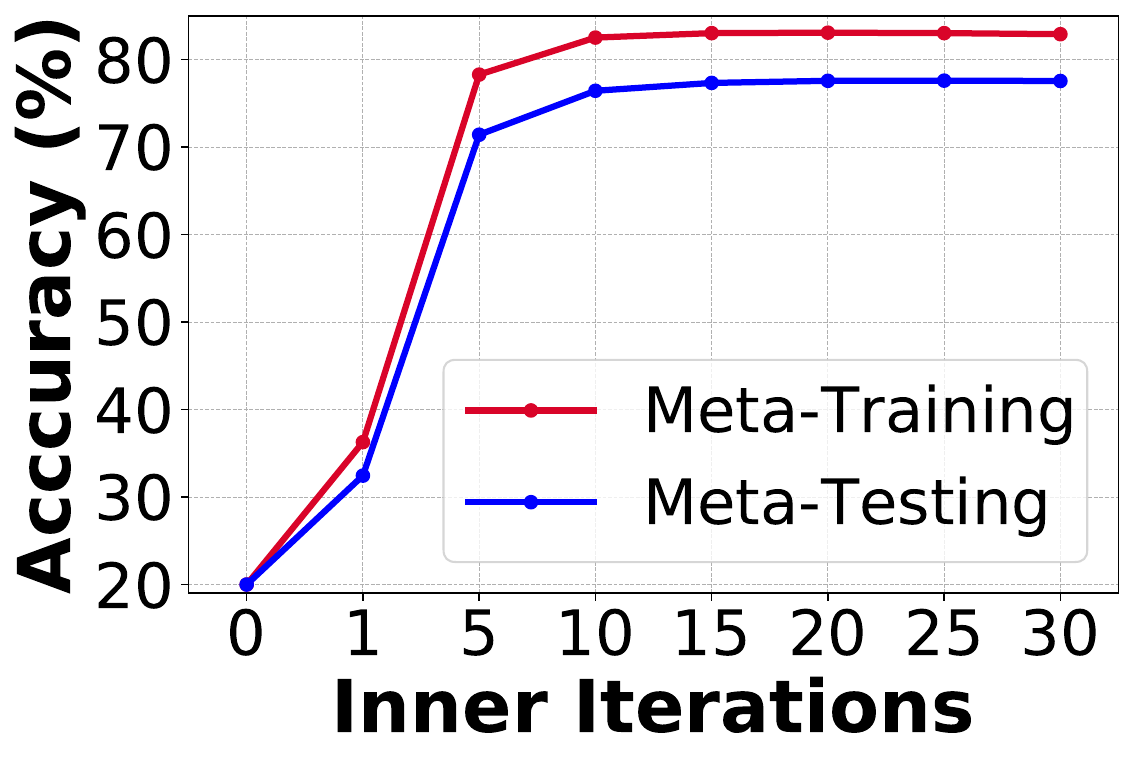}}\\
\mbox{CUB, ConvNet}
\endminipage
\caption{We plot the change of the five-way five-shot classification accuracy (on the query set), averaged over 10,000 tasks sampled from either the meta-training (red) or meta-testing classes (blue), along with the process of inner loop updates, using the best model initialization learned by MAML.}\label{fig:multistep2}
\end{figure}

\section{Additional Experimental Results}
\label{suppl_s_CUB}

Similar to \autoref{fig:heatmap}, we plot the meta-testing accuracy of five-way five-shot tasks on the three datasets over ResNet and ConvNet backbones in \autoref{fig:heatmap}. We get a similar trend with \autoref{fig:heatmap} where MAML achieves higher and much more stable accuracy (w.r.t. the learning rate) when $M$ is larger than $15$.
Specifically, for {\emph{Mini}ImageNet} with ResNet, the highest accuracy $83.44\%$ is obtained with $M=20$, higher than $80.81\%$ with $M=5$. 

We also plot the change of five-way five-shot classification accuracy (on the query set), averaged over 10,000 tasks sampled from either the meta-training (red) or meta-testing classes (blue), along with the process of inner loop updates, using the best model initialization learned by MAML in \autoref{fig:multistep2} for each pair of dataset and backbone. There indicates a similar phenomenon as \autoref{fig:multistep}, where MAML needs a large $M$. 

We further evaluate \ourmethod on CUB dataset with ResNet-12 backbone and on {\it Mini}ImageNet with the four-layer ConvNet backbone. The results are listed in  
\autoref{tab:cub} and \autoref{tab:mini_conv}, respectively. By comparing \ourmethod with others, we find the carefully tuned MAML shows promising results and \ourmethod outperforms the existing methods.

\begin{table}[t]
  \centering
  \caption{5-Way 1/5-Shot classification accuracy and 95\% confidence interval on CUB, evaluated over 10,000 tasks with a ResNet-12 backbone. $\ddagger$: methods with a ResNet-18 backbone. $\dagger$: We train MAML with 5 inner loop steps in both meta-training and meta-testing. $\star$: we carefully select the number of inner loop steps for MAML, based on the meta-validation set.}
    \begin{tabular}{c|cc}
    \addlinespace
    \toprule
    ResNet-12 & 1-Shot & 5-Shot \\
    \midrule
    MatchNet~\citep{VinyalsBLKW16Matching} &66.09 $\pm$ 0.92  & 82.50 $\pm$ 0.58 \\
    ProtoNet~\citep{SnellSZ17Prototypical}   &71.87 $\pm$ 0.85  & 85.08 $\pm$ 0.57 \\
    DeepEMD~\citep{Zhang2020Deep}        &75.65 $\pm$ 0.83   & 88.69 $\pm$ 0.50 \\
    Baseline++~\citep{chen2019closer}$^\ddagger$        &67.02 $\pm$ 0.90   & 83.58 $\pm$ 0.54\\
    AFHN$^\dagger${~\citep{li2020adversarial}}             & 70.53 $\pm$ 1.01   & 83.95 $\pm$ 0.63  \\
    Neg-Cosine~\citep{liu2020negative}$^\ddagger$        &72.66 $\pm$ 0.85   & 89.40 $\pm$ 0.43 \\
    Align~\citep{afrasiyabi2020associative}$^\ddagger$        &74.22 $\pm$ 1.09   & 88.65 $\pm$ 0.55 \\
    \midrule
    MAML (5-Step$^\dagger$) & 76.53 $\pm$ 0.20 & 88.34 $\pm$ 0.16 \\
    MAML (our reimplementation$^\star$) & 77.67 $\pm$ 0.20 & 90.35 $\pm$ 0.16 \\
    \ourmethod & \bf 78.07 $\pm$ 0.20 & {\bf 91.67 $\pm$ 0.16} \\
    \bottomrule
    \end{tabular}
  \label{tab:cub}
\end{table}

\begin{table}[t]
  \centering
  \caption{5-Way 1-Shot and 5-Shot classification accuracy and 95\% confidence interval on {\it Mini}ImageNet over 10,000 tasks with a four-layer ConvNet backbone. $\dagger$: We train MAML with 5 inner loop steps in both meta-training and meta-testing. $\star$: we carefully select the number of inner loop steps for MAML, based on the meta-validation set.}
    \begin{tabular}{c|cc}
    \addlinespace
    \toprule
    ConvNet & 1-Shot & 5-Shot \\
        \midrule
    MAML~\citep{FinnAL17Model}  & 48.70  $\pm$  1.84 & 63.11  $\pm$  0.92 \\
    MAML++~\citep{Antoniou2018How} & 52.15  $\pm$  0.26 & 68.32  $\pm$  0.44 \\
    Reptile~\citep{Nichol2018Reptile} & 49.97  $\pm$  0.32 & 65.99  $\pm$  0.58 \\
    FEAT~\citep{ye2020fewshot}  & 55.15  $\pm$  0.20 & 71.61  $\pm$  0.16 \\
    KTN (visual)~\citep{Peng2019Few} & 54.61  $\pm$  0.80 & 71.21  $\pm$  0.66 \\
    PARN~\citep{Wu2019PARN}  & 55.22  $\pm$  0.84 & 71.55  $\pm$  0.66 \\
    MAML-MMCF~\citep{yao2021improving} & 50.35  $\pm$  1.82 & 64.91  $\pm$  0.96 \\
	\midrule
	MAML (5-Step$^\dagger$) & 53.15 $\pm$ 0.20 & 67.01 $\pm$ 0.16 \\
    MAML (our reimplementation$^\star$) & 54.89  $\pm$  0.20 & 71.72  $\pm$  0.16 \\
    \ourmethod & {\bf 55.70  $\pm$  0.20} & {\bf 72.68  $\pm$  0.16} \\
    \bottomrule
    \end{tabular}
  \label{tab:mini_conv}
\end{table}

\section{Additional Explanations of Our Studied Methods} 
\label{suppl_s_dropout}

We provide some more explanations on the ensemble and forced permutation-invariant methods introduced in \autoref{s_pi_test}. For the ensemble method, given a few-shot task, we can permute $\{\vw_c\}_{c=1}^N$ to pair them differently with such a task. We can then perform different inner loop optimizations to obtain a set of five-way classifiers that we can perform ensemble upon. In the main text, we average the posterior probabilities of these five-way classifiers to make the final predictions.

Since the permutation affects the meta-training phase as well, we can interpret the meta-training phase as follows. Ever time we sample a few-shot task $\sT = (\sS, \sQ)$, we also sample a permutation $\pi: [N]\mapsto[N]$ to re-label the classes.
(We note that, this is implicitly done when few-shot tasks are sampled.) We then take $\sT_\pi = (\sS_\pi, \sQ_\pi)$ to optimize $\vtheta$ in the inner loop. That is, in meta-training, the objective function in \autoref{eq:maml_obj} can indeed be re-written as
\begin{align}
\mathop{\mathbb{E}}_{(\sS,\sQ)\sim p(\sT), \pi\sim p(\pi)}\sL(\sQ_\pi, \vtheta') = 
\mathop{\mathbb{E}}_{(\sS,\sQ)\sim p(\sT), \pi\sim p(\pi)} \sL(\sQ_\pi, \IL(\sS_\pi, \vtheta, M)),
\label{suppl_eq:maml_obj}
\end{align}
where $p(\pi)$ is a uniform distribution over all possible permutations. \autoref{suppl_eq:maml_obj} can be equivalently re-written as
\begin{align}
\mathop{\mathbb{E}}_{(\sS,\sQ)\sim p(\sT), \pi\sim p(\pi)}\sL(\sQ, \vtheta'_\pi) = 
\mathop{\mathbb{E}}_{(\sS,\sQ)\sim p(\sT), \pi\sim p(\pi)} \sL(\sQ, \IL(\sS, \vtheta_\pi, M)),
\label{suppl_eq:maml_obj_alter}
\end{align}
where $\vtheta_\pi$ means that the initialization of the linear classifiers $\{\vw_c\}_{c=1}^N$ are permuted; $\vtheta'_\pi$ is the corresponding updated model.
This additional \emph{sampling process} of $\pi$ is reminiscent of dropout~\citep{srivastava2014dropout}, which randomly masks out a neural network's neurons or edges to prevent an over-parameterized neural network from over-fitting. During testing, dropout takes expectation over the 
masks. We also investigate a similar idea during meta-testing, by taking expectation (\ie, average) on the learned initiation of the linear classifiers over different permutations.
This results in a new initialization during the meta-testing phase: $\vw_c\leftarrow\frac{1}{N}\sum_{c'=1}^{N} \vw_{c'}$.

%% file: related.tex

\section{Backgrounds on Meta-Learning and Few-Shot Learning}
\label{suppl_s_related}
Training a model under data budgets is important in machine learning, computer vision, and many other application fields, since the costs of collecting data and labeling them are by no means negligible. This is especially the case for deep learning models in visual recognition~\citep{he2016deep,dosovitskiy2020image,simonyan2015very,szegedy2015going,krizhevsky2012imagenet,huang2017densely}, which usually need thousands of, millions of, or even more images to train~\citep{RussakovskyDSKS15ImageNet,deng2009imagenet,guo2016ms,thomee2015yfcc100m,mahajan2018exploring,joulin2016learning} in a conventional supervised manner.
Different from training 
a model to predict at the \emph{instance} level, meta-learning attempts to learn the inductive bias across training {\em tasks}~\citep{Baxter2000Inductive,Vilalta2002Meta}. More specifically, meta-learning aims to train a ``meta-model'' to summarize the common characteristics of tasks and generalize them to those \emph{novel} but related tasks~\citep{Maurer09Transfer,MaurerPR16The,Denevi2018Learning,Ye2021LastShot}.
Meta-learning has been applied in various fields, including few-shot learning~\citep{Sachin2017,SnellSZ17Prototypical,WangH16Learning,ye2020fewshot,ye2020few,Flood2017Learning,zhang2018metagan,Wang2018Low,Tseng2020Cross,Fei2021MELR}, optimization~\citep{AndrychowiczDCH16Learning,wichrowska2017learned,li2017learning,bello2017neural}, reinforcement and imitation learning~\citep{stadie2018importance,frans2017meta,wang2016RL,duan2016rl,duan2017one,Yu2018One}, unsupervised learning~\citep{garg2018supervising,metz2018meta,edwards2017towards,Reed2017Few}, continual learning~\citep{riemer2019learning,kaiser2017learning,al2018continuous}, imbalance learning~\citep{WangRH17Learning,Ren2018Reweight},  transfer and multi-task learning~\citep{MotiianJID17Few,balaji2018metareg,ying2018transfer,zhang2018learning,li2019feature,li2018learning},  active learning~\citep{ravi2018meta,sharma2018learning, bachman2017learning,pang2018meta}, data compression~\citep{Wang2018Dataset},
architecture search~\citep{elsken2018neural}, recommendation systems~\citep{Vartak2017Meta}, 
data augmentation~\citep{ratner2017learning}, teaching~\citep{fan2018learning}, hyper-parameter tuning~\citep{Franceschi2017Bridge,Probst2019Tunability}, etc.

In few-shot learning~(FSL), meta-learning is applied to learn the ability of \emph{``how to build a classifier using limited data''}
that can be generalized across tasks. Such an inductive bias is first learned over few-shot tasks composed of ``base'' classes, and then evaluated on tasks composed of ``novel'' classes. For example, few-shot classification can be implemented in a non-parametric way with soft nearest neighbor~\citep{VinyalsBLKW16Matching} or nearest center classifiers~\citep{SnellSZ17Prototypical}, so that the feature extractor is learned and acts at the task level. 
The learned features pull similar instances together and push dissimilar ones far away, such that a test instance can be classified even with a few labeled training examples~\citep{koch2015siamese}. Considering the complexity of a hypothesis class, the model training configurations (\ie, hyper-parameters) also serve as a type of inductive biases. \citet{AndrychowiczDCH16Learning, Sachin2017} meta-learn the optimization strategy for each task, including the learning rate and update directions. 
Other kinds of inductive biases are also explored. \citet{HariharanG17Low,Wang2018Low} learn a data generation prior to augment examples given few images; \citet{DaiMWTZ17Logical} extract logical derivations from related tasks; \citet{WangLSHHS17Multi,ShyamGD17Attentive} learn the prior to attend images.

Model-agnostic meta-learning~(MAML)~\citep{FinnAL17Model} proposes another inductive bias, \ie, the model initialization. After the model initialization that is shared among tasks has been meta-trained, the classifier of a new few-shot task can be fine-tuned with several steps of gradient descent from that initial point. 
The universality of this MAML-type updates is proved in~\citep{Finn2018Meta}. MAML has been applied in various scenarios, such as uncertainty estimation~\citep{Finn2018Probabilistic}, robotics control~\citep{Yu2018One,clavera2019learning}, neural translation~\citep{Gu2018Meta}, language generation~\citep{Huang2018Natural}, etc. 

Despite the success, there are still problems with MAML. For example, \citet{Nichol2018Reptile} handle the computational burden by presenting a family of approaches using first-order approximations. \citet{Rajeswaran2019Meta} propose to leverage implicit differentiation, making the calculation of the meta-gradients much efficient and accurate.
\citet{Antoniou2018How} provide a bunch of tricks to train and stabilize the MAML framework. \citet{Bernacchia2021Meta} points out that negative rates of gradient updates help in some scenarios. \citet{rajendran2020meta,yao2021improving,yin2020meta} argue that the learned initialization by MAML may be at high risk of (a) memorization over-fitting, where it solves meta-training tasks without the need of inner loop optimization, or (b) learner over-fitting, where it over-fits to the meta-training
tasks and fails to generalize to the meta-testing
tasks. They thus propose to improve MAML by imposing a regularizer or performing data augmentation.

Since MAML applies a uniform initialization to all the tasks (\ie, the same set of $\{\vw_c\}_{c=1}^N$ and $\vphi$), recent methods explore ways to better incorporate task characteristics. \citet{Lee2019Meta,Bertinetto2018Meta,raghu2020rapid} optimize the linear classifiers $\{\vw_c\}_{c=1}^N$ (not the feature $f_{\vphi}$) till convergence in the inner loop; \citet{Triantafillou2019Meta,ye2020few} initialize the linear classifiers
using class prototypes (\ie, aggregated features per class) so they are task-aware even before the inner loop optimization.
Another direction is to enable task-specific initialization for the entire model~\citep{Requeima2019Fast,Vuorio2019Multimodal,Yao2019Hierarchically,Rusu2018LEO}, which often needs additional sub-networks.

Our work is complementary to the above improvements of MAML: we find an inherent permutation issue of MAML in meta-testing and conduct a detailed analysis. We then build upon it to improve MAML. We note that some of the above methods can be permutation-invariant even though they are not designed for the purpose. For example, LEO~\citep{Rusu2018LEO} computes class prototypes (\ie, averaged features per class) to represent each semantic class\footnote{We note that while \citet{Requeima2019Fast,Vuorio2019Multimodal,Yao2019Hierarchically} also enable task-specific initialization with additional sub-networks for task embedding, their methods cannot resolve the permutation issue. This is because they take an average of the feature embeddings over $N$ classes to represent a task.}. However, it introduces additional sub-networks.
MetaOptNet~\citep{Lee2019Meta} performs inner loop optimization only on $\{\vw_c\}_{c=1}^N$ (till convergence), making it a convex problem which is not sensitive to the initialization and hence the permutations. This method, however, has a high computational burden and needs careful hyper-parameter tuning for the additionally introduced regularizers.
Proto-MAML \citep{Triantafillou2019Meta} initializes the linear classifiers $\{\vw_c\}_{c=1}^N$ with the prototypes, which could be permutation-invariant but cannot achieve accuracy as high as our \ourmethod.
We note that one fundamental difference between our work and \citep{Dhillon2020Baseline}, LEO~\citep{Rusu2018LEO}, and Proto-MAML~\citep{Triantafillou2019Meta} is the motivation:  they aim to provide the initialization of the classifier head with better semantic meanings, while our goal is to resolve the sensitivity of MAML to the permutations.

%% file: main.bbl
\begin{thebibliography}{121}
\providecommand{\natexlab}[1]{#1}
\providecommand{\url}[1]{\texttt{#1}}
\expandafter\ifx\csname urlstyle\endcsname\relax
  \providecommand{\doi}[1]{doi: #1}\else
  \providecommand{\doi}{doi: \begingroup \urlstyle{rm}\Url}\fi

\bibitem[Afrasiyabi et~al.(2020)Afrasiyabi, Lalonde, and
  Gagn{\'e}]{afrasiyabi2020associative}
Arman Afrasiyabi, Jean-Fran{\c{c}}ois Lalonde, and Christian Gagn{\'e}.
\newblock Associative alignment for few-shot image classification.
\newblock In \emph{ECCV}, pp.\  18--35, 2020.

\bibitem[Al-Shedivat et~al.(2018)Al-Shedivat, Bansal, Burda, Sutskever,
  Mordatch, and Abbeel]{al2018continuous}
Maruan Al-Shedivat, Trapit Bansal, Yuri Burda, Ilya Sutskever, Igor Mordatch,
  and Pieter Abbeel.
\newblock Continuous adaptation via meta-learning in nonstationary and
  competitive environments.
\newblock In \emph{ICLR}, 2018.

\bibitem[Andrychowicz et~al.(2016)Andrychowicz, Denil, Colmenarejo, Hoffman,
  Pfau, Schaul, and de~Freitas]{AndrychowiczDCH16Learning}
Marcin Andrychowicz, Misha Denil, Sergio~Gomez Colmenarejo, Matthew~W. Hoffman,
  David Pfau, Tom Schaul, and Nando de~Freitas.
\newblock Learning to learn by gradient descent by gradient descent.
\newblock In \emph{NIPS}, pp.\  3981--3989, 2016.

\bibitem[Antoniou et~al.(2019)Antoniou, Edwards, and Storkey]{Antoniou2018How}
Antreas Antoniou, Harrison Edwards, and Amos~J. Storkey.
\newblock How to train your {MAML}.
\newblock In \emph{ICLR}, 2019.

\bibitem[Arnold \& Sha(2021)Arnold and Sha]{Seb2021Embedding}
S{\'{e}}bastien M.~R. Arnold and Fei Sha.
\newblock Embedding adaptation is still needed for few-shot learning.
\newblock \emph{CoRR}, abs/2104.07255, 2021.

\bibitem[Bachman et~al.(2017)Bachman, Sordoni, and
  Trischler]{bachman2017learning}
Philip Bachman, Alessandro Sordoni, and Adam Trischler.
\newblock Learning algorithms for active learning.
\newblock In \emph{ICML}, pp.\  301--310, 2017.

\bibitem[Balaji et~al.(2018)Balaji, Sankaranarayanan, and
  Chellappa]{balaji2018metareg}
Yogesh Balaji, Swami Sankaranarayanan, and Rama Chellappa.
\newblock Metareg: Towards domain generalization using meta-regularization.
\newblock In \emph{NeurIPS}, pp.\  1006--1016, 2018.

\bibitem[Baxter(2000)]{Baxter2000Inductive}
Jonathan Baxter.
\newblock A model of inductive bias learning.
\newblock \emph{Journal of Artificial Intelligence Research}, 12:\penalty0
  149--198, 2000.

\bibitem[Bello et~al.(2017)Bello, Zoph, Vasudevan, and Le]{bello2017neural}
Irwan Bello, Barret Zoph, Vijay Vasudevan, and Quoc~V Le.
\newblock Neural optimizer search with reinforcement learning.
\newblock In \emph{ICML}, pp.\  459--468, 2017.

\bibitem[Bernacchia(2021)]{Bernacchia2021Meta}
Alberto Bernacchia.
\newblock Meta-learning with negative learning rates.
\newblock In \emph{ICLR}, 2021.

\bibitem[Bertinetto et~al.(2019)Bertinetto, Henriques, Torr, and
  Vedaldi]{Bertinetto2018Meta}
Luca Bertinetto, Jo{\~{a}}o~F. Henriques, Philip H.~S. Torr, and Andrea
  Vedaldi.
\newblock Meta-learning with differentiable closed-form solvers.
\newblock In \emph{ICLR}, 2019.

\bibitem[Breiman(1996)]{breiman1996bagging}
Leo Breiman.
\newblock Bagging predictors.
\newblock \emph{Machine learning}, 24\penalty0 (2):\penalty0 123--140, 1996.

\bibitem[Chen et~al.(2019)Chen, Liu, Kira, Wang, and Huang]{chen2019closer}
Wei-Yu Chen, Yen-Cheng Liu, Zsolt Kira, Yu-Chiang~Frank Wang, and Jia-Bin
  Huang.
\newblock A closer look at few-shot classification.
\newblock In \emph{ICLR}, 2019.

\bibitem[Chen et~al.(2020)Chen, Wang, Tang, and Muandet]{Chen2020MATE}
Xiaohan Chen, Zhangyang Wang, Siyu Tang, and Krikamol Muandet.
\newblock {MATE:} plugging in model awareness to task embedding for meta
  learning.
\newblock In \emph{NeurIPS}, 2020.

\bibitem[Clavera et~al.(2019)Clavera, Nagabandi, Liu, Fearing, Abbeel, Levine,
  and Finn]{clavera2019learning}
Ignasi Clavera, Anusha Nagabandi, Simin Liu, Ronald~S Fearing, Pieter Abbeel,
  Sergey Levine, and Chelsea Finn.
\newblock Learning to adapt in dynamic, real-world environments through
  meta-reinforcement learning.
\newblock In \emph{ICLR}, 2019.

\bibitem[Dai et~al.(2017)Dai, Muggleton, Wen, Tamaddoni-Nezhad, and
  Zhou]{DaiMWTZ17Logical}
Wang-Zhou Dai, Stephen Muggleton, Jing Wen, Alireza Tamaddoni-Nezhad, and
  Zhi-Hua Zhou.
\newblock Logical vision: One-shot meta-interpretive learning from real images.
\newblock In \emph{ILP}, pp.\  46--62, 2017.

\bibitem[Denevi et~al.(2018)Denevi, Ciliberto, Stamos, and
  Pontil]{Denevi2018Learning}
Giulia Denevi, Carlo Ciliberto, Dimitris Stamos, and Massimiliano Pontil.
\newblock Learning to learn around {A} common mean.
\newblock In \emph{NeurIPS}, pp.\  10190--10200, 2018.

\bibitem[Deng et~al.(2009)Deng, Dong, Socher, Li, Li, and
  Fei-Fei]{deng2009imagenet}
Jia Deng, Wei Dong, Richard Socher, Li-Jia Li, Kai Li, and Li~Fei-Fei.
\newblock Imagenet: A large-scale hierarchical image database.
\newblock In \emph{CVPR}, 2009.

\bibitem[Dhillon et~al.(2020)Dhillon, Chaudhari, Ravichandran, and
  Soatto]{Dhillon2020Baseline}
Guneet~Singh Dhillon, Pratik Chaudhari, Avinash Ravichandran, and Stefano
  Soatto.
\newblock A baseline for few-shot image classification.
\newblock In \emph{ICLR}, 2020.

\bibitem[Dietterich(2000)]{dietterich2000ensemble}
Thomas~G Dietterich.
\newblock Ensemble methods in machine learning.
\newblock In \emph{International workshop on multiple classifier systems},
  2000.

\bibitem[Dosovitskiy et~al.(2021)Dosovitskiy, Beyer, Kolesnikov, Weissenborn,
  Zhai, Unterthiner, Dehghani, Minderer, Heigold, Gelly,
  et~al.]{dosovitskiy2020image}
Alexey Dosovitskiy, Lucas Beyer, Alexander Kolesnikov, Dirk Weissenborn,
  Xiaohua Zhai, Thomas Unterthiner, Mostafa Dehghani, Matthias Minderer, Georg
  Heigold, Sylvain Gelly, et~al.
\newblock An image is worth 16x16 words: Transformers for image recognition at
  scale.
\newblock In \emph{ICLR}, 2021.

\bibitem[Duan et~al.(2016)Duan, Schulman, Chen, Bartlett, Sutskever, and
  Abbeel]{duan2016rl}
Yan Duan, John Schulman, Xi~Chen, Peter~L Bartlett, Ilya Sutskever, and Pieter
  Abbeel.
\newblock Rl$^2$: Fast reinforcement learning via slow reinforcement learning.
\newblock \emph{CoRR}, abs/1611.02779, 2016.

\bibitem[Duan et~al.(2017)Duan, Andrychowicz, Stadie, Ho, Schneider, Sutskever,
  Abbeel, and Zaremba]{duan2017one}
Yan Duan, Marcin Andrychowicz, Bradly Stadie, OpenAI~Jonathan Ho, Jonas
  Schneider, Ilya Sutskever, Pieter Abbeel, and Wojciech Zaremba.
\newblock One-shot imitation learning.
\newblock In \emph{NIPS}, pp.\  1087--1098, 2017.

\bibitem[Edwards \& Storkey(2017)Edwards and Storkey]{edwards2017towards}
Harrison Edwards and Amos Storkey.
\newblock Towards a neural statistician.
\newblock In \emph{ICLR}, 2017.

\bibitem[Elsken et~al.(2019)Elsken, Metzen, and Hutter]{elsken2018neural}
Thomas Elsken, Jan~Hendrik Metzen, and Frank Hutter.
\newblock Neural architecture search: {A} survey.
\newblock \emph{Journal of Machine Learning Research}, 20:\penalty0
  55:1--55:21, 2019.

\bibitem[Fan et~al.(2018)Fan, Tian, Qin, Li, and Liu]{fan2018learning}
Yang Fan, Fei Tian, Tao Qin, Xiang-Yang Li, and Tie-Yan Liu.
\newblock Learning to teach.
\newblock In \emph{ICLR}, 2018.

\bibitem[Fei et~al.(2021)Fei, Lu, Xiang, and Huang]{Fei2021MELR}
Nanyi Fei, Zhiwu Lu, Tao Xiang, and Songfang Huang.
\newblock {MELR:} meta-learning via modeling episode-level relationships for
  few-shot learning.
\newblock In \emph{ICLR}, 2021.

\bibitem[Finn(2018)]{finn2018learning3}
Chelsea Finn.
\newblock \emph{Learning to Learn with Gradients}.
\newblock PhD thesis, UC Berkeley, 2018.

\bibitem[Finn \& Levine(2018)Finn and Levine]{Finn2018Meta}
Chelsea Finn and Sergey Levine.
\newblock Meta-learning and universality: Deep representations and gradient
  descent can approximate any learning algorithm.
\newblock In \emph{ICLR}, 2018.

\bibitem[Finn et~al.(2017)Finn, Abbeel, and Levine]{FinnAL17Model}
Chelsea Finn, Pieter Abbeel, and Sergey Levine.
\newblock Model-agnostic meta-learning for fast adaptation of deep networks.
\newblock In \emph{ICML}, pp.\  1126--1135, 2017.

\bibitem[Finn et~al.(2018)Finn, Xu, and Levine]{Finn2018Probabilistic}
Chelsea Finn, Kelvin Xu, and Sergey Levine.
\newblock Probabilistic model-agnostic meta-learning.
\newblock In \emph{NeurIPS}, pp.\  9537--9548, 2018.

\bibitem[Franceschi et~al.(2017)Franceschi, Donini, Frasconi, and
  Pontil]{Franceschi2017Bridge}
Luca Franceschi, Michele Donini, Paolo Frasconi, and Massimiliano Pontil.
\newblock A bridge between hyperparameter optimization and larning-to-learn.
\newblock \emph{CoRR}, abs/1712.06283, 2017.

\bibitem[Frans et~al.(2018)Frans, Ho, Chen, Abbeel, and
  Schulman]{frans2017meta}
Kevin Frans, Jonathan Ho, Xi~Chen, Pieter Abbeel, and John Schulman.
\newblock Meta learning shared hierarchies.
\newblock In \emph{ICLR}, 2018.

\bibitem[Garg \& Kalai(2018)Garg and Kalai]{garg2018supervising}
Vikas Garg and Adam Kalai.
\newblock Supervising unsupervised learning.
\newblock In \emph{NeurIPS}, pp.\  4996--5006, 2018.

\bibitem[Ghiasi et~al.(2018)Ghiasi, Lin, and Le]{Ghiasi2018Drop}
Golnaz Ghiasi, Tsung-Yi Lin, and Quoc~V. Le.
\newblock Dropblock: {A} regularization method for convolutional networks.
\newblock In \emph{NeurIPS}, pp.\  10750--10760, 2018.

\bibitem[Gu et~al.(2018)Gu, Wang, Chen, Li, and Cho]{Gu2018Meta}
Jiatao Gu, Yong Wang, Yun Chen, Victor O.~K. Li, and Kyunghyun Cho.
\newblock Meta-learning for low-resource neural machine translation.
\newblock In \emph{EMNLP}, pp.\  3622--3631, 2018.

\bibitem[Guo et~al.(2016)Guo, Zhang, Hu, He, and Gao]{guo2016ms}
Yandong Guo, Lei Zhang, Yuxiao Hu, Xiaodong He, and Jianfeng Gao.
\newblock Ms-celeb-1m: {A} dataset and benchmark for large-scale face
  recognition.
\newblock In \emph{ECCV}, pp.\  87--102, 2016.

\bibitem[Hariharan \& Girshick(2017)Hariharan and Girshick]{HariharanG17Low}
Bharath Hariharan and Ross~B. Girshick.
\newblock Low-shot visual recognition by shrinking and hallucinating features.
\newblock In \emph{ICCV}, pp.\  3037--3046, 2017.

\bibitem[He et~al.(2016)He, Zhang, Ren, and Sun]{he2016deep}
Kaiming He, Xiangyu Zhang, Shaoqing Ren, and Jian Sun.
\newblock Deep residual learning for image recognition.
\newblock In \emph{CVPR}, pp.\  770--778, 2016.

\bibitem[Hospedales et~al.(2020)Hospedales, Antoniou, Micaelli, and
  Storkey]{hospedales2020meta}
Timothy~M. Hospedales, Antreas Antoniou, Paul Micaelli, and Amos~J. Storkey.
\newblock Meta-learning in neural networks: {A} survey.
\newblock \emph{CoRR}, abs/2004.05439, 2020.

\bibitem[Hsu et~al.(2019)Hsu, Levine, and Finn]{Hsu2018Unsupervised}
Kyle Hsu, Sergey Levine, and Chelsea Finn.
\newblock Unsupervised learning via meta-learning.
\newblock In \emph{ICLR}, 2019.

\bibitem[Huang et~al.(2017)Huang, Liu, Van Der~Maaten, and
  Weinberger]{huang2017densely}
Gao Huang, Zhuang Liu, Laurens Van Der~Maaten, and Kilian~Q Weinberger.
\newblock Densely connected convolutional networks.
\newblock In \emph{CVPR}, pp.\  2261--2269, 2017.

\bibitem[Huang et~al.(2018)Huang, Wang, Singh, tau Yih, and
  He]{Huang2018Natural}
Po-Sen Huang, Chenglong Wang, Rishabh Singh, Wen tau Yih, and Xiaodong He.
\newblock Natural language to structured query generation via meta-learning.
\newblock In \emph{ACL}, pp.\  732--738, 2018.

\bibitem[Joulin et~al.(2016)Joulin, Van Der~Maaten, Jabri, and
  Vasilache]{joulin2016learning}
Armand Joulin, Laurens Van Der~Maaten, Allan Jabri, and Nicolas Vasilache.
\newblock Learning visual features from large weakly supervised data.
\newblock In \emph{ECCV}, pp.\  67--84, 2016.

\bibitem[Kaiser et~al.(2017)Kaiser, Nachum, Roy, and
  Bengio]{kaiser2017learning}
{\L}ukasz Kaiser, Ofir Nachum, Aurko Roy, and Samy Bengio.
\newblock Learning to remember rare events.
\newblock In \emph{ICLR}, 2017.

\bibitem[Koch et~al.(2015)Koch, Zemel, and Salakhutdinov]{koch2015siamese}
Gregory Koch, Richard Zemel, and Ruslan Salakhutdinov.
\newblock Siamese neural networks for one-shot image recognition.
\newblock In \emph{ICML Deep Learning Workshop}, volume~2, 2015.

\bibitem[Krizhevsky et~al.(2012)Krizhevsky, Sutskever, and
  Hinton]{krizhevsky2012imagenet}
Alex Krizhevsky, Ilya Sutskever, and Geoffrey~E Hinton.
\newblock Imagenet classification with deep convolutional neural networks.
\newblock In \emph{NIPS}, pp.\  1106--1114, 2012.

\bibitem[Lee et~al.(2019)Lee, Maji, Ravichandran, and Soatto]{Lee2019Meta}
Kwonjoon Lee, Subhransu Maji, Avinash Ravichandran, and Stefano Soatto.
\newblock Meta-learning with differentiable convex optimization.
\newblock In \emph{CVPR}, pp.\  10657--10665, 2019.

\bibitem[Lemke et~al.(2015)Lemke, Budka, and Gabrys]{lemke2015metalearning}
Christiane Lemke, Marcin Budka, and Bogdan Gabrys.
\newblock Metalearning: a survey of trends and technologies.
\newblock \emph{Artificial intelligence review}, 44\penalty0 (1):\penalty0
  117--130, 2015.

\bibitem[Li et~al.(2018)Li, Yang, Song, and Hospedales]{li2018learning}
Da~Li, Yongxin Yang, Yi-Zhe Song, and Timothy~M Hospedales.
\newblock Learning to generalize: Meta-learning for domain generalization.
\newblock In \emph{AAAI}, pp.\  3490--3497, 2018.

\bibitem[Li et~al.(2020)Li, Zhang, Li, and Fu]{li2020adversarial}
Kai Li, Yulun Zhang, Kunpeng Li, and Yun Fu.
\newblock Adversarial feature hallucination networks for few-shot learning.
\newblock In \emph{CVPR}, pp.\  13470--13479, 2020.

\bibitem[Li \& Malik(2017)Li and Malik]{li2017learning}
Ke~Li and Jitendra Malik.
\newblock Learning to optimize.
\newblock In \emph{ICLR}, 2017.

\bibitem[Li et~al.(2019)Li, Yang, Zhou, and Hospedales]{li2019feature}
Yiying Li, Yongxin Yang, Wei Zhou, and Timothy~M Hospedales.
\newblock Feature-critic networks for heterogeneous domain generalization.
\newblock In \emph{ICML}, pp.\  3915--3924, 2019.

\bibitem[Liu et~al.(2020)Liu, Cao, Lin, Li, Zhang, Long, and
  Hu]{liu2020negative}
Bin Liu, Yue Cao, Yutong Lin, Qi~Li, Zheng Zhang, Mingsheng Long, and Han Hu.
\newblock Negative margin matters: Understanding margin in few-shot
  classification.
\newblock In \emph{ECCV}, pp.\  438--455, 2020.

\bibitem[Mahajan et~al.(2018)Mahajan, Girshick, Ramanathan, He, Paluri, Li,
  Bharambe, and Van Der~Maaten]{mahajan2018exploring}
Dhruv Mahajan, Ross Girshick, Vignesh Ramanathan, Kaiming He, Manohar Paluri,
  Yixuan Li, Ashwin Bharambe, and Laurens Van Der~Maaten.
\newblock Exploring the limits of weakly supervised pretraining.
\newblock In \emph{ECCV}, pp.\  181--196, 2018.

\bibitem[Maurer(2009)]{Maurer09Transfer}
Andreas Maurer.
\newblock Transfer bounds for linear feature learning.
\newblock \emph{Machine Learning}, 75\penalty0 (3):\penalty0 327--350, 2009.

\bibitem[Maurer et~al.(2016)Maurer, Pontil, and
  Romera{-}Paredes]{MaurerPR16The}
Andreas Maurer, Massimiliano Pontil, and Bernardino Romera{-}Paredes.
\newblock The benefit of multitask representation learning.
\newblock \emph{Journal of Machine Learning Research}, 17:\penalty0
  81:1--81:32, 2016.

\bibitem[Metz et~al.(2019)Metz, Maheswaranathan, Cheung, and
  Sohl-Dickstein]{metz2018meta}
Luke Metz, Niru Maheswaranathan, Brian Cheung, and Jascha Sohl-Dickstein.
\newblock Meta-learning update rules for unsupervised representation learning.
\newblock In \emph{ICLR}, 2019.

\bibitem[Motiian et~al.(2017)Motiian, Jones, Iranmanesh, and
  Doretto]{MotiianJID17Few}
Saeid Motiian, Quinn Jones, Seyed~Mehdi Iranmanesh, and Gianfranco Doretto.
\newblock Few-shot adversarial domain adaptation.
\newblock In \emph{NIPS}, pp.\  6673--6683, 2017.

\bibitem[Nichol et~al.(2018)Nichol, Achiam, and Schulman]{Nichol2018Reptile}
Alex Nichol, Joshua Achiam, and John Schulman.
\newblock On first-order meta-learning algorithms.
\newblock \emph{CoRR}, abs/1803.02999, 2018.

\bibitem[Pang et~al.(2018)Pang, Dong, Wu, and Hospedales]{pang2018meta}
Kunkun Pang, Mingzhi Dong, Yang Wu, and Timothy Hospedales.
\newblock Meta-learning transferable active learning policies by deep
  reinforcement learning.
\newblock \emph{CoRR}, abs/1806.04798, 2018.

\bibitem[Peng et~al.()Peng, Li, Zhang, Li, Qi, and Tang]{Peng2019Few}
Zhimao Peng, Zechao Li, Junge Zhang, Yan Li, Guo-Jun Qi, and Jinhui Tang.
\newblock Few-shot image recognition with knowledge transfer.
\newblock In \emph{ICCV}, pp.\  441--449.

\bibitem[Probst et~al.(2019)Probst, Boulesteix, and
  Bischl]{Probst2019Tunability}
Philipp Probst, Anne-Laure Boulesteix, and Bernd Bischl.
\newblock Tunability: Importance of hyperparameters of machine learning
  algorithms.
\newblock \emph{Journal of Machine Learning Research}, 20:\penalty0
  53:1--53:32, 2019.

\bibitem[Qiao et~al.(2018)Qiao, Liu, Shen, and Yuille]{qiao2018few}
Siyuan Qiao, Chenxi Liu, Wei Shen, and Alan~L Yuille.
\newblock Few-shot image recognition by predicting parameters from activations.
\newblock In \emph{CVPR}, pp.\  7229--7238, 2018.

\bibitem[Raghu et~al.(2020)Raghu, Raghu, Bengio, and Vinyals]{raghu2020rapid}
Aniruddh Raghu, Maithra Raghu, Samy Bengio, and Oriol Vinyals.
\newblock Rapid learning or feature reuse? towards understanding the
  effectiveness of maml.
\newblock In \emph{ICLR}, 2020.

\bibitem[Rajendran et~al.(2020)Rajendran, Irpan, and Jang]{rajendran2020meta}
Janarthanan Rajendran, Alex Irpan, and Eric Jang.
\newblock Meta-learning requires meta-augmentation.
\newblock In \emph{NeurIPS}, 2020.

\bibitem[Rajeswaran et~al.(2019)Rajeswaran, Finn, Kakade, and
  Levine]{Rajeswaran2019Meta}
Aravind Rajeswaran, Chelsea Finn, Sham~M. Kakade, and Sergey Levine.
\newblock Meta-learning with implicit gradients.
\newblock In \emph{NeurIPS}, pp.\  113--124, 2019.

\bibitem[Ratner et~al.(2017)Ratner, Ehrenberg, Hussain, Dunnmon, and
  R{\'e}]{ratner2017learning}
Alexander~J Ratner, Henry Ehrenberg, Zeshan Hussain, Jared Dunnmon, and
  Christopher R{\'e}.
\newblock Learning to compose domain-specific transformations for data
  augmentation.
\newblock In \emph{NIPS}, pp.\  3236--3246, 2017.

\bibitem[Ravi \& Larochelle(2017)Ravi and Larochelle]{Sachin2017}
Sachin Ravi and Hugo Larochelle.
\newblock Optimization as a model for few-shot learning.
\newblock In \emph{ICLR}, 2017.

\bibitem[Ravi \& Larochelle(2018)Ravi and Larochelle]{ravi2018meta}
Sachin Ravi and Hugo Larochelle.
\newblock Meta-learning for batch mode active learning.
\newblock In \emph{ICLR Workshop}, 2018.

\bibitem[Reed et~al.(2018)Reed, Chen, Paine, van~den Oord, Eslami, Rezende,
  Vinyals, and de~Freitas]{Reed2017Few}
Scott~E. Reed, Yutian Chen, Thomas Paine, A{\"{a}}ron van~den Oord, S.~M.~Ali
  Eslami, Danilo~Jimenez Rezende, Oriol Vinyals, and Nando de~Freitas.
\newblock Few-shot autoregressive density estimation: Towards learning to learn
  distributions.
\newblock In \emph{ICLR}, 2018.

\bibitem[Ren et~al.(2018{\natexlab{a}})Ren, Triantafillou, Ravi, Snell,
  Swersky, Tenenbaum, Larochelle, and Zemel]{ren2018meta}
Mengye Ren, Eleni Triantafillou, Sachin Ravi, Jake Snell, Kevin Swersky,
  Joshua~B Tenenbaum, Hugo Larochelle, and Richard~S Zemel.
\newblock Meta-learning for semi-supervised few-shot classification.
\newblock In \emph{ICLR}, 2018{\natexlab{a}}.

\bibitem[Ren et~al.(2018{\natexlab{b}})Ren, Zeng, Yang, and
  Urtasun]{Ren2018Reweight}
Mengye Ren, Wenyuan Zeng, Bin Yang, and Raquel Urtasun.
\newblock Learning to reweight examples for robust deep learning.
\newblock In \emph{ICML}, pp.\  4331--4340, 2018{\natexlab{b}}.

\bibitem[Requeima et~al.(2019)Requeima, Gordon, Bronskill, Nowozin, and
  Turner]{Requeima2019Fast}
James Requeima, Jonathan Gordon, John Bronskill, Sebastian Nowozin, and
  Richard~E. Turner.
\newblock Fast and flexible multi-task classification using conditional neural
  adaptive processes.
\newblock In \emph{NeurIPS}, pp.\  7957--7968, 2019.

\bibitem[Riemer et~al.(2019)Riemer, Cases, Ajemian, Liu, Rish, Tu, and
  Tesauro]{riemer2019learning}
Matthew Riemer, Ignacio Cases, Robert Ajemian, Miao Liu, Irina Rish, Yuhai Tu,
  and Gerald Tesauro.
\newblock Learning to learn without forgetting by maximizing transfer and
  minimizing interference.
\newblock In \emph{ICLR}, 2019.

\bibitem[Russakovsky et~al.(2015)Russakovsky, Deng, Su, Krause, Satheesh, Ma,
  Huang, Karpathy, Khosla, Bernstein, Berg, and Li]{RussakovskyDSKS15ImageNet}
Olga Russakovsky, Jia Deng, Hao Su, Jonathan Krause, Sanjeev Satheesh, Sean Ma,
  Zhiheng Huang, Andrej Karpathy, Aditya Khosla, Michael~S. Bernstein,
  Alexander~C. Berg, and Fei-Fei Li.
\newblock Imagenet large scale visual recognition challenge.
\newblock \emph{IJCV}, 115\penalty0 (3):\penalty0 211--252, 2015.

\bibitem[Rusu et~al.(2019)Rusu, Rao, Sygnowski, Vinyals, Pascanu, Osindero, and
  Hadsell]{Rusu2018LEO}
Andrei~A. Rusu, Dushyant Rao, Jakub Sygnowski, Oriol Vinyals, Razvan Pascanu,
  Simon Osindero, and Raia Hadsell.
\newblock Meta-learning with latent embedding optimization.
\newblock In \emph{ICLR}, 2019.

\bibitem[Sharma et~al.(2018)Sharma, Jha, Hegde, and
  Ravindran]{sharma2018learning}
Sahil Sharma, Ashutosh Jha, Parikshit Hegde, and Balaraman Ravindran.
\newblock Learning to multi-task by active sampling.
\newblock In \emph{ICLR}, 2018.

\bibitem[Shin et~al.(2021)Shin, Lee, Gong, and Hwang]{shin2021large}
Jaewoong Shin, Hae~Beom Lee, Boqing Gong, and Sung~Ju Hwang.
\newblock Large-scale meta-learning with continual trajectory shifting.
\newblock In \emph{ICML}, pp.\  9603--9613, 2021.

\bibitem[Shyam et~al.(2017)Shyam, Gupta, and Dukkipati]{ShyamGD17Attentive}
Pranav Shyam, Shubham Gupta, and Ambedkar Dukkipati.
\newblock Attentive recurrent comparators.
\newblock In \emph{ICML}, pp.\  3173--3181, 2017.

\bibitem[Simon et~al.(2020)Simon, Koniusz, Nock, and
  Harandi]{Simon2020Adaptive}
Christian Simon, Piotr Koniusz, Richard Nock, and Mehrtash Harandi.
\newblock Adaptive subspaces for few-shot learning.
\newblock In \emph{CVPR}, pp.\  4135--4144, 2020.

\bibitem[Simonyan \& Zisserman(2015)Simonyan and Zisserman]{simonyan2015very}
Karen Simonyan and Andrew Zisserman.
\newblock Very deep convolutional networks for large-scale image recognition.
\newblock In \emph{ICLR}, 2015.

\bibitem[Snell et~al.(2017)Snell, Swersky, and Zemel]{SnellSZ17Prototypical}
Jake Snell, Kevin Swersky, and Richard~S. Zemel.
\newblock Prototypical networks for few-shot learning.
\newblock In \emph{NIPS}, pp.\  4080--4090, 2017.

\bibitem[Srivastava et~al.(2014)Srivastava, Hinton, Krizhevsky, Sutskever, and
  Salakhutdinov]{srivastava2014dropout}
Nitish Srivastava, Geoffrey Hinton, Alex Krizhevsky, Ilya Sutskever, and Ruslan
  Salakhutdinov.
\newblock Dropout: a simple way to prevent neural networks from overfitting.
\newblock \emph{Journal of Machine Learning Research}, 15\penalty0
  (1):\penalty0 1929--1958, 2014.

\bibitem[Stadie et~al.(2018)Stadie, Yang, Houthooft, Chen, Duan, Wu, Abbeel,
  and Sutskever]{stadie2018importance}
Bradly Stadie, Ge~Yang, Rein Houthooft, Peter Chen, Yan Duan, Yuhuai Wu, Pieter
  Abbeel, and Ilya Sutskever.
\newblock The importance of sampling in meta-reinforcement learning.
\newblock In \emph{NeurIPS}, pp.\  9300--9310, 2018.

\bibitem[Sun et~al.(2019)Sun, Liu, Chen, Chua, and Schiele]{Sun2019Meta}
Qianru Sun, Yaoyao Liu, Zhaozheng Chen, Tat-Seng Chua, and Bernt Schiele.
\newblock Meta-transfer learning through hard tasks.
\newblock \emph{CoRR}, abs/1910.03648, 2019.

\bibitem[Sung et~al.(2018)Sung, Yang, Zhang, Xiang, Torr, and
  Hospedales]{Flood2017Learning}
Flood Sung, Yongxin Yang, Li~Zhang, Tao Xiang, Philip H.~S. Torr, and
  Timothy~M. Hospedales.
\newblock Learning to compare: Relation network for few-shot learning.
\newblock In \emph{CVPR}, pp.\  1199--1208, 2018.

\bibitem[Szegedy et~al.(2015)Szegedy, Liu, Jia, Sermanet, Reed, Anguelov,
  Erhan, Vanhoucke, and Rabinovich]{szegedy2015going}
Christian Szegedy, Wei Liu, Yangqing Jia, Pierre Sermanet, Scott Reed, Dragomir
  Anguelov, Dumitru Erhan, Vincent Vanhoucke, and Andrew Rabinovich.
\newblock Going deeper with convolutions.
\newblock In \emph{CVPR}, pp.\  1--9, 2015.

\bibitem[Thomee et~al.(2016)Thomee, Shamma, Friedland, Elizalde, Ni, Poland,
  Borth, and Li]{thomee2015yfcc100m}
Bart Thomee, David~A Shamma, Gerald Friedland, Benjamin Elizalde, Karl Ni,
  Douglas Poland, Damian Borth, and Li-Jia Li.
\newblock Yfcc100m: The new data in multimedia research.
\newblock \emph{Communications of {ACM}}, 59\penalty0 (2):\penalty0 64--73,
  2016.

\bibitem[Thrun \& Pratt(2012)Thrun and Pratt]{thrun2012learning}
Sebastian Thrun and Lorien Pratt.
\newblock \emph{Learning to learn}.
\newblock Springer Science \& Business Media, 2012.

\bibitem[Tian et~al.(2020)Tian, Wang, Krishnan, Tenenbaum, and
  Isola]{Tian2020Rethinking}
Yonglong Tian, Yue Wang, Dilip Krishnan, Joshua~B. Tenenbaum, and Phillip
  Isola.
\newblock Rethinking few-shot image classification: {A} good embedding is all
  you need?
\newblock In \emph{ECCV}, pp.\  266--282, 2020.

\bibitem[Triantafillou et~al.(2020)Triantafillou, Zhu, Dumoulin, Lamblin, Xu,
  Goroshin, Gelada, Swersky, Manzagol, and Larochelle]{Triantafillou2019Meta}
Eleni Triantafillou, Tyler Zhu, Vincent Dumoulin, Pascal Lamblin, Kelvin Xu,
  Ross Goroshin, Carles Gelada, Kevin Swersky, Pierre-Antoine Manzagol, and
  Hugo Larochelle.
\newblock Meta-dataset: {A} dataset of datasets for learning to learn from few
  examples.
\newblock In \emph{ICLR}, 2020.

\bibitem[Tseng et~al.(2020)Tseng, Lee, Huang, and Yang]{Tseng2020Cross}
Hung-Yu Tseng, Hsin-Ying Lee, Jia-Bin Huang, and Ming-Hsuan Yang.
\newblock Cross-domain few-shot classification via learned feature-wise
  transformation.
\newblock In \emph{ICLR}, 2020.

\bibitem[Vanschoren(2018)]{vanschoren2018meta}
Joaquin Vanschoren.
\newblock Meta-learning: A survey.
\newblock \emph{CoRR}, abs/1810.03548, 2018.

\bibitem[Vartak et~al.(2017)Vartak, Thiagarajan, Miranda, Bratman, and
  Larochelle]{Vartak2017Meta}
Manasi Vartak, Arvind Thiagarajan, Conrado Miranda, Jeshua Bratman, and Hugo
  Larochelle.
\newblock A meta-learning perspective on cold-start recommendations for items.
\newblock In \emph{NIPS}, pp.\  6907--6917, 2017.

\bibitem[Vilalta \& Drissi(2002)Vilalta and Drissi]{Vilalta2002Meta}
Ricardo Vilalta and Youssef Drissi.
\newblock A perspective view and survey of meta-learning.
\newblock \emph{Artificial Intelligence Review}, 18\penalty0 (2):\penalty0
  77--95, 2002.

\bibitem[Vinyals et~al.(2016)Vinyals, Blundell, Lillicrap, Kavukcuoglu, and
  Wierstra]{VinyalsBLKW16Matching}
Oriol Vinyals, Charles Blundell, Tim Lillicrap, Koray Kavukcuoglu, and Daan
  Wierstra.
\newblock Matching networks for one shot learning.
\newblock In \emph{NIPS}, pp.\  3630--3638, 2016.

\bibitem[Vuorio et~al.(2019)Vuorio, Sun, Hu, and Lim]{Vuorio2019Multimodal}
Risto Vuorio, Shao-Hua Sun, Hexiang Hu, and Joseph~J. Lim.
\newblock Multimodal model-agnostic meta-learning via task-aware modulation.
\newblock In \emph{NeurIPS}, pp.\  1--12, 2019.

\bibitem[Wah et~al.(2011)Wah, Branson, Welinder, Perona, and
  Belongie]{WahCUB_200_2011}
C.~Wah, S.~Branson, P.~Welinder, P.~Perona, and S.~Belongie.
\newblock {The Caltech-UCSD Birds-200-2011 Dataset}.
\newblock Technical Report CNS-TR-2011-001, California Institute of Technology,
  2011.

\bibitem[Wang et~al.(2017{\natexlab{a}})Wang, Kurth-Nelson, Tirumala, Soyer,
  Leibo, Munos, Blundell, Kumaran, and Botvinick]{wang2016RL}
Jane~X Wang, Zeb Kurth-Nelson, Dhruva Tirumala, Hubert Soyer, Joel~Z Leibo,
  Remi Munos, Charles Blundell, Dharshan Kumaran, and Matt Botvinick.
\newblock Learning to reinforcement learn.
\newblock In \emph{CogSci}, 2017{\natexlab{a}}.

\bibitem[Wang et~al.(2017{\natexlab{b}})Wang, Liu, Shen, Huang, van~den Hengel,
  and Shen]{WangLSHHS17Multi}
Peng Wang, Lingqiao Liu, Chunhua Shen, Zi~Huang, Anton van~den Hengel, and
  Heng~Tao Shen.
\newblock Multi-attention network for one shot learning.
\newblock In \emph{CVPR}, pp.\  6212--6220, 2017{\natexlab{b}}.

\bibitem[Wang et~al.(2018{\natexlab{a}})Wang, Zhu, Torralba, and
  Efros]{Wang2018Dataset}
Tongzhou Wang, Jun-Yan Zhu, Antonio Torralba, and Alexei~A. Efros.
\newblock Dataset distillation.
\newblock \emph{CoRR}, abs/1811.10959, 2018{\natexlab{a}}.

\bibitem[Wang et~al.(2019)Wang, Chao, Weinberger, and van~der
  Maaten]{wang2019simpleshot}
Yan Wang, Wei-Lun Chao, Kilian~Q Weinberger, and Laurens van~der Maaten.
\newblock Simpleshot: Revisiting nearest-neighbor classification for few-shot
  learning.
\newblock \emph{CoRR}, abs/1911.04623, 2019.

\bibitem[Wang et~al.(2020)Wang, Yao, Kwok, and Ni]{wang2020generalizing}
Yaqing Wang, Quanming Yao, James~T Kwok, and Lionel~M Ni.
\newblock Generalizing from a few examples: A survey on few-shot learning.
\newblock \emph{ACM Computing Surveys (CSUR)}, 53\penalty0 (3):\penalty0 1--34,
  2020.

\bibitem[Wang \& Hebert(2016)Wang and Hebert]{WangH16Learning}
Yu-Xiong Wang and Martial Hebert.
\newblock Learning to learn: Model regression networks for easy small sample
  learning.
\newblock In \emph{ECCV}, pp.\  616--634, 2016.

\bibitem[Wang et~al.(2017{\natexlab{c}})Wang, Ramanan, and
  Hebert]{WangRH17Learning}
Yu{-}Xiong Wang, Deva Ramanan, and Martial Hebert.
\newblock Learning to model the tail.
\newblock In \emph{NIPS}, pp.\  7032--7042, 2017{\natexlab{c}}.

\bibitem[Wang et~al.(2018{\natexlab{b}})Wang, Girshick, Hebert, and
  Hariharan]{Wang2018Low}
Yu-Xiong Wang, Ross~B. Girshick, Martial Hebert, and Bharath Hariharan.
\newblock Low-shot learning from imaginary data.
\newblock In \emph{CVPR}, pp.\  7278--7286, 2018{\natexlab{b}}.

\bibitem[Wichrowska et~al.(2017)Wichrowska, Maheswaranathan, Hoffman,
  Colmenarejo, Denil, de~Freitas, and Sohl-Dickstein]{wichrowska2017learned}
Olga Wichrowska, Niru Maheswaranathan, Matthew~W Hoffman, Sergio~Gomez
  Colmenarejo, Misha Denil, Nando de~Freitas, and Jascha Sohl-Dickstein.
\newblock Learned optimizers that scale and generalize.
\newblock In \emph{ICML}, pp.\  3751--3760, 2017.

\bibitem[Wu et~al.(2019)Wu, Li, Guo, and Jia]{Wu2019PARN}
Ziyang Wu, Yuwei Li, Lihua Guo, and Kui Jia.
\newblock {PARN:} position-aware relation networks for few-shot learning.
\newblock In \emph{ICCV}, pp.\  6658--6666, 2019.

\bibitem[Yao et~al.(2019)Yao, Wei, Huang, and Li]{Yao2019Hierarchically}
Huaxiu Yao, Ying Wei, Junzhou Huang, and Zhenhui Li.
\newblock Hierarchically structured meta-learning.
\newblock In \emph{ICML}, pp.\  7045--7054, 2019.

\bibitem[Yao et~al.(2021)Yao, Huang, Zhang, Wei, Tian, Zou, Huang,
  et~al.]{yao2021improving}
Huaxiu Yao, Long-Kai Huang, Linjun Zhang, Ying Wei, Li~Tian, James Zou, Junzhou
  Huang, et~al.
\newblock Improving generalization in meta-learning via task augmentation.
\newblock In \emph{ICML}, pp.\  11887--11897, 2021.

\bibitem[Ye et~al.(2020{\natexlab{a}})Ye, Hu, Zhan, and Sha]{ye2020fewshot}
Han-Jia Ye, Hexiang Hu, De-Chuan Zhan, and Fei Sha.
\newblock Few-shot learning via embedding adaptation with set-to-set functions.
\newblock In \emph{CVPR}, pp.\  8805--8814, 2020{\natexlab{a}}.

\bibitem[Ye et~al.(2020{\natexlab{b}})Ye, Sheng, and Zhan]{ye2020few}
Han-Jia Ye, Xiang-Rong Sheng, and De-Chuan Zhan.
\newblock Few-shot learning with adaptively initialized task optimizer: a
  practical meta-learning approach.
\newblock \emph{Machine Learning}, 109\penalty0 (3):\penalty0 643--664,
  2020{\natexlab{b}}.

\bibitem[Ye et~al.(2021)Ye, Ming, Zhan, and Chao]{Ye2021LastShot}
Han-Jia Ye, Lu~Ming, De-Chuan Zhan, and Wei-Lun Chao.
\newblock Few-shot learning with a strong teacher.
\newblock \emph{CoRR}, abs/2107.00197, 2021.

\bibitem[Yin et~al.(2020)Yin, Tucker, Zhou, Levine, and Finn]{yin2020meta}
Mingzhang Yin, George Tucker, Mingyuan Zhou, Sergey Levine, and Chelsea Finn.
\newblock Meta-learning without memorization.
\newblock In \emph{ICLR}, 2020.

\bibitem[Ying et~al.(2018)Ying, Zhang, Huang, and Yang]{ying2018transfer}
Wei Ying, Yu~Zhang, Junzhou Huang, and Qiang Yang.
\newblock Transfer learning via learning to transfer.
\newblock In \emph{ICML}, pp.\  5072--5081, 2018.

\bibitem[Yu et~al.(2018)Yu, Finn, Dasari, Xie, Zhang, Abbeel, and
  Levine]{Yu2018One}
Tianhe Yu, Chelsea Finn, Sudeep Dasari, Annie Xie, Tianhao Zhang, Pieter
  Abbeel, and Sergey Levine.
\newblock One-shot imitation from observing humans via domain-adaptive
  meta-learning.
\newblock In \emph{Robotics: Science and Systems}, 2018.

\bibitem[Zhang et~al.(2020)Zhang, Cai, Lin, and Shen]{Zhang2020Deep}
Chi Zhang, Yujun Cai, Guosheng Lin, and Chunhua Shen.
\newblock Deepemd: Few-shot image classification with differentiable earth
  mover's distance and structured classifiers.
\newblock In \emph{CVPR}, pp.\  12200--12210, 2020.

\bibitem[Zhang et~al.(2018{\natexlab{a}})Zhang, Che, Ghahramani, Bengio, and
  Song]{zhang2018metagan}
Ruixiang Zhang, Tong Che, Zoubin Ghahramani, Yoshua Bengio, and Yangqiu Song.
\newblock Metagan: An adversarial approach to few-shot learning.
\newblock In \emph{NeurIPS}, pp.\  2371--2380, 2018{\natexlab{a}}.

\bibitem[Zhang et~al.(2018{\natexlab{b}})Zhang, Wei, and
  Yang]{zhang2018learning}
Yu~Zhang, Ying Wei, and Qiang Yang.
\newblock Learning to multitask.
\newblock In \emph{NeurIPS}, pp.\  5776--5787, 2018{\natexlab{b}}.

\bibitem[Zhou(2012)]{zhou2012ensemble}
Zhi-Hua Zhou.
\newblock \emph{Ensemble methods: foundations and algorithms}.
\newblock Chapman and Hall/CRC, 2012.

\end{thebibliography}
